%% file: main.tex
\documentclass[11pt]{article}

\PassOptionsToPackage{table,xcdraw,usenames,dvipsnames}{xcolor}
\PassOptionsToPackage{hyperfootnotes=false}{hyperref}

\usepackage[preprint]{acl}

\usepackage{times}
\usepackage{latexsym}
\usepackage[T1]{fontenc}
\usepackage[utf8]{inputenc}
\usepackage{microtype}
\usepackage{inconsolata}

\usepackage{booktabs}
\usepackage{graphicx}
\usepackage{amsfonts}
\usepackage{nicefrac}
\usepackage{amsmath}
\usepackage{amssymb}
\usepackage{mathtools}
\usepackage{amsthm}
\usepackage[capitalize,noabbrev]{cleveref}
\usepackage{dsfont}
\usepackage{enumitem}
\usepackage{wrapfig}
\usepackage{adjustbox}
\usepackage{multirow}
\usepackage{subcaption}
\usepackage{xspace}
\usepackage{pifont}
\usepackage{algorithmic}

\input{math_commands.tex}

\newcommand{\ds}[1]{}

\theoremstyle{plain}
\newtheorem{theorem}{Theorem}[section]

\theoremstyle{definition}

\theoremstyle{remark}

\title{Towards Structurally Explainable Machine-Generated Text Detection: A Graph-Perspective Framework}

\author{
  \textbf{Xu Zheng\textsuperscript{1}},
  \textbf{Zhuomin Chen\textsuperscript{1}},
  \textbf{Esteban Schafir\textsuperscript{1}},
  \textbf{Sipeng Chen\textsuperscript{2}},
  \textbf{Hojat Allah Salehi\textsuperscript{1}},\\
  \textbf{Haifeng Chen\textsuperscript{3}},
  \textbf{Farhad Shirani\textsuperscript{1}},
  \textbf{Mo Sha\textsuperscript{1}},
  \textbf{Wei Cheng\textsuperscript{3}},
  \textbf{Dongsheng Luo\textsuperscript{4}}\Thanks{Corresponding author. This work was primarily conducted while Dongsheng Luo was at Florida International University. Email: \texttt{luodongsheng01@gmail.com}.}\\
  \textsuperscript{1}Florida International University, Miami, FL, USA\\
  \textsuperscript{2}Florida State University, Tallahassee, FL, USA\\
  \textsuperscript{3}NEC Laboratories America, Princeton, NJ, USA\\
  \textsuperscript{4}Singapore Management University, Singapore
}

\newcommand{\ours}{\textsc{LM$^2$otifs}}

\begin{document}
\maketitle

\begin{abstract}
Despite the success of machine-generated text detectors, the black-box nature remains a critical limitation.
Traditional explainability methods rely on token-level saliency, insufficient to reveal the high-order structural dependencies that distinguish LLM outputs. In this paper, we propose {\ours}, a principled framework that shifts detection from linear sequences to graph-structured manifolds. We first provide a theoretical grounding based on probabilistic graphical models, demonstrating that detection performance is more distinguishable in the graph-topological space. Driven by this theory, {\ours} transforms text into lexical co-occurrence graphs to preserve latent structural fingerprints. The framework employs Graph Neural Networks for robust detection and utilizes graph-specific explainers to extract interpretable motifs.
Crucially, our experiments reveal that these structural motifs achieve higher faithfulness compared to traditional methods. This empirical evidence confirms the existence of high-order structural explanations that linear methods fail to capture. Experimental results show that {\ours} achieves state-of-the-art performance while providing multi-level \textit{distinct linguistic fingerprints} that are more faithful to the model's decision.
\end{abstract}

\input{1_introduction}

\input{2_preliminary}

\input{2_5_Theortical_analysis}

\input{3_method}

\input{4_relatedwork}
\input{5_Experiments}

\input{6_conclusion}

\bibliography{example_paper}

\newpage
\appendix
\input{7_appendix}

\end{document}

%% file: math_commands.tex
\usepackage{amsmath,amsfonts,bm}

\def\eqref#1{equation~\ref{#1}}

\def\1{\bm{1}}

\def\mA{{\bm{A}}}

\def\mH{{\bm{H}}}

\def\mX{{\bm{X}}}

\def\mZ{{\bm{Z}}}

\DeclareMathAlphabet{\mathsfit}{\encodingdefault}{\sfdefault}{m}{sl}
\SetMathAlphabet{\mathsfit}{bold}{\encodingdefault}{\sfdefault}{bx}{n}

\def\sR{{\mathbb{R}}}

\DeclareMathOperator*{\argmin}{arg\,min}

\newcommand{\stitle}[1]{\vspace{1ex}\noindent{\bf #1}}

\newcommand{\ms}[2]{{#1\tiny{$\pm$#2}}}

%% file: 1_introduction.tex
\section{Introduction}
The proliferation of Large Language Models (LLMs), such as ChatGPT~\cite{chatgpt}, Llama~\cite{touvron2023llama}, and Claude~\cite{claude}, has fundamentally reshaped digital content creation. 
While these models exhibit human-level proficiency in writing~\cite{yuan2022wordcraft} and reasoning~\cite{zhang2024pybench}, coding~\cite{zhang2024pybench}, and question answering~\cite{NEURIPS2023_9cb2a749}, their widespread adoption has raised profound concerns regarding content authenticity, including the spread of misinformation~\cite{chen2024combating}, fake news~\citep{ahmed2021detecting}, and academic plagiarism~\cite{lee2023language}. 
As the distinction between machine-generated text (MGT) and human-generated text (HGT) becomes increasingly difficult for humans to perceive~\cite{gehrmann2019gltr}, developing reliable detectors has become essential.

\begin{figure}[t]
  \centering
        \includegraphics[width=0.49\textwidth]{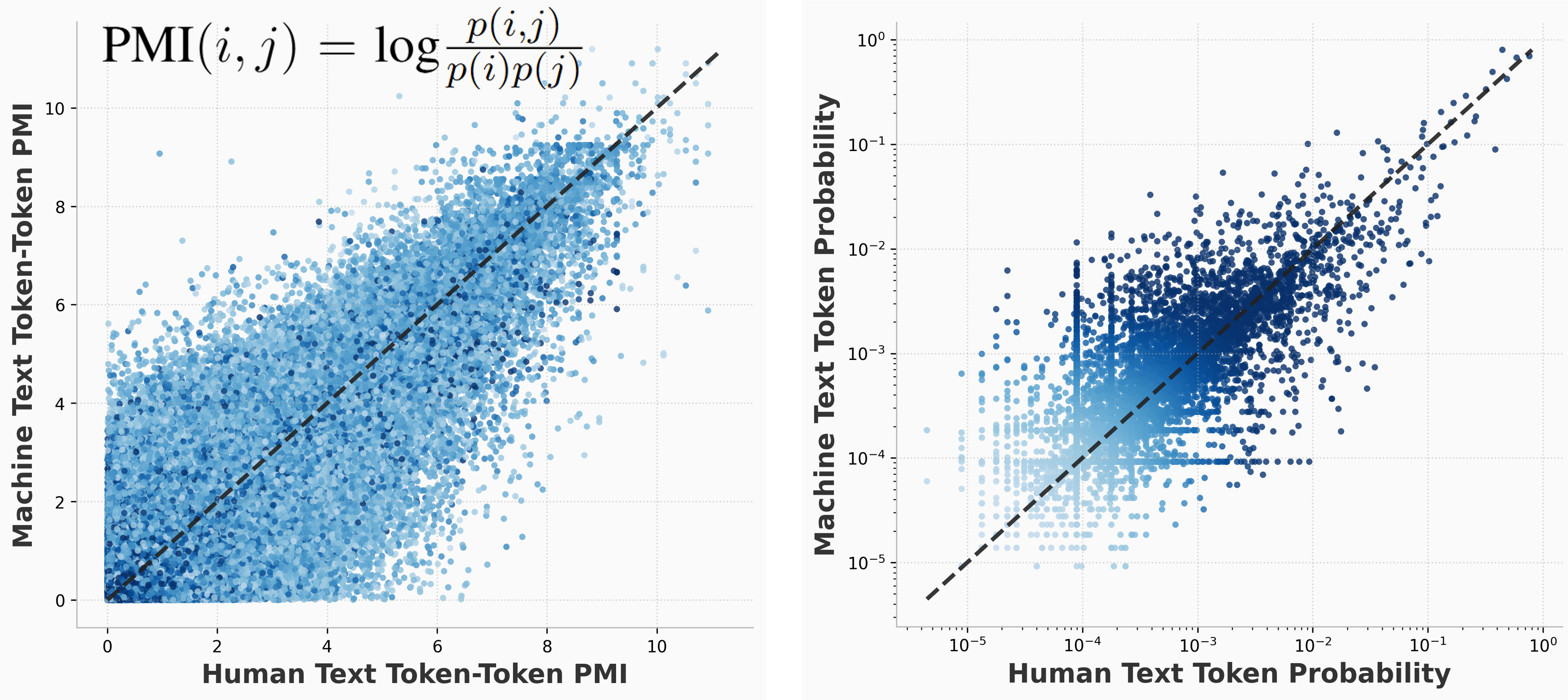}
  
\vspace{-0.2cm}
\caption{Statistical comparison of token and co-occurrence distributions between GPT-4-generated text and human-generated text on the Yelp Dataset dataset~\cite{mao2024raidar}. We compute PMI (pointwise mutual information) for token pairs within a fixed context window. Colors indicate the overall probability density, while the coordinates correspond to values in HGT and MGT, respectively. Points close to the diagonal line represent similar statistical patterns between the two distributions. The results show that co-occurrence patterns exhibit substantially stronger discriminative power than token probabilities for detection tasks.}

\label{fig:motivation_coorrance}
\vspace{-0.6cm}
\end{figure}
Existing detection paradigms~\cite{yangdna,guo2024detective} typically treat text as linear sequences, operating either as white-box models that analyze token-level probability distributions~\cite{su2023detectllm} or black-box classifiers that learn latent representations~\cite{sotofew,zhangdetecting,nguyen2024simllm} from a pretrained model. 
However, these methods are largely trapped in an ``accuracy-first'' paradigm. Despite achieving high accuracy, they suffer from a critical transparency gap~\cite{wu2025survey} that they output a mere probability score without providing interpretable evidence of why the text is machine-generated. In high-risk scenarios such as academic integrity hearings, medical record auditing, and legal investigations, such opaque predictions are inadequate for definitive judgments, creating an urgent need for ``white-box'' interpretability.
Furthermore, traditional explainability techniques are ill-suited for this domain. Gradient-based methods~\cite{sundararajan2017axiomatic, selvaraju2017grad} are computationally prohibitive for modern LLMs, while attention-based saliency~\cite{jain2019attention} often yields fragmented token-level heatmaps. 
More importantly, as illustrated in Figure~\ref{fig:explainable_baseline_avg}, existing token-level explanations often exhibit low faithfulness, failing to capture the detector's true decision rationale. This suggests that the distinction between MGT and HGT is not primarily encoded in isolated keywords, but in high-order structural dependencies. As further evidenced in Figure~\ref{fig:motivation_coorrance}, token co-occurrence distribution has a substantially distinguishable pattern over token distribution. This structural fingerprint explains the limitation of conventional sequence-based detectors and motivates higher-order representations for robust and explainable detection.

We attribute the limitations of current detectors to their reliance on a sequential perspective. The fundamental architecture of modern LLM builds upon the principle of autoregressive next-token prediction, which models the joint probability distribution of a sequence as $P(s_1,s_2,\cdots,s_T)\approx \prod_{t=1}^{T}P_{\theta}(s_t|s_{1:t-1})$, where $\theta$ is the (trainable) model parameter, $s_i$ is the word/token at the $i$th position, and $T$ is bounded by the context length ~\cite{radford2019language,bengio2000neural}. While effective for generation, this linear factorization often overlooks the underlying distribution's structural manifold. 
By revisiting the problem through the lens of Probabilistic Graphical Models (PGM)~\cite{bishop2006pattern}, we observe a fundamental asymmetry that while generative tasks require accurately approximating the posterior distribution, detection tasks only require identifying a discriminative graph structure that separates MGT from HGT. This insight suggests that a graph-based formulation is particularly well-suited for detection, as it allows the model to extract discriminative structural dependencies between tokens that distinguish MGT from HGT, while avoiding the unnecessary complexity of full generative modeling.

Drawing inspiration from PGM, we introduce a principled and structurally explainable framework for MGT detection, {\ours}~(\textit{Language Model Motifs}). Instead of treating text as a flat sequence, {\ours} projects it into a lexical co-occurrence graph, capturing latent structural artifacts that sequence-based models often smooth over. By shifting the detection task to a graph-structured space, we can leverage eXplainable Graph Neural Networks (XGNNs) to extract interpretable motifs, subgraph structures that serve as explicit evidence for machine authorship.
Crucially, our theoretical analysis shows that the detection criterion can be more effectively characterized in a graph-topological space, where the structural artifacts of machine logic become explicit. The main contributions of this paper are: 
\begin{itemize}[leftmargin=*,itemsep=1.5pt,topsep=2pt,parsep=0pt]
    \item[\ding{72}] We propose {\ours}, a novel framework that integrates co-occurrence graphs with GNNs, converting the MGT detection paradigm from linear sequences to topological patterns.  By leveraging explainable graph methods, {\ours} provides the ability to extract structural linguistic fingerprints from graph-based detectors.
    \item[\ding{72}] We provide a theoretical analysis of the detection rationale from a PGM perspective, formally justifying why graph-based representations provide a more discriminative accuracy for MGT. 
    \item[\ding{72}] Extensive experiments on diverse benchmarks demonstrate that {\ours} achieves comparable performance. Moreover, the explanation evaluation results show that incorporating structural motifs significantly enhances explanation faithfulness, suggesting that token-level patterns alone are insufficient to capture all the features utilized by the detector.
\end{itemize}

%% file: 2_preliminary.tex
\begin{figure*}[t]
\centering
\includegraphics[width=1.0\textwidth]{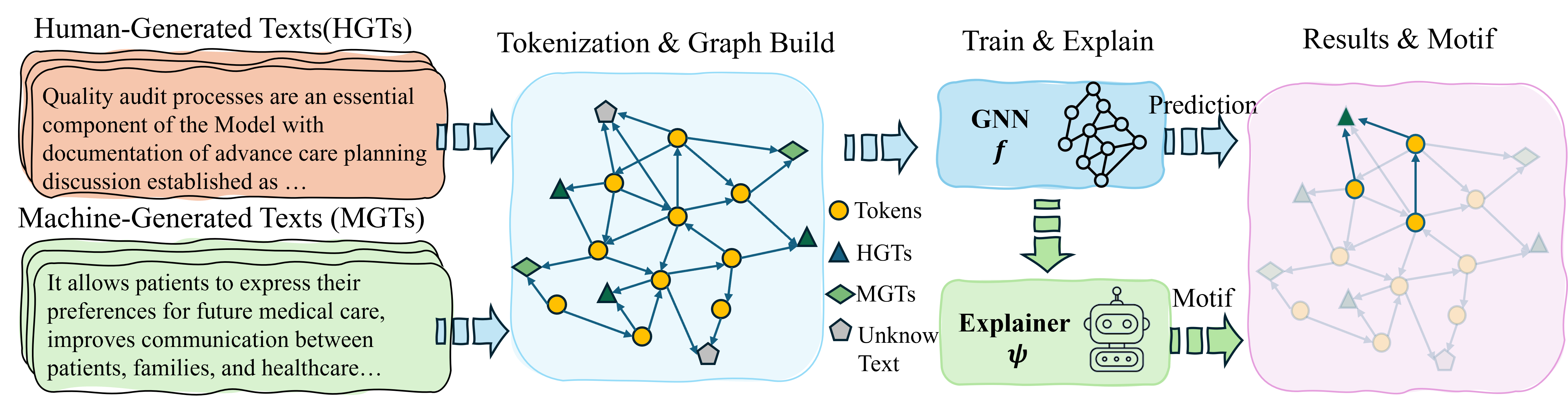}
\vspace{-0.6cm}
\caption{Overall pipeline, including tokenization, graph building, detector training, and motifs extraction.}
\vspace{-0.4cm}
\label{fig:framework}
\end{figure*}

\section{Preliminary}
\textbf{MGT Detection.} The MGT detection problem can be formulated as a classification task. 
Take an example of a binary hypothesis testing task. Given a pair of training sets, 
\begin{align*}
    &\mathcal{T}_{h}=\{\mathbf{S}_{h,i}=(S_{h,i,1},S_{h,i,2},\cdots,S_{h,i,L_i})\}_{i\in |\mathcal{T}_h|},
    \\&\mathcal{T}_m=\{\mathbf{S}_{m,i}=(S_{m,i,1},S_{m,i,2},\cdots,S_{m,i,L'_i})\}_{i\in |\mathcal{T}_m|},
\end{align*}
consisting of human-generated and machine-generated text sequences, respectively, drawn from the distributions\footnote{The length of the observed text sequences is not fixed and can be modeled as a random variable. This variability is implicitly captured in the distributions $P_h$ and $P_m$.} $P_h$ and $P_m$, the objective is to classify a newly observed text sequence $\mathbf{S}_o$ as either human-generated or machine-generated.

A detection mechanism is a function $f:(\mathcal{T}_h,\mathcal{T}_m,\mathbf{S}_o)\mapsto \widehat{Y}$, where $\widehat{Y}\in \{0,1\}$, the index $0$ represents the null hypothesis (human generated) and  $1$ represents the alternative hypothesis (machine generated).  Notably, the function $f$ takes input of $\mathbf{S}_o$ and $\mathcal{T}_h,\mathcal{T}_m$ are the support samples. In training-based methods, $\mathcal{T}_h,\mathcal{T}_m$ are used to train models, while in zero-shot methods, they are used to design the function, such as log-rank information in DetectLLM~\citep{su2023detectllm}.
The detection error is quantified by the risk function $P(f(\mathcal{T}_h,\mathcal{T}_m,\mathbf{S}_o) \neq Y)$, where $Y \in \{0,1\}$ denotes the ground-truth hypothesis label.

\stitle{Probabilistic Graphical Models}. PGM offers an efficient framework for representing probabilistic models, incorporating insightful properties such as conditional independence.
Given a graph $G=\{\mathcal{V},\mathcal{E}\}$, the node set $\mathcal{V}$ corresponds to random variables, and the link set  $\mathcal{E}$ captures probabilistic dependencies between these variables. Given a sequence of three tokens $\mathbf{S}=(s_1,s_2,s_3)$, the joint distribution is $P(s_1,s_2,s_3) = P(s_3|s_1,s_2)P(s_2|s_1)P(s_1)$. It can be represented by a graph with $\mathcal{V}=\{s_1,s_2,s_3\}$ and $\mathcal{E}=\{(s_1,s_2),(s_1,s_3),(s_2,s_3)\}$. Generally, for any sequence of tokens, a PGM can be used to represent the probabilistic dependencies among tokens.

\stitle{Node Classification}.
A graph $G$ consists of a set of nodes $\mathcal{V}=\{v_1,v_2,\cdots,v_n\}$, where $n\in \mathbb{N}$, and
a set of edges $\mathcal{E} \subseteq \mathcal{V} \times \mathcal{V} $. The adjacency matrix $\mA\in \{0,1\}^{n\times n}$ encodes the graph edges, where $A_{i,j} =\mathds{1}((v_i,v_j)\in \mathcal{E})$.
Each node may be associated with a feature vector, collectively represented by the matrix $\mX \in \sR^{n\times d }$, 
where the $i$-th row is the feature vector associated with the $i$-th node, and $d\in \mathbb{N}$ is the dimension. Each node $v$ is related to a label $Y_v \in \mathcal{Y}$, where $\mathcal{Y}$ is the collection of possible labels.
In this work, we reformulate the author detection problem as a node classification task. This reformulation is elaborated on in the subsequent sections. 
The objective in node classification is to train a classifier $f:(\mathbf{G},\mX,v)\mapsto \widehat{Y}_v$, which, given a graph $\mathbf{G}$, node feature matrices $\mX$, and a node index $v$, produces an estimate $\widehat{Y}_v$ of the node label $Y_v$. The accuracy of the classifier is defined as $P_{V,\mathbf{G},\mX,Y_V}(f(V,\mathbf{G},\mX)\neq Y_V)$, where $V$ is uniformly distributed over $\mathcal{V}$, and  $\mathbf{G},\mX,Y_V$ follow a joint distribution $P_{\mathbf{G},\mX,Y_V}$.

\stitle{Post-hoc Explainable Graph Neural Networks}. Given a graph or node classification task, the goal of XGNN is to find an explanation function $\Psi(\cdot)$, which maps the input graph $G$ to a \textit{minimal} and \textit{sufficient} explanation subgraph $G_{exp}$.
Minimality restricts the size of the explanatory subgraph and is enforced by the constraint $|G_{exp}| \leq s \cdot |G|$, where 
$|G|$ denotes the number of edges in $G$ and $s\in [0,1]$ is the size parameter.
Sufficiency is quantified by the KL divergence term $d_{KL}(P_{Y|G,\mX,V}|| P_{Y|G_{exp},\mX,V})$. The explainer is optimally sufficient if it minimizes the KL divergence subject to minimality constraints. Given $s\in [0,1]$, an \textit{optimal} explainer $\Psi^*$ is defined as:
\begin{align}
\label{eq:obj}
\argmin_{\Psi: |G_{exp}|\leq s|G|} d_{KL}(P_{Y|G,\mX,V}|| P_{Y|G_{exp},\mX,V})     
\end{align}

%% file: 2_5_Theortical_analysis.tex
\section{Theoretical Analysis}
\label{sec:theory}
As discussed in the prequel, prior works in MGT detection, such as Fast-DetectGPT \citep{baofast}, have employed sequential data models to design detection mechanisms. 
Drawing inspiration from TextGCN~\cite{yao2019graph}, we formulate the MGT detection problem using a graph-based approach where both tokens and documents are represented as nodes. Building upon this foundation, we demonstrate that GNN-based detectors achieve strictly improved detection accuracy compared to such approaches. This section provides theoretical justifications for this claim. The subsequent sections provide further verification through empirical analysis over several benchmark datasets. 

We formally define a class of baseline \textit{empirical sequential-based} (ESB) detectors that capture the essential characteristics of existing approaches. An ESB detector operates in two steps. First, it uses the human-generated training set $\mathcal{T}_h$ to construct the empirical conditional distribution estimates $\widehat{P}_{h}(s_t|s_{1:t-1})$ for human-generated text sequences,
where $t\in [T]$, and $T$ is a hyperparameter capturing the maximum context length. Similarly, the empirical estimates $\widehat{P}_{m}(s_t|s_{1:t-1})$ are computed based on the machine generated training set $\mathcal{T}_m$. In the second step, the detector uses (a potentially trainable) mapping $g_{s}: ((\widehat{P}_{h}(s_t|s_{1:t-1}),\widehat{P}_{m}(s_t|s_{1:t-1}))_{t\in [T]},\mathbf{S}_o)\mapsto \widehat{Y}$, where $\mathbf{S}_o$ is the to-be-classified sequence. An ESB detector is completely characterized by the mapping $g_s(\cdot)$. 
We denote the collection of ESB detectors by $\mathcal{F}_{\text{ESB}}$.
We introduce the class of PGB MGT detectors. A PGB detector operates on a specially constructed graph with two types of nodes: token nodes and text sequence nodes~\citep{yao2019graph}. Formally, let $\mathcal{V}=\mathcal{S}\cup\mathcal{D}$ denote the complete node set, where
\begin{align*}
& \mathcal{S}= \{s| \exists \mathbf{S}\in \mathcal{T}_h\cup \mathcal{T}_m, i\in [|\mathbf{S}|]: s_i=s\},
\\& \mathcal{D}= \{\mathbf{S}| \mathbf{S}\in \mathcal{T}_h\cup \mathcal{T}_m\cup \{\mathbf{S}_o\}\}.
\end{align*}
Here, $\mathcal{S}$ represents the set of all unique tokens in either human or machine-generated texts, and $\mathcal{D}$ comprises all text sequences from both sources and the to-be-classified text.

The edge structure of the graph captures both token co-occurrences and token-sequence relationships. Two tokens $s_i,s_j\in \mathcal{S}$ are connected if they co-occur in at least $\lambda$ sequences within $\mathcal{T}_h\cup \mathcal{T}_m$, where $\lambda$ is a hyperparameter. Additionally, each token node is connected to sequence nodes containing that token.
Edge weights are defined by two distinct functions. For token-token edges $(s_i,s_j)$, the PGB first computes embedding vectors for each token using
\[e_{\ell}: \mathcal{J}_\ell(s_i)\times (\mathcal{I}_{\ell,j}(s_i))_{j\in \mathcal{J}_\ell(s_i)}\mapsto \mathbf{e}_{\ell,i}, \ell\in \{h,m\},\] 
where for each token $s_i$, the set $\mathcal{J}_\ell(s_i)=\{j| s_i\in \mathbf{S}_{\ell,j}\}$ indexes the sequences containing $s_i$, while $\mathcal{I}_{\ell,j}(s_i)= \{k| S_{\ell,j,k}=s_i\}$ indexes the positions where $s_i$ appears in sequence $\mathbf{S}_{\ell,j}$. The token-token edge weights are then computed as $A_t(\mathbf{e}_{h,i},\mathbf{e}_{m,i},\mathbf{e}_{h,j},\mathbf{e}_{m,j})$, where $\mathbf{e}_{h,i}$ and $\mathbf{e}_{m,i}$ are the embeddings from human and machine-generated texts, respectively. For token-sequence edges $(s,\mathbf{S})$, the weight is simply $A_s(N_{s|\mathbf{S}})$, where $N_{s|\mathbf{S}}$ counts occurrences of token $s$ in sequence $\mathbf{S}$. 
Examples of these edge weight functions $A_t(\cdot)$ and $A_s(\cdot)$ are provided in Section~\ref{section:graph} and used in our empirical evaluations.

Token nodes are initialized with one-hot features and sequence nodes with all-zeros features. The GNN operates by several rounds of message passing among connected nodes. The PGB detector applies $K$ rounds of message passing over the constructed graph, where at each round, node embeddings are updated based on messages received from neighboring nodes. After $K$ iterations, the detector computes the final node embeddings, denoted by $\mathbf{h}^{(K)}$. The classification output is obtained via a function $g_{p}: (\mathbf{h}^{(K)},\mathbf{S}_o) \mapsto \widehat{Y}$ that maps the collection of node embeddings to the binary decision $\widehat{Y}$. 
A PGB detector is completely characterized by the tuple $(K,\lambda, e_{h},e_m,A_t,A_s,g_p)$.
We denote the collection of PGB detectors by $\mathcal{F}_{\text{PGB}}$.

The following theorem shows that the PGB class of detectors strictly subsumes the ESB class in terms of achievable detection accuracy.

\begin{theorem}
    \label{th:1}
    For every ESB detector $f_{\text{ESB}} \in \mathcal{F}_{\text{ESB}}$, there exists a PGB detector $f_{\text{PGB}} \in \mathcal{F}_{\text{PGB}}$ such that the detection accuracy of $f_{\text{PGB}}$ matches that of $f_{\text{ESB}}$, i.e.,  
   \begin{align*}
       P(f_{\text{PGB}}(\mathcal{T}_h, \mathcal{T}_m, \mathbf{S}_o) = Y) =  \\
       P(f_{\text{ESB}}(\mathcal{T}_h, \mathcal{T}_m, \mathbf{S}_o) = Y),
   \end{align*}  
   for all pairs of probability distributions $(P_h,P_m)$. Furthermore, the PGB class of detectors improves upon the ESB class in terms of detection accuracy. That is, for any fixed set of hyperparameters $T,K,\lambda$,    there exists $(P_h,P_m)$ and $f_{\text{PGB}} \in \mathcal{F}_{\text{PGB}}$ for which:
   \begin{align*}
        P(f_{\text{PGB}}(\mathcal{T}_h, \!\mathcal{T}_m,\! \mathbf{S}_o)\! =\! Y) \!\! \geq \\
        \!\!\!\!\!\max_{f_{\text{ESB}} \in \mathcal{F}_{ESB}} \!\!\!\!\!P(f_{\text{ESB}}(\mathcal{T}_h, \!\mathcal{T}_m, \!\mathbf{S}_o) \!=\! Y),
   \end{align*}
\end{theorem}
The proof is provided in \textbf{Appendix \ref{App:th:1}}.

%% file: 3_method.tex
\section{Methodology}
In this section, drawing upon the theoretical foundations of PGM in the prequel, we present the practical implementation of our probabilistic graph-based (PGB) detector framework, {\ours}.  Our implementation encompasses three key components: graph construction based on token co-occurrences, GNN-based authorship detection, and explainable motif extraction. The complete pipeline of {\ours} is illustrated in Figure~\ref{fig:framework}.

\begin{figure}
  \centering
    \includegraphics[width=0.9\linewidth]{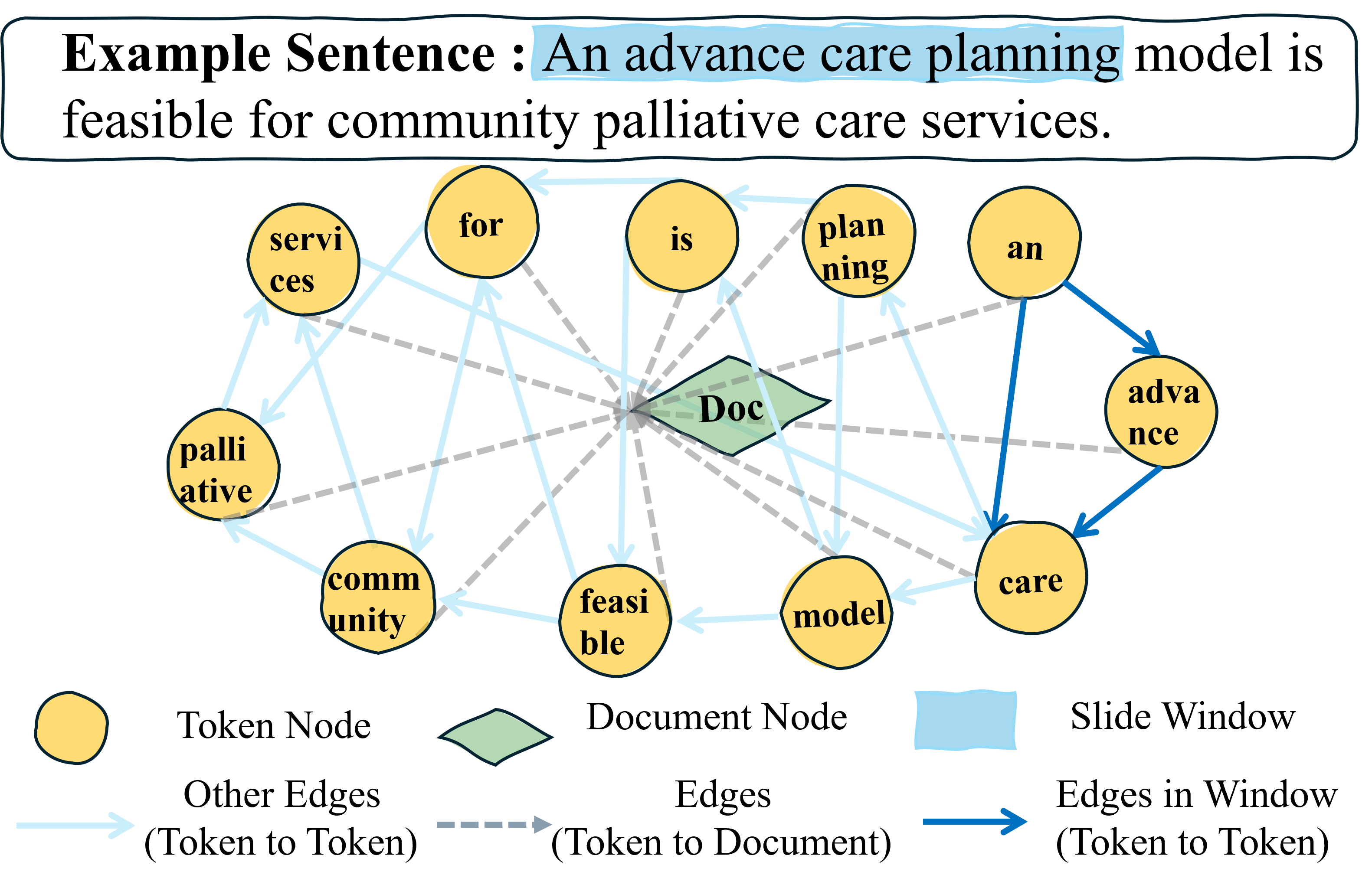}
  \vspace{-0.2cm}
    \caption{An example of graph construction with a fixed sliding window size 3.}
  \label{fig:pomdp}
    \vspace{-0.6cm}
\end{figure}

\subsection{Graph Construction} 
\label{section:graph}
Following our PGB framework, we implement an efficient graph construction method based on word co-occurrence principles.
{ Our pipeline consists of two stages. In the first stage, we capture the dependencies among words/tokens. As shown in Figure~\ref{fig:pomdp}, a word-dependency graph (solid lines) is constructed using a sliding window. In the second stage, we add document nodes and connect them to the corresponding words (dashed lines).
During testing, we similarly add test-document nodes to the existing word-dependency graph.} 
Finally, our graph consists of two types of nodes representing tokens and documents, corresponding to the node sets $\mathcal{S}$ and $\mathcal{D}$. As specified in our framework, tokens are initialized with one-hot features and documents with zero vectors.

To construct edges that capture textual relationships, we consider both document-token connections and token co-occurrences. The adjacency matrix $A$ is defined as:
\begin{align*}
\vspace{-0.2cm}
\label{eq:attr}
A_{ij} = \left \{ \begin{array}{cc}
1 & i,j~ \text{are token}, \text{PMI}(i,j)>0 \\
1 &  j~ \text{is document}, i~ \text{is token in}~ j\\
1 &  i=j\\
0 &  \text{otherwise}\\
\end{array} \right. ,
\end{align*}
where $\text{PMI}(i,j) = \text{log}\frac{p(i,j)}{p(i)p(j)}$, point-wise mutual information, is used to determine if these token co-occurrences are significant. 

To strictly prevent data leakage, all statistical probabilities are pre-computed exclusively on the training corpus. 
$p(i)$ represents the probability of the $i$-th token within a fixed-length sliding window across the training set, and $p(i,j)$ denotes the co-occurrence probability of tokens $i$ and $j$ within the same training bounds. As discussed in Section~\ref{sec:theory}, in the most general sense, the edge weights may be continuous-valued and generated using a learnable function. However, our experimental evaluation shows that the above binary-valued edge weights are sufficient for reliable detection. During inference, Out-Of-Vocabulary~(OOV) is a potential problem. To preserve the co-occurrence graph topology without recomputing global statistics, we adopt a semantic fallback mechanism. Instead of discarding OOV tokens, we project them into the known vocabulary space using semantic embeddings and replace each token with its nearest semantic neighbor from the training set.

\subsection{GNN Detection}
Having constructed the graph structure, we implement the detection mechanism outlined in our framework through a GNN architecture. For a given text sequence $\mathbf{S}_o$, our goal is to learn a function $f$ that determines whether the text is machine-generated or human-authored. This corresponds to the PGB detector operating over $K$ message passing rounds.
\noindent Each GNN layer implements one round of message passing, with the update rule:
\begin{align*}
a_v^{(l)} &= \text{AGG}^{(l)}\left({h_u^{(l-1)}: u\in \mathcal{N}(v)}\right), \\
h_v^{(l)}  &= \text{COMBINE}^{(l)}\left(h_v^{(l-1)},a_v^{(l)}\right),
\end{align*}

{where $a_v^{(l)}$ represents the aggregated message at $l$-th layer, $h_v^{(l)}$ is the node feature, $\mathcal{N}(v)$ denotes the neighbors of node $v$, and $\text{AGG}(\cdot)$ and $\text{COMBINE}(\cdot)$  are the regular aggregation and combination functions in GNNs, following the definition from previous work~\citep{xu2018how}. }
After $K$ layers, we obtain the final node embeddings $\mH$.
For classification, we apply a softmax function to the final embeddings to obtain prediction probabilities $\mZ = \text{softmax}(\mH)$. The model is trained by minimizing the cross-entropy loss over labeled document nodes:
\begin{equation*}
\mathcal{L} = - \sum_{d \in \mathcal{Y}_D}\sum_{\ell\in \{h,m\}} Y_{d\ell}\ln Z_{d\ell},
\end{equation*}
where $\mathcal{Y}_D$ represents the set of document nodes in the training set and $Y_{d\ell}$ is the ground-truth label.
While our goal focuses on binary classification (human-authored vs. machine-generated) in this paper, the framework naturally extends to scenarios with multiple classes, such as texts generated by different language models.

\subsection{Explainable Motifs Extraction}
Beyond detection accuracy, our core strength lies in the ability to provide unparalleled interpretable insights through the extraction of distinguishing motifs. 
Drawing inspiration from graph analysis techniques~\citep{luo2020parameterized}, we transform the interpretability challenge into a subgraph identification problem, where meaningful token dependencies in our constructed graph serve as distinguishing motifs. These motifs capture characteristic patterns of word usage and dependencies that differentiate between human and machine-generated content~\citep{kim2024threads}, providing concrete insights beyond simple token-level statistics.

Specifically, we formulate a practical optimization objective using cross-entropy loss and explicit size constraints. The objective function balances the prediction accuracy of the explanation subgraph against its complexity:
\begin{equation*}
\Psi^*(\cdot)=\argmin_{\Psi: G\mapsto G_{exp}} \text{CE}(Y;f(G_{exp})) + \lambda |G_{exp}|
\end{equation*}
where $\text{CE}(Y;f(G_{exp}))$ measures how well the explainer preserves the model's prediction capability, $|G_{exp}|$ denotes the size of the explanation subgraph, and $\lambda$ controls the trade-off between explanation fidelity and complexity. This formulation is an approximation of the theoretical requirements from Equation~\ref{eq:obj}, where the cross-entropy term ensures sufficiency and the size penalty enforces minimality. The optimization is performed through gradient descent, with the edge weights of $G_{exp}$ being learned and then discretized through thresholding.

%% file: 4_relatedwork.tex
\section{Related Work}
\stitle{AI-generated text Detection}. Detecting machine-generated texts approaches can be categorized into three main categories. The first category focuses on watermarking LLM-generated content~\citep{chang2024postmark,ajith2024downstream,yang2023watermarking,wuresilient,molenda2024waterjudge}. Most watermarking methods operate in a white-box setting, where researchers can modify the decoding process or token distribution directly~\citep{ajith2024downstream,wuresilient,molenda2024waterjudge}. 
The black-box setting can be achieved by implementing post-processing modules to embed watermarks~\citep{chang2024postmark,yang2023watermarking}. 
The second category encompasses training-based detection methods that leverage trained neural networks~\citep{guo2024detective,solaiman2019release,zhangdetecting,kim2024threads,sotofew}. OpenAI developed GPT-2 detectors using RoBERTa~\citep{liu2019roberta} as their foundation model~\citep{solaiman2019release}. 
Additionally, researchers have explored fine-tuning language models specifically for detection purposes~\citep{li2023deepfake,Koike:OUTFOX:2024,guo-etal-2023-hc3,zhang2024detecting}. 
The third category consists of zero-shot detection methods~\citep{nguyen2024simllm,zeng2024dlad,yangdna,tian2024detecting,ma2024zero}, which utilize existing tools like LLMs without additional training. For example, SimLLM~\citep{nguyen2024simllm} generates comparative text samples to identify machine-generated content through similarity analysis. R-Detect~\citep{song2025deep} suggests a non-parametric kernel relative test to check if a text's distribution is closer to HGT than MGT.

\stitle{Explainable LLMs \& GNNs}. Large language models often function as black-box systems, presenting inherent risks for downstream applications~\citep{zhao2024explainability}. To address this limitation, researchers have developed various explanation methods~\citep{wu2020perturbed,li2016understanding,enguehard2023sequential,chen2023rev}, which can be divided into local and global approaches. Local explanation methods aim to illuminate how an LLM arrives at predictions for specific inputs~\citep{wu2020perturbed,li2016understanding,chen2023rev}. For example, the leave-one-out technique represents a fundamental approach to measuring input feature importance~\citep{wu2020perturbed,li2016understanding}. 
Global explanation methods focus on understanding how specific model components operate, including hidden layers and language model mechanisms. 
For instance, researchers have tracked attention layers to extract semantic information~\citep{wu2020perturbed}. 
SASC~\citep{singh2023explaining} employs pre-trained models to generate explanations for various LLM components. 

Various approaches have emerged for extracting subgraph explanations using GNNs~\citep{yuan2022explainability,lin2021generative,fang2023cooperative,xie2022task,chengenerating}. These methods can be categorized into several groups. Gradient-based traditional approaches, including SA~\citep{sa_baldassarre2019explainability} and Grad-CAM~\citep{GradCAM_pope2019explainability}, leverage gradient information to derive explanations.
Model-agnostic techniques encompass three main categories. First, perturbation-based methods such as GNNExplainer~\citep{ying2019gnnexplainer}, PGExplainer~\citep{luo2020parameterized}, and ReFine~\citep{wang2021towards} identify important features and subgraph structures through systematic perturbations. Second, surrogate methods~\citep{vu2020pgm,duval2021graphsvx} approximate local predictions using surrogate models to generate explanations. Third, generation-based approaches~\citep{yuan2020xgnn,shan2021reinforcement,wang2022gnninterpreter} employ generative models to produce both instance-level and global-level explanations.

%% file: 5_Experiments.tex
\section{Experiments}
We conduct extensive experiments to evaluate ours. 
Rather than solely chasing cross-domain accuracy at the expense of transparency, we evaluate $\ours$ across two primary dimensions, including fundamental detection capability within specific domains, and the ability to extract highly faithful, explainable structural motifs. Due to space limitations, detailed detection results, ablation studies, time complexity, implementation details, detailed explanation evaluation results, and motif statistical analysis are provided in Appendix~\ref{extend:detailedexp}.

\subsection{Setups}
\stitle{Datasets}. 
Following established benchmarks in MGT detection~\cite{yangdna,zeng2024dlad}, we evaluate {\ours} on six comprehensive datasets: HC3~\cite{guo-etal-2023-hc3}, M4~\cite{wang-etal-2024-m4}, RAID~\cite{dugan-etal-2024-raid}, Yelp~\cite{mao2024raidar}, Creative, and Essay~\cite{verma2023ghostbuster,guo2024biscope}. 
We select diverse domains, including open-qa, medicine, finance, reddit, arxiv, and IMDB reviews. We evaluate across a wide spectrum of LLMs, including ChatGPT, DaVinci, {Cohere}, Dolly, BloomZ, Llama2, GPT-4, Mistral, Claude-3 families, and Gemini-1.0-Pro. Dataset details and abbreviation for LLMs are available in Appendix~\ref{app:dataset}.

\stitle{Baselines}. To rigorously evaluate both detection and explainability, we compare {\ours} against a diverse suite of baselines.
For detection baselines, we consider state-of-the-art training-based and zero-shot methods, such as DetectLLM~\cite{su2023detectllm}, DeTeCtive~\cite{guo2024detective}, and DNAGPT~\cite{yangdna}. Both training-based and zero-shot methods use pre-trained models for a unified comparison protocol. For explanation baselines, we compare against feature attribution methods (LIME~\cite{ribeiro2016should}, SHAP~\cite{NIPS2017_8a20a862}), statistical visualizers (GLTR~\cite{gehrmann-etal-2019-gltr}), LLM-based explainers (GPT-4o), and a Random Motif control. For all explanation baselines, we utilize RoBERTa-QA as the competitive base model to be explained. Detailed baseline information is provided in Appendix~\ref{app:baseline}.

\input{experiments/nips/chatgpt-acc}

\input{experiments/nips/llms-acc}

\subsection{Detection Performance Comparison}
\label{exp:detection}
 
We compare {\ours} against 13 baselines, including training-based and zero-shot methods, to comprehensively evaluate detection performance. We report both accuracy(ACC) and area under the receiver operating characteristic(AUC) results. For the in-domain setting, we train and test our method on the same domain. In Table~\ref{tab:humanchatgpt_avg}, we report the average results of ChatGPT-based text detection on three datasets. {\ours} achieves the best performance under ACC and AUC metrics. In Table~\ref{tab:llms-acc}, we study the performance across various LLMs. As the results show, the performance is aligned with Table~\ref{tab:humanchatgpt_avg}. Under the ACC metric, {\ours} is the best performance on average, demonstrating the ability for MGT detection. The detailed results are available in {Appendix~\ref{extend:detailedexp}}. Due to the limitation of pages, we provide more experiments about cross-domain evaluation, and statistical significance analysis in {Appendix~\ref{extend:detections}}.

\input{experiments/figure_baselines_comp_avg}

\subsection{Explanation Evaluation}
\label{exp:motifs}
\stitle{Quantitative Analysis}. 
The primary contribution of {\ours} lies in its transparent, white-box nature. However, due to the lack of ground-truth, evaluating explanation quality remains challenging. Therefore, following prior work~\cite{NEURIPS2019_fe4b8556, zhengtowards, zheng2025ffidelity}, we adopt the MORF~(Most Relevant First) and LERF~(Least Relevant First) protocols, which are widely used evaluation methods in explainable AI (XAI). These protocols assess explanation faithfulness by measuring how model predictions change when the most or least relevant input attributions are sequentially removed according to the generated explanations. In our graph setting, we carefully remove topological edges while preserving nodes, ensuring that the core semantic structure remains intact while isolating the impact of structural connections.

As shown in Figure~\ref{fig:explainable_baseline_avg}, {\ours} significantly outperforms all explanation baselines on the HC3 dataset across faithfulness evaluations. Under the MORF protocol, removing the top 20\% most important edges identified by our method leads to the average detection accuracy dropping by nearly 15\%. In contrast, removing features identified by LIME, SHAP, or GPT-4o leads to an accuracy decline of less than 10\%. Conversely, under the LERF protocol, {\ours} maintains noticeably higher accuracy even after removing 30\% of the least important features, outperforming all baselines and confirming that our framework successfully separates true predictive signals from structural noise.

From the behavior of the baselines, we observe that token-level saliency maps do not provide sufficient evidence to reliably explain the model's decisions, suggesting that sequence-based detectors capture more than localized token patterns. In contrast, our token co-occurrence-based framework explicitly models higher-order dependency structures, enabling more faithful explanations. These results quantitatively demonstrate that the decision boundary between MGTs and HGTs is governed by high-order structural relationships rather than simple token-frequency patterns. 
To further validate the effectiveness of our approach, we provide additional comparisons on the M4 and RAID datasets, together with random baselines, in Appendix~\ref{extend:motifs_eval}.

\begin{figure}[t]
  \centering
  \begin{subfigure}[h]{0.23\textwidth}
        \includegraphics[width=\textwidth, trim=50 30 25 25, clip]{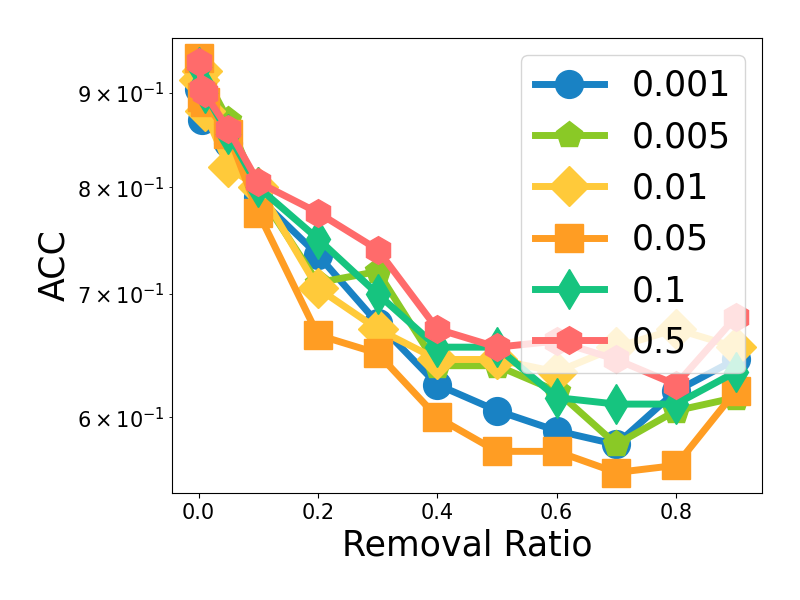}
        \caption{wiki-csai}
        \label{fig:lambda_hc3_csai}
  \end{subfigure}
  \begin{subfigure}[h]{0.23\textwidth}
        \includegraphics[width=\textwidth, trim=50 30 25 25, clip]{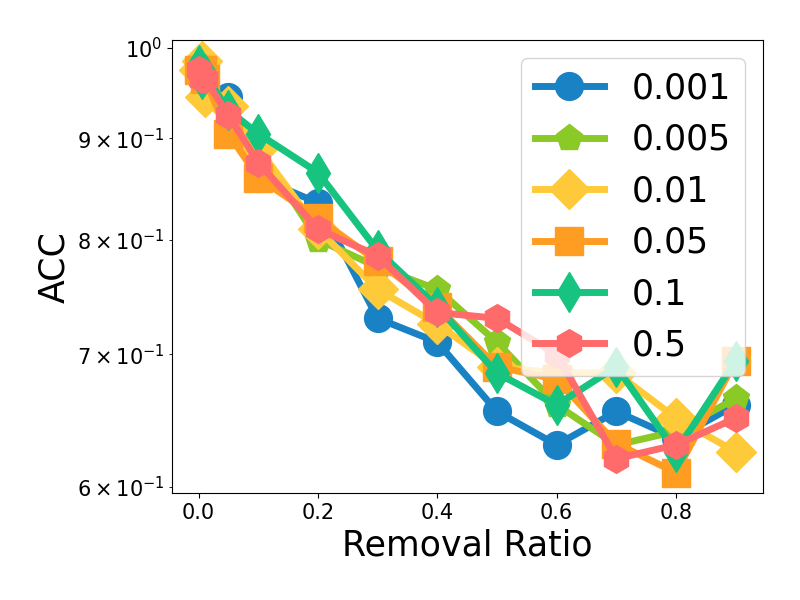}
        \caption{finance}
        \label{fig:lambda_hc3_finance}
  \end{subfigure}
\vspace{-0.2cm}
\caption{The hyperparameter analysis of $\lambda$ with MORF protocol for {\ours}.}
\label{fig:lambda_ablation}
\vspace{-0.8cm}
\end{figure}

\stitle{Hyperparameter Sensitivity Analysis}. 
To analyze the sensitivity of motif extraction, we conduct an ablation study on the explainer hyperparameter $\lambda$ using the HC3 dataset, and evaluate the resulting explanations under the MORF protocol. As shown in Figure~\ref{fig:lambda_ablation}, our method demonstrates strong robustness across a wide range of $\lambda$ values, indicating that the discriminative structural patterns learned by the graph-based detector can be consistently extracted without requiring careful hyperparameter tuning. 
Notably, performance remains stable even under relatively large perturbations of $\lambda$, suggesting that the optimization landscape for motif extraction is smooth and does not rely on finely balanced parameter configurations. This result further suggests that the detected motifs are stable and intrinsic to the model's decision process, rather than artifacts caused by a specific parameter setting. Overall, the observed insensitivity to $\lambda$ strengthens the claim that the learned structural explanations reflect meaningful and persistent patterns in the underlying graph representations.

Overall, detectors make predictions by combining linguistic and structural features such as word distributions, token co-occurrence patterns, and higher-order contextual dependencies. 
For example, in watermarking-based detection methods~\cite{li2025statistical}, the probability of generating certain words is manipulated through predefined green and red lists, allowing machine-generated text to be identified through deviations in word frequencies. Importantly, these signals go beyond simple statistics and capture meaningful distinctions between text types, revealing that different language models leave \textit{distinct and visible fingerprints}.

%% file: experiments/nips/chatgpt-acc.tex
\begin{table}[t]
\centering
\caption{Detection comparisons on HGTs and ChatGPT-generated texts. 
The best and second-best results are shown in bold font and underlined.  F-DetectGPT and R-QA are short for Fast-DetectGPT and RoBERTa-QA. } 
\vspace{-0.2cm}
\scalebox{0.73}{
    \begin{tabular}{lc@{\hspace{0.2cm}}c@{\hspace{0.2cm}}c@{\hspace{0.1cm}}c@{\hspace{0.1cm}}c@{\hspace{0.2cm}}c@{\hspace{0.2cm}}c@{\hspace{0.1cm}}c}
        \toprule
        & \multicolumn{4}{c}{ACC} & \multicolumn{4}{c}{AUC} \\
        \cmidrule(r){2-5}\cmidrule(r){6-9}
         Method & HC3 & M4 & RAID & Avg.& HC3 & M4 & RAID & Avg. \\
     \midrule
     
    \rowcolor{gray!20} \multicolumn{9}{c}{\textbf{Zero-Shot Methods}} \\
    NPR          & 0.83	&	0.71	&	0.79	&	0.78 &	\textbf{1.00}	&	0.93	&	\underline{0.97}	&	0.97 \\
    LRR          & 0.96	&	0.86	&	0.87	&	0.90 &	\textbf{1.00}	&	0.98	&	0.96	&	0.98 \\
    
    Rank         & 0.53	&	0.58	&	0.56	&	0.56 &	0.89	&	0.95	&	0.91	&	0.92 \\
    Entropy      & 0.77	&	0.73	&	0.66	&	0.72 &	\underline{0.95}	&	0.79	&	0.89	&	0.88 \\
    LogRank      & 0.70	&	0.87	&	0.84	&	0.81 &	\textbf{1.00}	&	0.94	&	\underline{0.97}	&	0.97 \\
    Likelihood   & 0.75	&	0.88	&	0.85	&	0.83 &	\textbf{1.00}	&	0.90	&	0.98	&	0.96 \\
    
    Glimpse      &\underline{0.98}   &   0.94    &   0.91    &   0.94 &  \textbf{1.00}    & 0.98     &  0.96    & 0.98  \\
    Binoculars   &\underline{0.98}   &   0.94    &   \textbf{0.99}    &  \underline{0.97} &  \textbf{1.00}    &  0.98    &  \textbf{1.00}   &  \underline{0.99}  \\
    DNAGPT       & 0.73	&	0.68	&	0.72	&	0.71 &	0.88	&	0.86	&	0.93	&	0.89 \\ 
    DetectGPT    & 0.63	&	0.61	&	0.62	&	0.62 &	0.56	&	0.63	&	0.78	&	0.66 \\
    F-DetectGPT      & 0.97	&	\underline{0.96}	&	\underline{0.97}	&	\underline{0.97} &	\textbf{1.00}	&	\underline{0.99}	&	\textbf{1.00}	&	\underline{0.99} \\
    \midrule
    \rowcolor{gray!20} \multicolumn{9}{c}{\textbf{Training-Based Methods}} \\ 
    R-QA   & \textbf{1.00}	&	0.95	&	0.80	&	0.91 &	\textbf{1.00}	&	\underline{0.99}	&	0.96	&	0.98 \\
    RADAR        & 0.66	&	0.76	&	0.77	&	0.73 &	0.52	&	0.83	&	0.95	&	0.76 \\
    GPTZero      & 0.77	&	0.75	&	0.68	&	0.73 &	0.77	&	0.75	&	0.68	&	0.73 \\ 
    DeTeCtive    & 0.92	&	0.93	&	0.96	&	0.93 &	0.93	&	0.94	&	0.98	&	0.95 \\
    \midrule
    {\ours}     & {0.97}	&	\textbf{0.98}	&	\textbf{0.99}	&	\textbf{0.98} &	\textbf{1.00}	&	\textbf{1.00}	&	\textbf{1.00}	&	\textbf{1.00} \\
    \bottomrule
    \end{tabular}
    }   
\label{tab:humanchatgpt_avg}
\vspace{-0.4cm}
\end{table}

%% file: experiments/nips/llms-acc.tex
\begin{table*}[t]
\centering
\caption{Detection performance comparisons on HGT and MGT based on ACC.
The best and second-best results are shown in bold and underlined, respectively.
YEC represents the combination of the Yelp, Essay, and Creative datasets. F-DetectGPT and R-QA are short for Fast-DetectGPT and RoBERTa-QA.}
\vspace{-0.2cm}
\setlength\tabcolsep{3.3pt}
\scalebox{0.84}{
    \begin{tabular}{lc@{\hspace{0.4cm}}c@{\hspace{0.4cm}}c@{\hspace{0.4cm}}c@{\hspace{0.4cm}}c@{\hspace{0.4cm}}c@{\hspace{0.4cm}}c@{\hspace{0.4cm}}c@{\hspace{0.4cm}}c@{\hspace{0.4cm}}c@{\hspace{0.4cm}}c@{\hspace{0.4cm}}c}
    
        \toprule
        & \multicolumn{4}{c}{M4} & \multicolumn{4}{c}{RAID} & \multicolumn{3}{c}{YEC} \\
        \cmidrule(r){2-5}\cmidrule(r){6-9}\cmidrule(r){10-12}
         Method & DaV. & Coh. & Dol. & Blo. & Lla. & GT4 & MPT & Mis. & Son. & Opu. & Gem. & Avg. \\
     \midrule
     \rowcolor{gray!20} \multicolumn{13}{c}{\textbf{Zero-Shot Methods}} \\
    NPR~\cite{su2023detectllm}         & 0.63	&	0.67	&	0.55	&	0.59	&	0.79	&	0.66	&	0.54	&	0.65	&	0.70	&	0.63	&	0.54	&	0.63
\\
    LRR~\cite{su2023detectllm}         & 0.77	&	0.75	&	0.74	&	0.77	&	0.87	&	0.71	&	0.55	&	0.71	&	0.74	&	0.72	&	0.53	&	0.72
\\
    Rank~\cite{gehrmann2019gltr}        & 0.51	&	0.54	&	0.53	&	0.53	&	0.53	&	0.53	&	0.51	&	0.52	&	0.52	&	0.52	&	0.52	&	0.52
\\
    Entropy~\cite{gehrmann2019gltr}     & 0.62	&	0.61	&	0.53	&	0.53	&	0.62	&	0.62	&	0.52	&	0.63	&	0.72	&	0.74	&	0.64	&	0.62
\\
    LogRank~\cite{ippolito2019automatic}     & 0.67	&	0.88	&	0.72	&	0.62	&	0.80	&	0.74	&	0.46	&	0.66	&	0.76	&	0.79	&	0.72	&	0.71
\\
    Likelihood~\cite{solaiman2019release}  & 0.69	&	0.87	&	0.66	&	0.54	&	0.79	&	0.75	&	0.50	&	0.65	&	0.80	&	0.83	&	0.74	&	0.71
\\
    Glimpse~\cite{bao2025glimpse}     & 0.74  & 0.94              & 0.69 & 0.61 & 0.88          & 0.77 &  0.68 & 0.77 & 0.85 & 0.85  & 0.76  &  0.69\\
    Binoculars~\cite{ma2024zero}   & 0.83  & \underline{0.97}  & 0.90 & 0.66 & \textbf{0.98} & 0.92 & 0.58 & 0.71 & 0.88 & \underline{0.91} & 0.81 & 0.83 \\
    DNAGPT~\cite{yangdna}      & 0.53	&	0.74	&	0.53	&	0.50	&	0.68	&	0.66	&	0.39	&	0.54	&	0.62	&	0.64	&	0.65	&	0.59
\\ 
    DetectGPT~\cite{mitchell2023detectgpt}   & 0.48	&	0.57	&	0.48	&	0.59	&	0.67	&	0.59	&	0.46	&	0.53	&	0.62	&	0.58	&	0.57	&	0.56
\\
    F-DetectGPT~\cite{baofast}     & 0.81	&	\textbf{0.98}	&	\underline{0.90}	&	0.54	&	0.94	&	0.85	&	0.48	&	0.64	&	0.85	&	0.88	&	0.76	&	0.78
\\

    \rowcolor{gray!20} \multicolumn{13}{c}{\textbf{Training-Based Methods}} \\
    R-QA~\cite{guo-etal-2023-hc3}  & 0.83	&	0.94	&	0.74	&	0.51	&	0.77	&	0.70	&	0.56	&	0.56	&	0.79	&	0.87	&	0.80	&	0.73
\\
    RADAR~\cite{hu2023radar}       & 0.76	&	0.77	&	0.65	&	0.63	&	0.68	&	0.69	&	0.64	&	0.72	&	0.80	&	0.83	&	0.78	&	0.72
\\
    GPTZero~\cite{gptzero}     & 0.74	&	0.80	&	0.61	&	0.53	&	0.65	&	0.60	&	0.54	&	0.55	&	0.69	&	0.71	&	0.54	&	0.63
\\   
    DeTeCtive~\cite{guo2024detective}   & \underline{0.90}	&	0.85	&	\underline{0.90}	&	\underline{0.92}	&	\underline{0.96}	&	\underline{0.97}	&	\textbf{0.92}	&	\underline{0.88}	&	\underline{0.94}	&	\underline{0.91}	&	\underline{0.86}	&	\underline{0.91}
\\
    \midrule
    {\ours}     &  \textbf{0.95}	&	\underline{0.97}	&	\textbf{0.91}	&	\textbf{0.98}	&	\textbf{0.98}	&	\textbf{1.00}	&	\underline{0.90}	&	\textbf{0.91}	&	\textbf{0.99}	&	\textbf{0.99}	&	\textbf{0.91}	&	\textbf{0.95}
\\
     
    \bottomrule
    \end{tabular}
    }   
\label{tab:llms-acc}
\vspace{-0.2cm}
\end{table*}

%% file: experiments/figure_baselines_comp_avg.tex
\begin{figure*}[t]
  \centering
    \begin{subfigure}[h]{0.28\textwidth}
        \includegraphics[width=\textwidth, trim=25 30 25 25, clip]{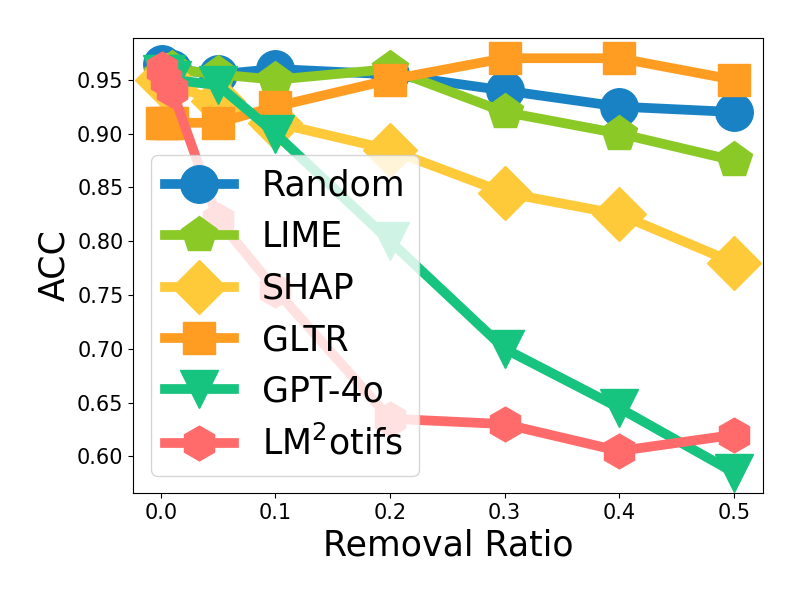}
        \caption{MORF on open-qa}
        \label{fig:baseline_openqa_morf}
  \end{subfigure}
  \begin{subfigure}[h]{0.28\textwidth}
        \includegraphics[width=\textwidth, trim=25 30 25 25, clip]{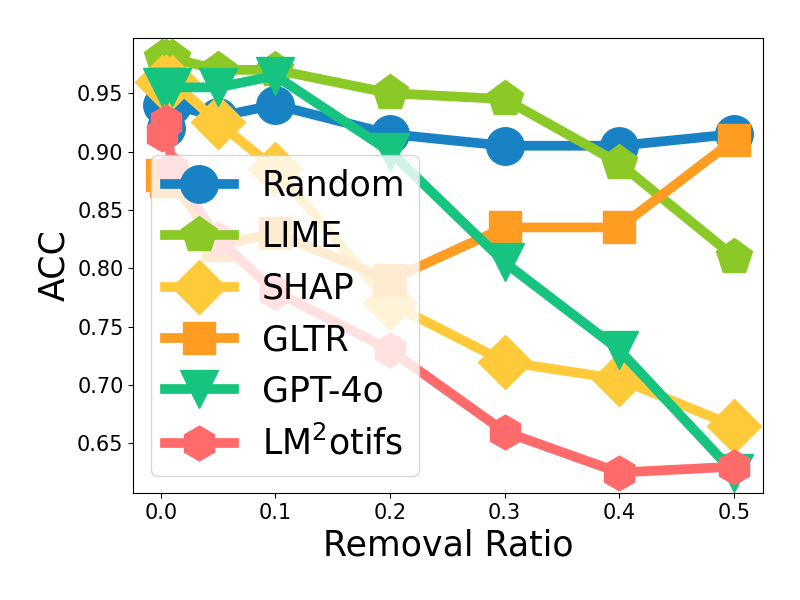}
        \caption{MORF on wiki-csai}
        \label{fig:baseline_wiki_morf}
  \end{subfigure}
  \begin{subfigure}[h]{0.28\textwidth}
        \includegraphics[width=\textwidth, trim=25 30 25 25, clip]{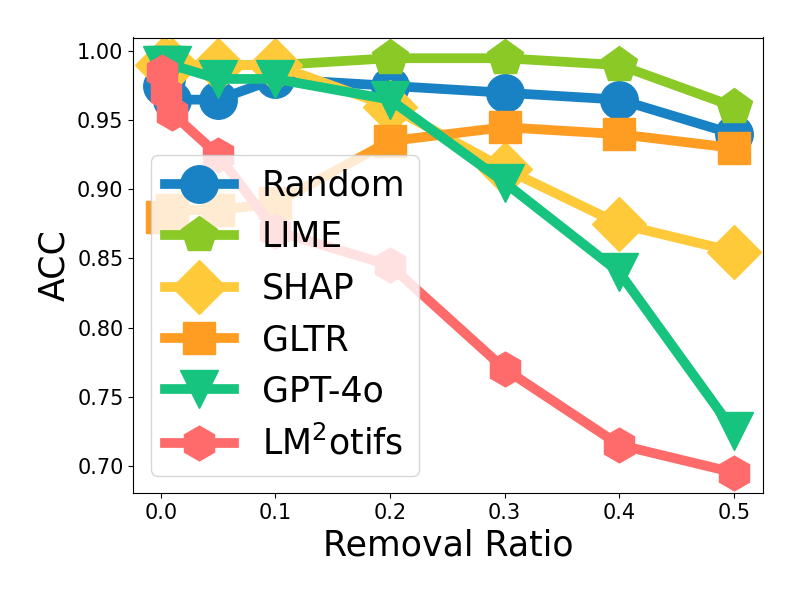}
        \caption{MORF on finance}
        \label{fig:baseline_finance_morf}
  \end{subfigure}
  
  \begin{subfigure}[h]{0.28\textwidth}
        \includegraphics[width=\textwidth, trim=25 25 25 25, clip]{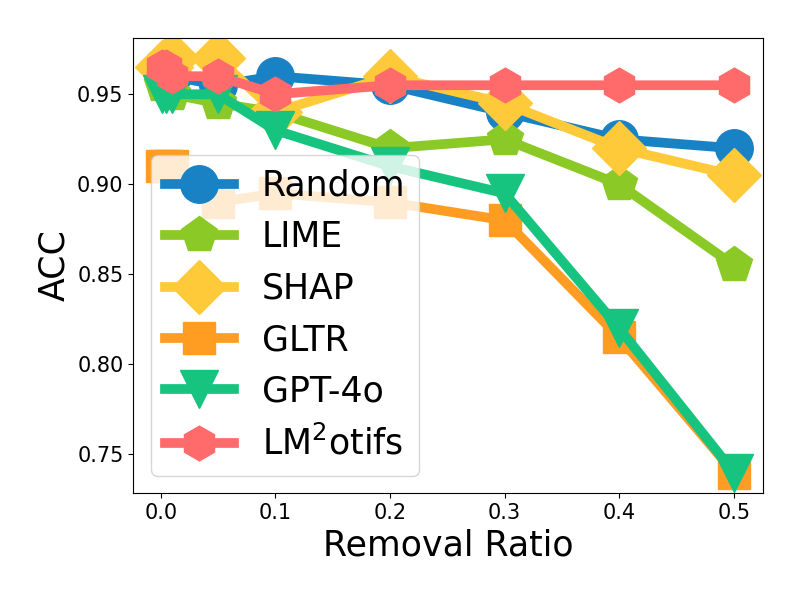}
        \caption{LERF on open-qa}
        \label{fig:baseline_avg_lerf}
  \end{subfigure}
  \begin{subfigure}[h]{0.28\textwidth}
        \includegraphics[width=\textwidth, trim=25 25 25 25, clip]{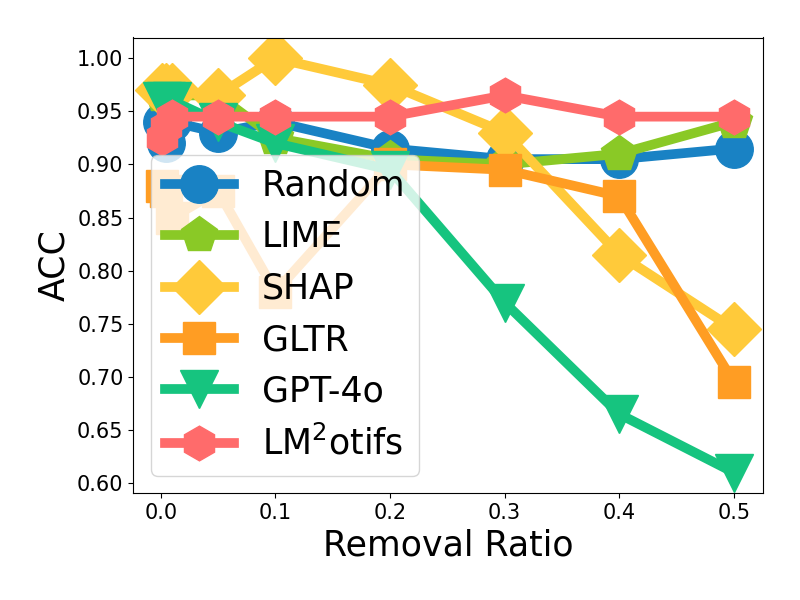}
        \caption{LERF on wiki-csai}
        \label{fig:baseline_wiki_lerf}
  \end{subfigure}
  \begin{subfigure}[h]{0.28\textwidth}
        \includegraphics[width=\textwidth, trim=25 25 25 25, clip]{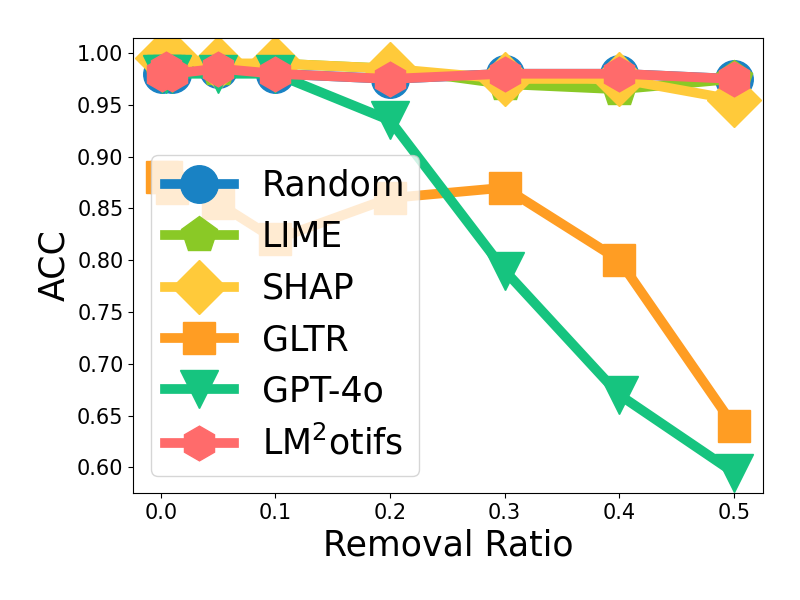}
        \caption{LERF on finance}
        \label{fig:baseline_finance_lerf}
  \end{subfigure}
  
\vspace{-0.2cm}
\caption{Comparison of explanation evaluation results between {\ours} and adapted baseline methods across three domains in the HC3 dataset, highlighting the effectiveness and consistency of our approach.}
\label{fig:explainable_baseline_avg}
\vspace{-0.5cm}
\end{figure*}

%% file: 6_conclusion.tex
\section{Conclusion}
We presented {\ours}, a principled framework for explainable MGT detection. By shifting from linear sequences to graph-structured manifolds, our approach captures the high-order structural dependencies of LLMs that traditional detectors overlook. Grounded in probabilistic graphical models, {\ours} provides a theoretical basis for identifying machine-specific \textit{fingerprints} through interpretable motifs.
Experimental results across diverse benchmarks show that {\ours} achieves state-of-the-art performance while ensuring significantly higher explanation faithfulness than baselines. By providing verifiable structural evidence alongside binary verdicts, {\ours} establishes a transparent and robust foundation for authorship authentication in the era of generative AI.

\section*{Limitations}
Our experimental results demonstrate our claims, but the impact of different GNN architectures or hyperparameter settings remains underexplored. Additionally, the quality of the explainable motifs is highly dependent on the quality of the underlying graph representation, which in turn requires a sufficient number of training samples to construct effectively. Currently, while {\ours} achieves state-of-the-art accuracy and unprecedented structural transparency in in-domain settings, its cross-domain generalization is constrained by the limited scale of the constructed graphs. Future work could investigate scaling up the graph construction across larger, more diverse datasets to enhance cross-domain generalization and explore a wider range of GNN backbones.

%% file: 7_appendix.tex
\section{Proof of Theorem \ref{th:1}}
\label{App:th:1}
We first prove that the ensemble of PGB detectors is at least as accurate as the ensemble of ESB detectors. To this end, let us recall that an ESB detector is completely characterized by  the mapping $g_{s}$ and the PGB detector by $(K,\lambda,e_{h},e_m,A_t,A_s,g_p)$.
Let us consider an arbitrary ESB detector by fixing the function $g_{s}(\cdot)$. The ESB detector computes $\widehat{P}_\ell(s_t|s_{1:t-1}),  \ell\in\{h,m\},t\in [T]$ empirically and uses $g_s((\widehat{P}_\ell(s_t|s_{1:t-1}))_{\ell\in\{h,m\},t\in [T]},\mathbf{S}_o)$ for detection.
On the other hand, the PGB uses the  embedding functions $e_\ell,A_t,A_s$ to compute the final node embeddings $\mathbf{h}^{(K)}$ and the mapping $g_{p}(\mathbf{h}^{(K)},\mathbf{S}_o)$ for detection. We take $K=T$ and $\lambda=1$. 
Then, to prove that there exists a PGB which matches the ESB in terms of detection accuracy, 
it suffices to show that there exist choices of embedding functions $e_\ell,A_t,A_s$, such that the empirical estimate $\widehat{P}_\ell(s_t|s_{1:t-1}), \ell\in \{h,m\},t\in [T]$ can be written as a function of the final node embeddings $\mathbf{h}^{(T)}$, i.e., there exists $r(\cdot)$ such that $r(\mathbf{h}^{(T)})=(\widehat{P}_\ell(s_t|s_{1:t-1}))_{\ell\in \{h,m\}, t\in [T]}$.
Then, the proof follows by taking $g_{p}(r(\mathbf{h}^{(T)}),\mathbf{S}_o)=g_{s}((\widehat{P}_\ell(s_t|s_{1:t-1}))_{\ell\in \{h,m\}, t\in [T]},\mathbf{S}_o)$, so that 
\begin{align*}
    P(f_{\text{PGB}}(\mathcal{T}_h, \mathcal{T}_m, \mathbf{S}_o) =f_{\text{ESB}}(\mathcal{T}_h, \mathcal{T}_m, \mathbf{S}_o) )=1.
\end{align*}

To this end, we take $e_{\ell}$ as the identity function and $A_t$ as a one-to-one parametrization function, so that for each token node $s_i$, the collection $\mathcal{J}_\ell(s_i)\times (\mathcal{I}_{\ell,j}(s_i))_{j\in \mathcal{J}_\ell(s_i)}$ can be computed from its connected edge weights, where $\mathcal{J}_{\ell}(s_i)$ is the training sequence indices in which the token is present and 
$\mathcal{I}_{\ell,j}(s_i)$ is the collection of indices in  the sequence $\mathbf{S}_{\ell,j}, j\in \mathcal{J}_{\ell}$ whose value is equal to $s_i$. 
We further note that
\begin{align*}
  & \widehat{P}_\ell(s_t|s_{1:t-1})  =  \\
  & \frac{1}{|\mathcal{J}_\ell(s_t)|}\sum_{i=1}^{|\mathcal{J}_\ell(s_t)|} \frac{\sum_{j=1}^{|\mathbf{S}_{\ell,i}|-t} \mathds{1}(\mathbf{S}_{\ell,i,j:j+t}=s_{1:t})}{\sum_{j=1}^{|\mathbf{S}_{\ell,i}|-t} \mathds{1}(\mathbf{S}_{\ell,i,j:j+t-1}=s_{1:t-1})},
\end{align*}
Furthermore, 
\begin{align*}
\mathds{1}(\mathbf{S}_{\ell,i,j:j+t}=s_{1:t})
= \prod_{s_i:i\in [t]}\mathds{1}((j+i)\in \mathcal{I}_{\ell,j}(s_i)),
\end{align*}
 Consequently, for each $t\in [T]$, the conditional distribution
$\widehat{P}_\ell(s_t|s_{1:t-1})  $ can be computed as a function of $\mathbf{h}^{(t)}$. As a result, the aggregate final node embedding $\mathbf{h}^{(T)}$ can yield $\widehat{P}_\ell(s_t|s_{1:t-1}),\ell\in \{h,m\}, t\in [T]$ as a function.   
This completes the first part of the proof. 

To prove strict improvements of PGM detectors over ESB detectors in terms of detection accuracy, we note that ESB detectors are restricted by their limited context length $T$. To provide a concrete example, consider a detection scenario characterized by the pair of probability distributions $P_h,P_m$, where all human and machine generated text sequences have length greater than $T$. That is, for any sequence $\mathbf{S}_{\ell}=(S_{\ell,1},S_{\ell,2},\cdots,S_{\ell,L})$ with $L\leq T$, we have $P_{\ell}(S_{\ell,1},S_{\ell,2},\cdots,S_{\ell,L})=0$, where $\ell\in \{h,m\}$. Furthermore, assume that the vocabulary consists of two tokens $\{a,b\}$. Both human and machine generated text sequences consist of tokens generated independently and with equal probability over the vocabulary for all indices in $\{1,2,\cdots,L-1\}$. The human generated text always ends with the token $a$ and machine generated text with the token $b$, i.e., $P(S_{h,L}=a)=P(S_{m,L}=b)=1$. 
Then, it is straightforward to see that a PGM can achieve accuracy equal to one, since the edge weights, which are functions of $ \mathcal{J}_\ell\times (\mathcal{I}_{\ell,j})_{j\in \mathcal{J}_\ell}$ 
can capture the fact that the human generated text ends in $a$ and machine generated text ends in $b$. On the other hand, for an ESB, it can be noted that  all of the empirical conditional distributions $\widehat{P}_\ell(s_t|s_{1:t-1}), t\in [T], \ell\in \{h,m\}$ converge to uniform Bernoulli distributions as $L\to \infty$. So, the ESB achieves an accuracy which is strictly less than 1 due to its limited context length, and its accuracy converges to $\frac{1}{2}$ as $L\to\infty$. This completes the proof. \qed

\section{ Experimental Setup Details}
\subsection{Datasets}
\label{app:dataset}
Our evaluation employs six distinct datasets, carefully selected to create a comprehensive benchmark spanning various domains and language models. Detailed statistics for the training, validation, and test sets, including their graph representations, are presented in Tables~\ref{tab:samples_chatgpt}, \ref{tab:samples_M4}, \ref{tab:samples_RAID}, and~\ref{tab:samples_yelp_creative_essay}.
\input{experiments/dataset_statistics}

\begin{itemize}[leftmargin=*]
\setlength{\itemsep}{-0.5pt}
\item \textbf{HC3} (Human-ChatGPT Comparison Corpus)~\citep{guo-etal-2023-hc3}: This dataset contains questions with both human-generated text (HGT) and machine-generated text (MGT) from ChatGPT. From its five available domains, we utilize four for our experiments: open-qa, wiki-csai, medicine, and finance.
\item \textbf{M4}~\citep{wang-etal-2024-m4}: The M4 dataset provides MGT from several LLMs, including Davinci, Dolly, and BloomZ, across diverse domains such as wiki-how, reddit, peerread, and arxiv. We consider four other LLMs, {DaVinci}~(DaV.), {Cohere}~(Coh.), {Dolly}~(Dol.), and {BloomZ}~(Blo.) in this dataset.
\item \textbf{RAID}~\citep{dugan-etal-2024-raid}: This large-scale dataset contains documents generated by 11 LLMs across 11 genres. Our benchmark includes four of these: recipe, book, poetry, and review. {Llama2}~(Lla.), {GPT-4}~(GT4), {MPT}, and {Mistral}~(Mis.)  are considered in this dataset.

\item \textbf{Yelp, Creative, and Essay}~\citep{mao2024raidar, verma2023ghostbuster, guo2024biscope}: For these three datasets, while texts from five LLMs are available, our analysis focuses on advanced models, specifically those generated by Claude-3-Sonnet~(Son.), Claude-3-Opus~(Opu.), and Gemini-1.0-Pro~(Gem.).
\end{itemize}

\input{experiments/nips/detailed-chatgpt}

\subsection{Experimental Setup}
\label{app:baseline}

\input{experiments/nips/detailed-chatgpt-acc-cross}

\input{experiments/nips/detailed-llm-acc}

\input{experiments/nips/detailed-llm-auc}

\stitle{Detection Baselines}.
To evaluate the fundamental detection competence of {\ours}, we compare it against leading zero-shot and training-based sequence models.
\begin{itemize}[leftmargin=*]
\setlength{\itemsep}{-0.5pt}
\item \textbf{Zero-Shot Detectors}: We include {Likelihood, Rank, Log-Rank, Entropy}~\citep{gehrmann2019gltr,solaiman2019release,ippolito2019automatic}, {DetectGPT}~\citep{mitchell2023detectgpt}, {DetectLLM} ({LRR} and {NPR})~\citep{su2023detectllm}, {DNA-GPT}~\citep{yangdna}, {Fast-DetectGPT}~\citep{baofast}, {Glimpse}~\citep{bao2025glimpse} and {Binoculars}~\citep{ma2024zero}. 
DetectGPT employs perturbations to approximate the probability distribution of the text. Fast-DetectGPT improves upon this by introducing a conditional probability curvature metric for detector optimization, bypassing traditional perturbation bottlenecks. DNA-GPT adopts a distinct approach by truncating the input text, then uses LLMs to generate the subsequent content, and finally analyzes the N-gram differences between the original and generated text.  To ensure a unified and fair comparison protocol, we utilize the OPT-2.7B model~\citep{zhang2022opt} as the default reference model for these zero-shot methods. For detailed implementation specifics, we followed the publicly available 
implementation of Fast-DetectGPT~\footnote{https://github.com/baoguangsheng/fast-detect-gpt}.
\item \textbf{Training-Based Detectors}: Our comparative evaluation includes {RoBERTa-QA}~\citep{guo-etal-2023-hc3}, {DeTeCtive}~\citep{guo2024detective}, and {RADAR}~\citep{hu2023radar}. We also present comparison results with the commercial {GPTZero} platform~\footnote{https://gptzero.me}. DeTeCtive is specifically designed for multi-source MGT detection. It employs contrastive learning to minimize the representational divergence among various MGT sources. During prediction, DeTeCtive utilizes k-nearest neighbors (KNN) to determine the classification. For our experiments, we use the DeTeCtive model trained on the OUTFOX dataset~\citep{Koike:OUTFOX:2024}. RoBERTa-QA, proposed in~\citep{guo-etal-2023-hc3} and trained on the HC3 dataset, leverages the pre-trained RoBERTa model~\citep{liu2019roberta} and fine-tunes a classification layer on the HC3 data.
\end{itemize}

\stitle{Explainable Baselines}. 
Because zero-shot models lack inherent interpretability, evaluating the faithfulness of our structural motifs requires comparing {\ours} against established post-hoc explanation methods and statistical visualizers applied to a highly accurate supervised model (RoBERTa-QA).
\begin{itemize}[leftmargin=*]
\setlength{\itemsep}{-0.5pt}
\item \textbf{LIME}~\citep{ribeiro2016should}: A post-hoc explainer that perturbs input tokens to approximate a local linear model around the prediction. This serves as our primary token-level attribution baseline, identifying which isolated words most influence the model's decision.
\item \textbf{SHAP}~\citep{NIPS2017_8a20a862}: A game-theoretic approach that assigns importance values to each token by evaluating all possible combinations of features. Like LIME, it provides a linear, token-centric explanation.
\item \textbf{GLTR}~\citep{gehrmann-etal-2019-gltr}: A statistical visualizer that highlights text based on the predictive ranking of tokens from a reference LLM. It represents explanations based purely on probability distributions rather than structural reasoning.
\item \textbf{GPT-4o (LLM-as-a-Judge)}: We employ GPT-4o as a zero-shot explainer, prompting it to analyze the text and highlight the semantic or stylistic features that indicate it is machine-generated.
\item \textbf{Random Motifs}: To rigorously verify our graph-extraction logic, we introduce a structural sanity check. In this baseline, the importance of each edge within the text's graph representation is randomly assigned. If an explanation method's MoRF/LeRF performance is indistinguishable from Random Motifs, the extracted patterns are deemed trivial noise rather than genuine linguistic fingerprints.
\end{itemize}

\stitle{Implementation.}  Our experiments are based on a three-layer GCN architecture, and GNNExplainer~\cite{ying2019gnnexplainer}, the $\lambda$ is set to default 0.05.
The input dimension of the first layer is dependent on the token size of the training set. The hidden dimension is 64, and the output dimensionality is fixed to the number of text categories. We use the Bert~\citep{devlin2019bert} tokenizer as the tokenizer.
We employ Adam~\citep{kingma2014adam} as the default optimizer with the learning rate 5E-4, 5000 epochs. For motif extraction, we adapt the GNNExplainer~\citep{ying2019gnnexplainer}to suit our analysis. Notably, the explanation method can be replaced by others, and we only use a basic post-hoc explainer here.  We follow the Refine~\citep{refine} to implement the GNNExplainer. The optimizer for GNNExplainer is Adam with a learning rate of 1E-3, 100 epochs. 
 All experiments are conducted on a Linux machine with 8 NVIDIA A100 GPUs, each with 40GB of memory. The software environment is CUDA 11.3 and Driver Version 550.54.15. We used Python 3.9.13, Pytorch 1.10.0, and torch-geometric 2.0.3 to construct our project.

\section{ Detailed Experiment Results}
\label{extend:detailedexp}

\subsection{Detection Experiments}
\label{extend:detections}
In our experiments, we consider in-domain detection and cross-domain detection in the same dataset and report the results in this section. We report the results under ACC and AUC metrics. For GPTZero, since it provides a binary output, we consider its ACC and AUC values to be equivalent.

\stitle{In-Domain Detection}. We provide the detailed experiment results for distinguishing HGTs and MGTs by ChatGPT in Table~\ref{tab:humanchatgpt_detailed_acc} and Table \ref{tab:humanchatgpt_detailed_auc}. The results demonstrate that {\ours} achieves the best performance across all domains under both ACC and AUC metrics, aligned with our analysis.  In addition, we also provide the experiment results between HGT and MGT by other LLMs in Table \ref{tab:humanm4_acc}, \ref{tab:humanraid_acc}, \ref{tab:humanyelp_acc}, \ref{tab:humanm4_auc}, \ref{tab:humanraid_auc}, and \ref{tab:humanyelp_auc}. {\ours} performs consistently well on various LLMs and achieves the best performance, indicating the effectiveness of PGM for MGT detection tasks.

\input{experiments/statistation_hc3}

\stitle{Cross-Domain Detection}. To further analyze the generality of {\ours}, we conduct cross-domain detection experiments. We use the open-qa, wiki-how, and books domains in HC, M4, and RAID datasets as the training domain and test on other domains, respectively. For the zero-shot baselines and RADAR, we use the training data as a reference to learn a threshold and apply it to the test domain. For the RoBERTa-QA, we follow its pipeline to fine-tune the RoBERTa on one domain and test on other domains. As Table~\ref{tab:humanchatgpt_detailed_acc_cross} shows, {\ours} performs poorly on some domains, such as the reddit domain on the M4 dataset. 
Rather than being an inherent limitation of our explainable framework, this performance gap is primarily due to the current in-domain training setting, which restricts the diversity of the learned structural motifs compared to models pre-trained on massive, large-scale corpora.

\input{experiments/ablation_tokenizer}

\stitle{Statistical Significance Analysis}. To further demonstrate the robustness of {\ours}, we conducted a Statistical Significance Analysis. Specifically, we repeated our experiments five times on the HC3 dataset, each with a distinct random seed, and the resulting performance metrics are detailed in Table~\ref{tab:statistic_model}. The consistently high performance across these different runs indicates the stable and reliable nature of {\ours}.

\stitle{Ablation Study}. To investigate the impact of different graph characteristics on the MGT detection task, we performed ablation experiments on graph categories, specifically comparing undirected versus directed graphs and weighted versus unweighted graphs. {To verify the influence of token semantics on detection performance, we also performed an ablation study on the token node initialization method, where token nodes are initialized using Bert token embeddings.}
In our experiments, we use {-U} and {-W} to represent undirected graphs and weighted graphs, {while {-Bert} indicates replacing Bert token embeddings with a simpler initialization method.}
\input{experiments/ablation_graph_hc3}

\input{experiments/time_comsumption}
We further investigated the impact of different tokenizers on the MGT detection task. Our default tokenizer is Bert's tokenizer. To assess the influence of tokenization, we conducted experiments using GPT-2's tokenizer. The results of this comparison are presented in Table~\ref{tab:tokenizers}. Our findings indicate that the choice between Bert's and GPT-2's tokenizers did not significantly affect the overall detection performance.

{To investigate the effect of sliding window size on detection performance, we conduct ablation studies, and the results are presented in Table~\ref{tab:window_ablation}. As the window size increases, the detection accuracy initially improves and then declines. Based on these results, we set 20 as the default sliding-window size in our experiments.}

\stitle{Time Consumption}. Compared to other training-based methods, {\ours} has an additional pipeline, the graph construction phase. Specifically, its time complexity for graph construction is $O(LW^2)$, where $L$ represents the length of the sentence and $W$ denotes the size of the sliding window. We also evaluated the test time efficiency of {\ours} in comparison to several other baselines. As detailed in Table~\ref {tab:inference_time}, {\ours} demonstrates the lowest time consumption during the testing phase.

\subsection{Extended Motifs Evaluation}
\label{extend:motifs_eval}
\stitle{XAI Protocol Evaluation}. We follow Section~\ref{exp:motifs} to report the explainable motifs evaluation results on the HGT and ChatGPT-generated datasets. As the detailed results show in Figure~\ref{fig:xai_motifs_evaluation}, the explainable motifs are effective in most cases and obtain better results than baselines from both LeRF and MoRF protocols. However, in the medicine domain in HC3, the explainable motifs are not better than random motifs. The potential reason could be the distributed nature of the explainable motifs across numerous nodes and edges. Consequently, the deletion of some edges does not drastically impede the graph network's ability to accurately perform detection. For instance, in the medicine domain of the HC3 dataset, a significant performance drop in the GNN is observed when the proportion of deleted edges surpasses 70\%.

\input{experiments/ablation_windowsize}

\input{experiments/morf_lerf}

\begin{figure}[t]
\centering
\includegraphics[width=\linewidth]{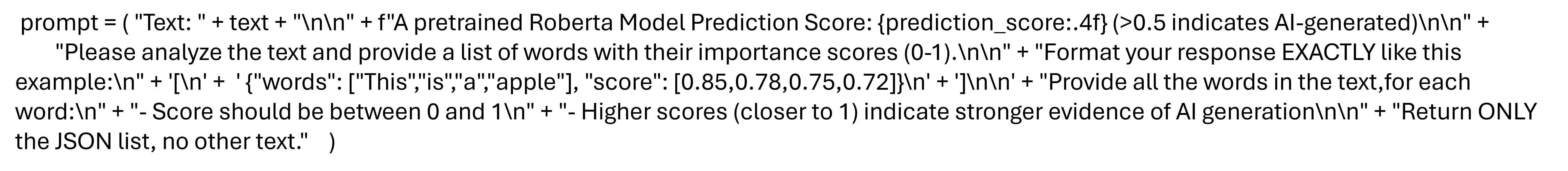}
\caption{The prompt of GPT-4o as an explainer.}
\label{fig:prompt}
\end{figure}

\input{experiments/figure_lerf_comparison}
\input{experiments/figure_morf_comparison}
\input{experiments/figure_lambda}
We also provide the detailed comparison results between ours and the baselines in Figure~\ref{fig:extensive_baseline_lerf} and ~\ref{fig:extensive_baseline_morf}. For these baselines, we use RoBERTa-QA, a well-trained model in the HC3 dataset, as the model to be explained. Notably, the reason we use RoBERTa-QA is that it has the best performance in the HC3 dataset, and it can be replaced by other models. GLTR is a tool to analyze a piece of text and visualize these statistical patterns, which uses a language model to determine the probability of each word appearing in its context. In this paper, we follow the original code to use GPT2 as the default language model. Besides, we also consider using the GPT-4o as a baseline for the LLM explainer. The prompt is shown in Figure ~\ref{fig:prompt}. 

Our results are provided in Figure~\ref{fig:extensive_baseline_lerf} and~\ref{fig:extensive_baseline_morf}. As shown, the motifs from {\ours} are more effective and consistently have a better performance than baselines. From the MoRF protocol, when the 20\% important edges are removed, the explainable motifs cause more than an average 15\% accuracy drop on the HC3 dataset, while other explanations get less than 10\% accuracy decline. Under the LeRF protocol, the explainable motifs cause a lower performance drop than other motifs. GPT-4o performs well under the MoRF setting but fails under the LeRF setting. 

\input{experiments/table_lambda}

{We conducted ablation experiments on the explainer's hyperparameter $\lambda$ using the HC3 dataset and evaluated the interpretation results following the MoRF protocol. The results are presented in Figure~\ref{fig:appendix_lambda_ablation} and Table~\ref{tab:lambda_ablation}. As shown, the explanation performance is largely robust to different choices of $\lambda$.}

\stitle{Motifs Statistical Analysis}. We provide more statistical analysis on M4 and RAID datasets. Table~\ref{tab:statistic_motifs},~\ref{tab:statistic_motifs_m4}, and~\ref{tab:statistic_motifs_raid} reveal distinct motif fingerprints, frequency variations between HGT and MGT across tokens(nodes) and token-token co-occurrences(edges). Selecting the top 0.05\% of edges as global explainable motifs highlights a notable difference: HGT shows a higher ratio of token and token-token co-occurrences compared to MGT. This suggests that for MGT detection, word-to-word connections are more influential than for HGT detection, given the same number of tokens. One possible explanation is that language models excel at utilizing diverse word collocations, while humans tend to rely on more conventional patterns.

\input{experiments/statistics_analysis}

\stitle{Visualizations}. To visualize the extracted motifs, we use the PubMed dataset, which contains machine-generated text (MGT) samples produced by multiple LLMs, including GPT-4, Claude-3, and Davinci. We analyze the motifs at two levels of granularity: word-level structures and higher-order phrase/sentence-level structures. Specifically, word-level motifs are extracted from one-hop neighbor subgraphs, and we visualize the top 20\% of tokens ranked by motif scores in Table~\ref{tab:fig:casestudy_words}. To capture more complex semantic and phrasal relationships, we further extract higher-order motifs from two-hop subgraphs and visualize the top 2\% motifs in Table~\ref{tab:fig:casestudy_phases}.

To better illustrate the structural differences among LLM-generated texts, we project the graph motifs back onto the original text space. Unlike traditional N-gram statistics that mainly reflect token occurrence frequencies, our graph motifs explicitly reveal how different LLMs organize and assemble linguistic structures. Word-level motifs primarily capture local co-occurrence preferences, while higher-order motifs uncover more complex semantic, syntactic, and phrasal dependencies.

Our visualization results reveal clear structural differences across models, even when they use highly similar domain vocabularies. For example, GPT-4 and Davinci frequently employ the same biomedical terms such as ``sleep'' and ``patients'', yet their graph topologies differ substantially. GPT-4 tends to form dense and globally connected action-oriented structures, whereas Davinci relies more heavily on localized and linear phrasal associations. These findings demonstrate that {\ours} can explicitly expose the latent structural organization patterns underlying different LLMs, providing an interpretable and human-auditable view of machine-generated text.

\section{LLM Usage}  
In this paper, we leverage LLMs, including ChatGPT and Gemini, to refine sentence-level writing.

\input{figures/case_table}

%% file: experiments/dataset_statistics.tex
\begin{table*}[!h]
\caption{The details of the dataset for detection between HGT and MGT generated by ChatGPT.} 
\vspace{-0.5em}
\centering
\setlength\tabcolsep{3.3pt}
\scalebox{0.75}{
     \begin{tabular}{lcccccccccccc}
        \toprule
         &  
        \multicolumn{4}{c}{HC3} &
        \multicolumn{4}{c}{M4} &
        \multicolumn{4}{c}{RAID} \\
        \cmidrule(r){2-5}\cmidrule(r){6-9}\cmidrule(r){10-13} 
          & {open-qa} & {wiki-csai} & {medicine} & finance & wiki-how & reddit & peerread & arxiv & recipe & book & poetry & review   \\
     \toprule    
     \# Training              & 2,000       &1,384        &2,000      &2,000      &2,000  & 872 & 2,000 & 2,000 & 2,000 & 2,000 & 2,000 & 1,793  \\
     \# Validation            & 100        &100         &100       &100       &100   &100  &100   &100  &100  &100  &100   &100 \\
     \# Test                  & 200        &200         &200       &200       &200  &200  &200  &200 
 &200  &200  &200  &200 \\
     \# Nodes         & 15,974     &12,069      &8,127     &9,581     &20,061  & 11,276 & 18,926 & 10,526 & 6,562 & 20,515 & 16,818 & 17,024    \\
     \# Edges  & 3,262K  &2,635K   &2,063K &2,326K   &6,823K & 4,658K & 8,591K & 3,595K & 2,119K & 7,076K & 5,059K & 5,448K\\
    \bottomrule
    \end{tabular}
    }   
\label{tab:samples_chatgpt}
\end{table*}
\begin{table*}[!h]
\caption{The details of the M4 dataset for detection between HGT and MGT generated by LLMs.} 
\vspace{-0.5em}
\centering
\setlength\tabcolsep{3.3pt}
\scalebox{0.78}{
     \begin{tabular}{lcccccccccccc}
        \toprule
         &  
        \multicolumn{4}{c}{reddit} &
        \multicolumn{4}{c}{peerread} & 
        \multicolumn{4}{c}{arxiv} \\
        \cmidrule(r){2-5}\cmidrule(r){6-9}\cmidrule(r){10-13} 
         & Davinci & Cohere & Dolly & BloomZ & Davinci & Cohere & Dolly & BloomZ & Davinci & Cohere & Dolly & BloomZ  \\
     \toprule    
     \# Training              & 2,000 &2,000 & 2,000  & 2,000  &872 & 824 &872 & 830 &2,000 &2,000 &2,000 &2,000 \\
     \# Validation            & 100  &100   &100   &100   &100   &100  &100   &98  &100  &100  &100   &100 \\
     \# Test                  & 200  &200   &200  &200   &200  &198  &200  &192  &200  &200  &200  &200 \\
     \# Nodes         & 20,867 &21,701 &21,344  & 20,944 & 11,059  & 10,837 &14,366 &11,340 &10,153 & 10,724 &12,039 &11,468 \\
     \# Edges  & 7,175K &7,055K &7,157K & 6,601K &4,139K &3,933K &5,924K &4,296K &3,343K &3,376K &4,132K & 4,011K\\
    \bottomrule
    \end{tabular}
    }   
\label{tab:samples_M4}
\end{table*}
\begin{table*}[!h]
\caption{The details of the RAID dataset for detection between HGT and MGT generated by LLMs.} 
\vspace{-0.5em}
\centering
\setlength\tabcolsep{3.3pt}
\scalebox{0.8}{
     \begin{tabular}{lcccccccccccc}
        \toprule
         &
        \multicolumn{4}{c}{recipes} &
        \multicolumn{4}{c}{poetry} & 
        \multicolumn{4}{c}{reviews} \\
        \cmidrule(r){2-5}\cmidrule(r){6-9}\cmidrule(r){10-13} 
         & Llama  & GPT-4 & MPT & Mistral & Llama 2  & GPT-4 & MPT & Mistral & Llama 2  & GPT-4 & MPT & Mistral \\
     \toprule    
     \# Training              & 2,000 & 2,000  & 2,000 &2,000      &2,000 &2,000 &2,000  & 2,000  & 1,793  & 1,793 & 1,793 & 1,793\\
     \# Validation            & 100  & 100   & 100  &100       &100   &100  &100   &100  &100  &100  &100   &100 \\
     \# Test                  & 200  & 200   & 200  &200       &200  &200  &200  &200  &200  &200  &200  &200 \\
     \# Nodes         & 6,904 & 6,701 & 13,466 &8,833   &16,696 &17,152 &19,476 &17,523 &17,004 & 17,387 & 19,371 &17,843\\
     \# Edges  & 2,125K & 2,163K & 4,066K &2,586K   &4,766K &4,527K &5,924K &4,835K &5,039K &5,694K &6,017K & 5,142K\\
    \bottomrule
    \end{tabular}
    }   
\label{tab:samples_RAID}
\end{table*}
\begin{table*}[!h]
\caption{The details of the dataset for detection between HGT and MGT generated by LLMs in Yelp, Creative, and Essay Dataset. Sonnet and Opus are short for Claude3-Sonnet and Claude-3-Opus. } 
\vspace{-0.5em}
\centering
\scalebox{0.9}{
     \begin{tabular}{lccccccccc}
        \toprule
         &
        \multicolumn{3}{c}{Yelp} &
        \multicolumn{3}{c}{Essay} & 
        \multicolumn{3}{c}{Creative} \\
        \cmidrule(r){2-4}\cmidrule(r){5-7}\cmidrule(r){8-10} 
         & Sonnet  & Opus & Gemini & Sonnet  & Opus & Gemini & Sonnet  & Opus & Gemini \\
     \toprule    
     \# Training              & 2,000  &2,000   &2,000   &1,500   &1,500   & 1,500   & 1,500   & 1,500     & 1,500  \\
     \# Validation            & 200   &200    &200    &100    &100    & 100    & 100    & 100      & 100 \\
     \# Test                  & 200   &200    &200    &200    &200    & 200    & 200    & 200      & 200 \\
     \# Nodes                & 11,581 &11,308 &11,350 &20,836 &20,748 & 20,868 & 20,597 & 20,057   & 19,936 \\
     \# Edges               & 1,940K  &1,886K &1,778K &8,422K &8,038K & 7,989K & 7,187K & 6,308K   & 6,250K  \\
    \bottomrule
    \end{tabular}
    }   
\label{tab:samples_yelp_creative_essay}
\end{table*}

%% file: experiments/nips/detailed-chatgpt.tex
\begin{table*}[!h]
\caption{Detection comparisons with SOTA methods on ACC between HGT and ChatGPT-generated texts. 
The best results are shown in bold font. The second-best results are shown in underlined.  F-DetectGPT and R-QA are short for Fast-DetectGPT and RoBERTa-QA. } 
\centering
\scalebox{0.80}{
    \begin{tabular}{lc@{\hspace{0.15cm}}c@{\hspace{0.15cm}}c@{\hspace{0.15cm}}cc@{\hspace{0.15cm}}c@{\hspace{0.15cm}}c@{\hspace{0.15cm}}cc@{\hspace{0.15cm}}c@{\hspace{0.15cm}}c@{\hspace{0.15cm}}cc}
        \toprule
        & \multicolumn{4}{c}{HC3} & \multicolumn{4}{c}{M4} & \multicolumn{4}{c}{RAID} \\
        \cmidrule(r){2-5}\cmidrule(r){6-9}\cmidrule(r){10-13}
         Method & open-qa & wiki-csai & medicine & finance & wiki-how & reddit & peerread & arxiv & recipe &	book & poetry & review & Avg. \\
     \midrule
     
     \rowcolor{gray!20} \multicolumn{14}{c}{\textbf{Zero-Shot Methods}} \\
    NPR         & 0.65	&	0.92	&	0.91	&	0.85	&	0.61	&	0.68	&	0.83	&	0.72	&	0.84	&	0.83	&	0.50	&	0.97	&	0.78
\\
    LRR         & 0.98	&	0.95	&	0.98	&	0.94	&	0.82	&	0.82	&	\textbf{1.00}	&	0.81	&	0.94	&	0.81	&	0.75	&	0.98	&	0.90
\\
    Rank        & 0.54	&	0.53	&	0.54	&	0.51	&	0.57	&	0.55	&	0.58	&	0.61	&	0.51	&	0.65	&	0.53	&	0.54	&	0.56
\\
  Entropy     & 0.92	&	0.76	&	0.77	&	0.61	&	0.83	&	0.81	&	0.68	&	0.60	&	0.80	&	0.71	&	0.49	&	0.62	&	0.72
\\
    LogRank     & 0.77	&	0.73	&	0.72	&	0.58	&	0.82	&	0.95	&	0.80	&	0.92	&	0.81	&	0.93	&	0.84	&	0.79	&	0.81
\\
  Likelihood  & 0.85	&	0.81	&	0.76	&	0.58	&	0.85	&	{0.96}	&	0.80	&	0.92	&	0.83	&	0.96	&	0.82	&	0.78	&	0.83
\\

    Glimpse      & 0.95	 &0.98	&\underline{0.99}	&\underline{0.99} & 0.97	& 0.91	& 0.87	& \underline{0.99} & 0.94	& 0.96	& 0.76	& 0.98 & 0.94
\\
    Binoculars   & 0.92	 &\textbf{1.00}	&\textbf{1.00}	&\textbf{1.00}  & 0.77	& \underline{0.97}	&\textbf{1.00}	&\textbf{1.00}  &\textbf{1.00}	&0.96	&\textbf{0.99}	&\underline{0.99}  &\underline{0.97}
\\
    DNAGPT      & 0.63	&	0.79	&	0.63	&	0.88	&	0.60	&	0.79	&	0.53	&	0.81	&	0.71	&	0.82	&	0.75	&	0.59	&	0.71
\\ 
    DetectGPT   & 0.46	&	0.63	&	0.76	&	0.68	&	0.58	&	0.66	&	0.59	&	0.61	&	0.56	&	0.66	&	0.59	&	0.68	&	0.62
\\
    F-DetectGPT     & 0.95	&	\underline{0.99}	&	0.98	&	0.97	&	0.88	&	0.94	&	\textbf{1.00}	&	\textbf{1.00}	&	\underline{0.99}	&	\underline{0.97}	&	0.93	&	\textbf{1.00}	&	0.97
\\

     \rowcolor{gray!20} \multicolumn{14}{c}{\textbf{Training-Based Methods}} \\
    
    R-QA  & \textbf{1.00}*	&	\textbf{1.00}*	&	\textbf{1.00}*	&	\underline{0.99}*	&	0.88	&	{0.96}	&	\underline{0.99}	&	0.95	&	0.83	&	0.86	&	0.50	&	\textbf{1.00}	&	0.91
\\
    RADAR       & 0.52	&	0.81	&	0.55	&	0.75	&	0.46	&	0.93	&	0.88	&	0.77	&	0.61	&	\underline{0.97}	&	0.61	&	0.89	&	0.73
\\
    GPTZero     & 0.58	&	0.69	&	0.96	&	0.84	&	0.54	&	0.82	&	0.96	&	0.69	&	0.61	&	0.84	&	0.48	&	0.78	&	0.73
\\  
    DeTeCtive   & \underline{0.99}	&	0.79	&	\underline{0.99}	&	0.89	&	\underline{0.89}	&	0.93	&	0.90	&	{0.98}	&	{0.94}	&	0.95	&	\underline{0.97}	&	{0.97}	&	{0.93}
\\
    \midrule
    {\ours}     & 0.97	&	0.96	&	0.98	&	{0.98}	&	\textbf{0.97}	&	\textbf{0.99}	&	0.98	&	0.96	&	\underline{0.99}	&	\textbf{1.00}	&	\textbf{0.99}	&	{0.96}	&	\textbf{0.98}
\\
     
    \bottomrule
    \end{tabular}
    }   
\label{tab:humanchatgpt_detailed_acc}
\end{table*}

\begin{table*}[!h]
\caption{MGT detection AUC performance comparisons with SOTA methods on HGT and ChatGPT-generated texts. 
The best results are shown in bold font. The second-best results are shown in underlined. F-DetectGPT and R-QA are short for Fast-DetectGPT and RoBERTa-QA. } 
\centering
\setlength\tabcolsep{3.3pt}
\scalebox{0.83}{
    \begin{tabular}{l|c@{\hspace{0.15cm}}c@{\hspace{0.15cm}}c@{\hspace{0.15cm}}cc@{\hspace{0.15cm}}c@{\hspace{0.15cm}}c@{\hspace{0.15cm}}cc@{\hspace{0.15cm}}c@{\hspace{0.15cm}}c@{\hspace{0.15cm}}cc}
        \toprule
        & \multicolumn{4}{c}{HC3} & \multicolumn{4}{c}{M4} & \multicolumn{4}{c}{RAID} \\
        \cmidrule(r){2-5}\cmidrule(r){6-9}\cmidrule(r){10-13}
         Method & open-qa & wiki-csai & medicine & finance & wiki-how & reddit & peerread & arxiv & recipe &	book & poetry & review & Avg. \\
     \midrule
     
     \rowcolor{gray!20} \multicolumn{14}{c}{\textbf{Zero-Shot Methods}} \\
    NPR         & \textbf{1.00}	&	\textbf{1.00}	&	\textbf{1.00}	&	\textbf{1.00}	&	0.95	&	\underline{0.99}	&	0.81	&	0.98	&	\textbf{1.00}	&	\textbf{1.00}	&	0.89	&	\textbf{1.00}	&	0.97
\\
    LRR         & \textbf{1.00}	&	\underline{0.99}	&	\textbf{1.00}	&	0.99	&	0.93	&	0.98	&	\textbf{1.00}	&	\underline{0.99}	&	\underline{0.99}	&	\underline{0.99}	&	0.85	&	\textbf{1.00}	&	0.98
\\
    Rank        & \textbf{1.00}	&	0.77	&	\underline{0.99}	&	0.81	&	0.94	&	0.92	&	\underline{0.97}	&	0.95	&	0.79	&	\underline{0.99}	&	0.87	&	\textbf{1.00}	&	0.92
\\
    Entropy     & \underline{0.99}	&	0.85	&	\underline{0.99}	&	0.97	&	0.91	&	0.91	&	0.60	&	0.75	&	0.97	&	0.88	&	0.75	&	0.97	&	0.88
\\
    LogRank     & \textbf{1.00}	&	\textbf{1.00}	&	\textbf{1.00}	&	\textbf{1.00}	&	0.95	&	\underline{0.99}	&	0.82	&	0.98	&	\textbf{1.00}	&	\textbf{1.00}	&	0.89	&	\textbf{1.00}	&	0.97
\\
    Likelihood  & \textbf{1.00}	&	\textbf{1.00}	&	\textbf{1.00}	&	\textbf{1.00}	&	0.95	&	\underline{0.99}	&	0.69	&	0.97	&	\textbf{1.00}	&	\textbf{1.00}	&	0.90	&	\textbf{1.00}	&	0.96
\\
   Glimpse      & \textbf{1.00} &	\textbf{1.00}	& \textbf{1.00}	& \textbf{1.00}	& \textbf{0.99}	& \textbf{1.00}	& 0.92 &	\textbf{1.00}		& \textbf{1.00}	& \textbf{1.00}	& 0.85	& \textbf{1.00}		& 0.98
\\
    Binoculars   & 0.98	&  \textbf{1.00}	& \textbf{1.00}	& \textbf{1.00}	 	& 0.90	& \textbf{1.00}	& \textbf{1.00}	& \textbf{1.00}	 	& \textbf{1.00}	& \textbf{1.00}	& \underline{0.99}	& \textbf{1.00}	 	& \underline{0.99} 
\\
    DNAGPT      & 0.72	&	0.95	&	0.91	&	0.94	&	0.97	&	0.95	&	0.56	&	0.94	&	0.94	&	0.97	&	0.84	&	\underline{0.98}	&	0.89
\\ 
    DetectGPT   & 0.35	&	0.59	&	0.70	&	0.61	&	0.68	&	0.71	&	0.70	&	0.43	&	0.65	&	0.77	&	0.84	&	0.84	&	0.66
\\
    F-DetectGPT     & \textbf{1.00}	&	\textbf{1.00}	&	\textbf{1.00}	&	\underline{0.98}	&	\underline{0.96}	&	\underline{0.99}	&	\textbf{1.00}	&	\textbf{1.00}	&	\textbf{1.00}	&	\textbf{1.00}	&	{0.98}	&	\textbf{1.00}	&	\underline{0.99}
\\

     \rowcolor{gray!20} \multicolumn{14}{c}{\textbf{Training-Based Methods}} \\
    R-QA  &\textbf{1.00}*	&	\textbf{1.00}*	&	\textbf{1.00}*	&	\textbf{1.00}*	&	0.94	&	\textbf{1.00}	&	\textbf{1.00}	&	\textbf{1.00}	&	0.90	&	0.99	&	0.95	&	\textbf{1.00}	&	0.98
 \\
    RADAR       & 0.20	&	0.77	&	0.41	&	0.68	&	0.40	&	0.97	&	\textbf{1.00}	&	0.95	&	\underline{0.99}	&	\textbf{1.00}	&	0.88	&	0.91	&	0.76
\\
    GPTZero     & 0.58	&	0.69	&	0.96	&	0.84	&	0.54	&	0.82	&	0.96	&	0.69	&	0.61	&	0.84	&	0.48	&	0.78	&	0.73
\\  
    DeTeCtive   & \textbf{1.00}	&	0.84	&	\underline{0.99}	&	0.89	&	0.90	&	0.98	&	0.90	&	\underline{0.99}	&	\underline{0.99}	&	0.97	&	0.97	&	0.97	&	0.95
\\

    \midrule
    {\ours}     & \textbf{1.00}	&	\underline{0.99}	&	\textbf{1.00}	&	\textbf{1.00}	&	\textbf{0.99}	&	\textbf{1.00}	&	\textbf{1.00}	&	\textbf{1.00}	&	\textbf{1.00}	&	\textbf{1.00}	&	\textbf{1.00}	&	\textbf{1.00}	&	\textbf{1.00}
 \\
    \bottomrule
    \end{tabular}
    }   
\label{tab:humanchatgpt_detailed_auc}
\end{table*}

%% file: experiments/nips/detailed-chatgpt-acc-cross.tex
\begin{table*}[!h]
\caption{Cross-domain MGT detection ACC performance comparisons with SOTA methods on HGT and ChatGPT-generated texts. 
The best results are shown in bold font. The second-best results are shown in underlined. F-DetectGPT and R-QA are short for Fast-DetectGPT and RoBERTa-QA. } 
\centering
\setlength\tabcolsep{3.3pt}
\scalebox{1.0}{
    \begin{tabular}{lcccccccccc}
        \toprule
        & \multicolumn{3}{c}{HC3} & \multicolumn{3}{c}{M4} & \multicolumn{3}{c}{RAID} \\
        \cmidrule(r){2-4}\cmidrule(r){5-7}\cmidrule(r){8-10}
         Method & wiki-csai & medicine & finance  & reddit & peerread & arxiv & recipe  & poetry & review & Avg. \\
     \midrule
     
     \rowcolor{gray!20} \multicolumn{11}{c}{\textbf{Zero-Shot Methods}} \\
    NPR         & 0.97	&	{0.97}	&	\underline{0.98}	&	\underline{0.93}	&	0.97	&	0.77	&	0.55	&	0.72	&	0.97	&	0.87

\\
    LRR         & {0.98}	&	0.94	&	{0.96}	&	\underline{0.93}	&	0.94	&	0.76	&	0.52	&	0.69	&	0.93	&	0.85

\\
    Rank        & 0.65	&	0.94	&	0.65	&	0.82	&	0.56	&	0.66	&	0.54	&	0.80	&	0.80	&	0.71

\\
    Entropy     & 0.71	&	0.91	&	0.87	&	0.61	&	0.75	&	0.50	&	0.50	&	0.58	&	0.78	&	0.69

\\
    
    LogRank     & {0.98}	&	{0.97}	&	\underline{0.98}	&	{0.92}	&	0.83	&	0.71	&	0.53	&	0.73	&	0.97	&	0.85

\\
    Likelihood  & 0.97	&	0.96	&	\underline{0.98}	&	0.88	&	0.80	&	0.67	&	0.57	&	0.70	&	\underline{0.99}	&	0.84

\\
     Glimpse      & \textbf{1.00} & \underline{0.98} &\textbf{ 1.00} & \textbf{0.96} & 0.88 & \textbf{0.99} & 0.96 & 0.78 & \textbf{1.00} & \underline{0.95}
\\
    Binoculars   & 0.97 & 0.97 & 0.97 & 0.89 & 0.89 & 0.89 & \textbf{1.00} & \textbf{1.0}0 & \textbf{1.00} & \underline{0.95}
\\
    DNAGPT      & 0.86	&	0.87	&	0.84	&	{0.92}	&	0.49	&	0.85	&	\underline{0.84}	&	0.67	&	0.96	&	0.81

\\ 
    DetectGPT   & 0.52	&	0.58	&	0.53	&	0.56	&	0.54	&	0.50	&	0.56	&	0.67	&	0.73	&	0.58

\\
    F-DetectGPT     & \underline{0.99}	&	\textbf{0.99}	&	0.95	&	\underline{0.93}	&	\textbf{1.00}	&	\underline{0.94}	&	\textbf{1.00}	&	{0.89}	&	\textbf{1.00}	&	\textbf{0.97}

\\

     \rowcolor{gray!20} \multicolumn{11}{c}{\textbf{Training-Based Methods}} \\
    R-QA  & 0.53	&	0.65	&	0.53	&	0.77	&	0.93	&	\textbf{0.99}	&	0.70	&	0.86	&	0.85	&	0.76

\\
    RADAR       & 0.79	&	0.53	&	0.74	&	{0.92}	&	0.77	&	0.87	&	0.65	&	0.74	&	0.88	&	0.77

\\
    DeTeCtive   & 0.68	&	0.58	&	0.60	&	0.76	&	0.76	&	0.65	&	0.48	&	0.72	&	0.89	&	0.68

\\
    \midrule
    {\ours}     & 0.71	&	0.82	&	0.64	&	0.59	&	\underline{0.99}	&	0.93	&	0.50	&	\underline{0.93}	&	0.97	&	0.79

\\
     
    \bottomrule
    \end{tabular}
    }   

\label{tab:humanchatgpt_detailed_acc_cross}
\end{table*}

\begin{table*}[!h]
\caption{MGT detection ACC performance comparisons with SOTA methods on HGT and MGT on M4 dataset. 
The best results are shown in bold font. The second-best results are shown in underlined. F-DetectGPT and R-QA are short for Fast-DetectGPT and RoBERTa-QA. } 
\centering
\setlength\tabcolsep{3.3pt}
\scalebox{0.85}{
    \begin{tabular}{lc@{\hspace{0.3cm}}c@{\hspace{0.3cm}}cc@{\hspace{0.3cm}}c@{\hspace{0.3cm}}cc@{\hspace{0.3cm}}c@{\hspace{0.3cm}}cc@{\hspace{0.3cm}}c@{\hspace{0.3cm}}cc}
        \toprule
        & \multicolumn{3}{c}{DaVinci} & \multicolumn{3}{c}{Cohere} & \multicolumn{3}{c}{Dolly} & \multicolumn{3}{c}{BloomZ}  \\
        \cmidrule(r){2-4}\cmidrule(r){5-7}\cmidrule(r){8-10}\cmidrule(r){11-13}
   Method & reddit & peerread & arxiv & reddit & peerread & arxiv & reddit & peerread & arxiv & reddit & peerread & arxiv & Avg. \\
     \midrule
     
     \rowcolor{gray!20} \multicolumn{14}{c}{\textbf{Zero-Shot Methods}} \\
     NPR         & 0.67	&	0.74	&	0.49	&	0.65	&	0.83	&	0.53	&	0.51	&	0.52	&	0.61	&	0.52	&	0.71	&	0.55	&	0.61 \\
    LRR         & 0.86	&	0.95	&	0.50	&	0.68	&	0.94	&	0.63	&	0.75	&	0.82	&	0.66	&	0.75	&	\underline{0.98}	&	0.59	&	0.76 \\
    
    Rank        &0.57	&	0.51	&	0.45	&	0.56	&	0.54	&	0.52	&	0.50	&	0.53	&	0.57	&	0.55	&	0.51	&	0.53	&	0.53\\
    Entropy     &0.78	&	0.71	&	0.37	&	0.73	&	0.53	&	0.58	&	0.54	&	0.46	&	0.58	&	0.63	&	0.34	&	0.62	&	0.57\\
    
    LogRank     &0.83	&	0.77	&	0.40	&	0.94	&	0.81	&	0.90	&	0.75	&	0.69	&	0.73	&	0.71	&	0.39	&	0.77	&	0.72\\
    Likelihood  &0.89	&	0.77	&	0.40	&	0.95	&	0.78	&	0.89	&	0.63	&	0.65	&	0.71	&	0.56	&	0.34	&	0.72	&	0.69 \\
    Glimpse    & 0.77	&0.95	&0.51		&0.95	&0.88	&1.00		&0.64	&0.68	&0.75		&0.52	&0.43	&0.88		&0.75
\\
    Binoculars & \textbf{0.98}	&\textbf{1.00}	&0.51		&\textbf{0.98}	&0.96	&\textbf{0.98}		&0.83	&\underline{0.99}	&\underline{0.87}		&0.58	&0.62	&0.77		&0.84
 \\
    DNAGPT      & 0.75	&	0.47	&	0.36	&	0.90	&	0.47	&	0.86	&	0.51	&	0.53	&	0.54	&	0.45	&	0.49	&	0.57	&	0.58 \\ 
    
    DetectGPT   & 0.56	&	0.53	&	0.35	&	0.63	&	0.60	&	0.47	&	0.54	&	0.46	&	0.45	&	0.58	&	0.57	&	0.62	&	0.53 \\
    F-DetectGPT     & \underline{0.97}	&	\textbf{1.00}	&	0.46	&	\underline{0.96}	&	\underline{0.99}	&	\textbf{0.98}	&	0.90	&	\underline{0.99}	&	{0.82}	&	0.43	&	0.51	&	0.69	&	0.81 \\
    
     \rowcolor{gray!20} \multicolumn{14}{c}{\textbf{Training-Based Methods}} \\
    R-QA  & {0.93}	&	1.00	&	0.55	&	0.95	&	0.97	&	0.89	&	0.95	&	0.55	&	0.71	&	0.50	&	0.50	&	0.52	&	0.75 \\
    RADAR       & 0.84	&	0.88	&	0.57	&	0.87	&	0.85	&	0.60	&	0.66	&	0.77	&	0.53	&	0.80	&	0.79	&	0.30	&	0.71 \\
    GPTZero     & 0.86	&	\underline{0.99}	&	0.36	&	0.84	&	0.92	&	0.65	&	0.76	&	0.58	&	0.50	&	0.61	&	0.53	&	0.46	&	0.67 \\ 
    DeTeCtive   & 0.90	&	0.85	&	\textbf{0.95}	&	0.84	&	0.76	&	\underline{0.95}	&	\textbf{0.96}	&	0.75	&	\textbf{0.98}	&	\underline{0.94}	&	0.89	&	\underline{0.92}	&	\underline{0.89}\\

    \midrule
    {\ours}     & \underline{0.97}	&	\textbf{1.00}	&	\underline{0.87}	&	\textbf{0.98}	&	\textbf{1.00}	&	0.94	&	\underline{0.95}	&	\textbf{1.00}	&	0.77	&	\textbf{1.00}	&	\textbf{1.00}	&	\textbf{0.95}	&	\textbf{0.95}\\
     
    \bottomrule
    \end{tabular}
    }   
\label{tab:humanm4_acc}
\end{table*}

%% file: experiments/nips/detailed-llm-acc.tex
\begin{table*}[!h]
\caption{MGT detection ACC performance comparisons with SOTA methods on HGT and MGT on RAID dataset. 
The best results are shown in bold font. The second-best results are shown in underlined. F-DetectGPT and R-QA are short for Fast-DetectGPT and RoBERTa-QA.} 
\centering
\setlength\tabcolsep{3.3pt}
\scalebox{0.85}{
    \begin{tabular}{lc@{\hspace{0.3cm}}c@{\hspace{0.3cm}}cc@{\hspace{0.3cm}}c@{\hspace{0.3cm}}cc@{\hspace{0.3cm}}c@{\hspace{0.3cm}}cc@{\hspace{0.3cm}}c@{\hspace{0.3cm}}cc}
        \toprule
        & \multicolumn{3}{c}{Llama} & \multicolumn{3}{c}{GPT-4} & \multicolumn{3}{c}{MPT} & \multicolumn{3}{c}{Mistral}  \\
        \cmidrule(r){2-4}\cmidrule(r){5-7}\cmidrule(r){8-10}\cmidrule(r){11-13}
   Method & recipe & poetry & review & recipe & poetry & review & recipe & poetry & review & recipe & poetry & review & Avg. \\
     \midrule
     
     \rowcolor{gray!20} \multicolumn{14}{c}{\textbf{Zero-Shot Methods}} \\
    NPR         & 0.92	&	0.50	&	0.94	&	0.73	&	0.50	&	0.76	&	0.56	&	0.53	&	0.54	&	0.58	&	0.53	&	0.84	&	0.66
\\
    LRR         & 0.91	&	0.81	&	0.90	&	0.78	&	0.60	&	0.74	&	0.57	&	0.53	&	0.56	&	0.66	&	0.61	&	0.86	&	0.71
 \\
    Rank        &0.51	&	0.53	&	0.54	&	0.51	&	0.53	&	0.54	&	0.50	&	0.53	&	0.50	&	0.50	&	0.53	&	0.54	&	0.52
\\
    Entropy     &0.78	&	0.47	&	0.61	&	0.76	&	0.49	&	0.60	&	0.29	&	0.65	&	0.62	&	0.55	&	0.68	&	0.65	&	0.60
\\
    LogRank     &0.79	&	0.81	&	0.79	&	0.80	&	0.64	&	0.78	&	0.30	&	0.63	&	0.44	&	0.43	&	0.77	&	0.78	&	0.66
\\
    Likelihood  &0.83	&	0.78	&	0.76	&	0.82	&	0.68	&	0.75	&	0.27	&	0.69	&	0.54	&	0.45	&	0.76	&	0.73	&	0.67
 \\
    
    Glimpse    & 0.93	&0.77	&0.95		&0.93	&0.60	&0.79		&0.70	&0.59	&0.75		&0.81	&0.67	&0.84		&0.78
\\
    Binoculars & \textbf{1.00}	&\textbf{0.98}	&0.95		&\textbf{0.99}	&0.81	&0.95		&0.43	&0.62	&0.68		&0.76	&0.72	&0.65		&0.80
\\
    DNAGPT      & 0.76	&	0.69	&	0.58	&	0.68	&	0.70	&	0.59	&	0.29	&	0.54	&	0.35	&	0.40	&	0.52	&	0.70	&	0.57
 \\ 
  DetectGPT   & 0.51	&	0.77	&	0.73	&	0.51	&	0.61	&	0.65	&	0.44	&	0.48	&	0.46	&	0.51	&	0.52	&	0.56	&	0.56
 \\
    F-DetectGPT     & \underline{0.92}	&	0.94	&	\underline{0.96}	&	0.96	&	0.79	&	0.80	&	0.39	&	0.63	&	0.41	&	0.48	&	0.79	&	0.64	&	0.73
\\

     \rowcolor{gray!20} \multicolumn{14}{c}{\textbf{Training-Based Methods}} \\
    R-QA  & 0.85	&	0.50	&	\underline{0.96}	&	0.76	&	0.50	&	0.83	&	0.46	&	0.53	&	0.69	&	0.44	&	0.55	&	0.69	&	0.65
 \\
    RADAR       & 0.58	&	0.59	&	0.86	&	0.63	&	0.57	&	0.86	&	0.59	&	0.73	&	0.59	&	0.64	&	\textbf{0.89}	&	0.63	&	0.68
 \\
     GPTZero     & 0.74	&	0.47	&	0.73	&	0.61	&	0.46	&	0.73	&	0.53	&	0.52	&	0.57	&	0.58	&	0.57	&	0.51	&	0.59
 \\ 
    DeTeCtive   & \textbf{1.00}	&	\underline{0.95}	&	0.94	&	\underline{0.97}	&	\underline{0.96}	&	\underline{0.97}	&	\underline{0.91}	&	\textbf{0.90}	&	\textbf{0.95}	&	\underline{0.87}	&	\underline{0.88}	&	\underline{0.90}	&	\underline{0.93}
\\
    \midrule
    {\ours}     & \textbf{1.00}	&	\textbf{0.98}	&	\textbf{0.97}	&	\textbf{0.99}	&	\textbf{1.00}	&	\textbf{1.00}	&	\textbf{0.95}	&	\underline{0.84}	&	\underline{0.90}	&	\textbf{0.94}	&	\underline{0.88}	&	\textbf{0.92}	&	\textbf{0.95}
\\
     
    \bottomrule
    \end{tabular}
    }   
\label{tab:humanraid_acc}
\end{table*}

\begin{table*}[!h]
\caption{MGT detection ACC performance comparisons with SOTA methods on HGT and MGT on Yelp, Essay, and Creative dataset. 
The best results are shown in bold font. The second-best results are shown in underlined. F-DetectGPT and R-QA are short for Fast-DetectGPT and RoBERTa-QA. } 
\centering
\setlength\tabcolsep{3.3pt}
\scalebox{1.0}{
    \begin{tabular}{lcccccccccc}
        \toprule
        & \multicolumn{3}{c}{Claude3-Sonnet} & \multicolumn{3}{c}{Claude3-Opus} & \multicolumn{3}{c}{Gemini}  \\
        \cmidrule(r){2-4}\cmidrule(r){5-7}\cmidrule(r){8-10}
   Method & Yelp & Essay & Creative & Yelp & Essay & Creative & Yelp & Essay & Creative & Avg. \\
     \midrule
     
     \rowcolor{gray!20} \multicolumn{11}{c}{\textbf{Zero-Shot Methods}} \\
    NPR         &0.62	&	0.67	&	0.80	&	0.62	&	0.58	&	0.68	&	0.50	&	0.57	&	0.56	&	0.62

\\
    LRR         & 0.55	&	0.90	&	0.78	&	0.52	&	0.91	&	0.73	&	0.45	&	0.58	&	0.56	&	0.66

 \\
    Rank        &0.51	&	0.54	&	0.51	&	0.50	&	0.55	&	0.51	&	0.50	&	0.55	&	0.51	&	0.52

\\
    Entropy     &0.60	&	0.87	&	0.69	&	0.58	&	0.92	&	0.71	&	0.53	&	0.85	&	0.53	&	0.70

\\
    LogRank     &0.57	&	0.91	&	0.79	&	0.54	&	0.93	&	0.89	&	0.54	&	0.94	&	0.68	&	0.75

\\
    Likelihood  &0.61	&	0.96	&	0.83	&	0.61	&	{0.97}	&	0.91	&	0.56	&	0.97	&	0.69	&	0.79

 \\
  Glimpse    & 0.69	&\textbf{1.00} &0.86 &0.69	&0.97 &0.90            &0.59			&0.96			&0.74								&0.82
\\
   Binoculars & 0.69	&\textbf{1.00} &0.94  &0.77	&\textbf{1.00}  &0.97            &0.68			&\underline{0.97}			&\textbf{0.78}								&0.87
\\
    DNAGPT      & 0.54	&	0.66	&	0.66	&	0.54	&	0.71	&	0.67	&	0.53	&	0.77	&	0.64	&	0.64

 \\  
    DetectGPT   & 0.49	&	0.68	&	0.69	&	0.44	&	0.62	&	0.69	&	0.42	&	0.66	&	0.62	&	0.59

 \\
    F-DetectGPT     & 0.66	&	\textbf{1.00}	&	0.88	&	0.72	&	\underline{0.99}	&	0.93	&	0.60	&	\textbf{0.98}	&	{0.69}	&	0.83

\\

     \rowcolor{gray!20} \multicolumn{11}{c}{\textbf{Training-Based Methods}} \\
    R-QA  & 0.72	&	0.86	&	0.79	&	0.82	&	0.87	&	0.93	&	0.81	&	0.86	&	0.72	&	0.82

 \\
    RADAR       & 0.62	&	0.94	&	0.84	&	0.64	&	0.95	&	0.91	&	0.64	&	0.96	&	0.74	&	0.80

 \\
    GPTZero     & 0.63	&	0.66	&	0.78	&	0.61	&	0.65	&	0.86	&	0.59	&	0.36	&	0.66	&	0.64

 \\ 
    DeTeCtive   & \underline{0.98}	&	0.86	&	\underline{0.97}	&	\underline{0.99}	&	0.79	&	\underline{0.96}	&	\underline{0.97}	&	0.85	&	\underline{0.77}	&	\underline{0.90}

\\
    \midrule
    {\ours}     & \textbf{0.99}	&	\underline{0.99}	&	\textbf{0.98}	&	\textbf{1.00}	&	\underline{0.99}	&	\textbf{0.98}	&	\textbf{0.99}	&	\underline{0.97}	&	\underline{0.77}	&	\textbf{0.96} \\   
    \bottomrule
    \end{tabular}
    }   

\label{tab:humanyelp_acc}
\end{table*}

%% file: experiments/nips/detailed-llm-auc.tex
\begin{table*}[!h]
\caption{MGT detection AUC performance comparisons with SOTA methods on HGT and MGT on M4 dataset. 
The best results are shown in bold font. The second-best results are shown in underlined. F-DetectGPT and R-QA are short for Fast-DetectGPT and RoBERTa-QA. } 
\centering
\setlength\tabcolsep{3.3pt}
\scalebox{0.85}{
    \begin{tabular}{lc@{\hspace{0.3cm}}c@{\hspace{0.3cm}}cc@{\hspace{0.3cm}}c@{\hspace{0.3cm}}cc@{\hspace{0.3cm}}c@{\hspace{0.3cm}}cc@{\hspace{0.3cm}}c@{\hspace{0.3cm}}cc}
        \toprule
        & \multicolumn{3}{c}{DaVinci} & \multicolumn{3}{c}{Cohere} & \multicolumn{3}{c}{Dolly} & \multicolumn{3}{c}{BloomZ}  \\
        \cmidrule(r){2-4}\cmidrule(r){5-7}\cmidrule(r){8-10}\cmidrule(r){11-13}
   Method & reddit & peerread & arxiv & reddit & peerread & arxiv & reddit & peerread & arxiv & reddit & peerread & arxiv & Avg. \\
     \midrule
     
     \rowcolor{gray!20} \multicolumn{14}{c}{\textbf{Zero-Shot Methods}} \\
    NPR         & {0.98}	&	\textbf{1.00}	&	0.28	&	0.97	&	\textbf{1.00}	&	0.97	&	0.86	&	0.97	&	0.80	&	0.86	&	0.97	&	0.86	&	0.88
 \\
    LRR         & 0.97	&	\textbf{1.00}	&	0.36	&	0.97	&	\textbf{1.00}	&	0.97	&	0.98	&	0.92	&	0.74	&	\underline{0.98}	&	\textbf{1.00}	&	\underline{0.93}	&	0.90
 \\
    Rank        &0.92	&	0.94	&	0.45	&	0.90	&	0.82	&	0.81	&	0.72	&	0.50	&	0.69	&	0.88	&	0.72	&	0.88	&	0.77
\\
    Entropy     &0.86	&	0.58	&	0.23	&	0.76	&	0.61	&	0.58	&	0.76	&	0.51	&	0.61	&	0.83	&	0.49	&	0.69	&	0.63
\\
    LogRank     &{0.98}	&	0.96	&	0.28	&	0.97	&	0.90	&	0.97	&	0.93	&	0.65	&	0.79	&	0.84	&	0.58	&	0.85	&	0.81
\\
    Likelihood  &{0.98}	&	0.83	&	0.27	&	0.96	&	0.78	&	0.96	&	0.93	&	0.60	&	0.80	&	0.70	&	0.47	&	0.78	&	0.76
\\
    Glimpse    & 0.92	&\textbf{1.00}	&0.51		&0.98	&0.96	&\textbf{1.00}		&0.83	&0.81	&0.91		&0.66	&0.42	&\textbf{0.98}		&0.83
\\
    Binoculars & \textbf{1.00}	&\textbf{1.00}	&0.51		&0.98	&\textbf{1.00}	&\textbf{1.00}		&\underline{0.98}	&\textbf{1.00}	&0.95		&0.53	&0.66	&0.85		&0.87
\\
    DNAGPT      & 0.84	&	0.27	&	0.32	&	0.94	&	0.35	&	0.93	&	0.72	&	0.55	&	0.70	&	0.47	&	0.11	&	0.69	&	0.57
 \\ 
    DetectGPT   & 0.59	&	0.74	&	0.29	&	0.72	&	0.76	&	0.43	&	0.61	&	0.54	&	0.42	&	0.73	&	0.71	&	0.63	&	0.60
\\
    F-DetectGPT     & \underline{0.99}	&	\textbf{1.00}	&	0.48	&	\underline{0.99}	&	\textbf{1.00}	&	\underline{0.99}	&	0.97	&	\textbf{1.00}	&	0.90	&	0.37	&	0.52	&	0.75	&	0.83
 \\

     \rowcolor{gray!20} \multicolumn{14}{c}{\textbf{Training-Based Methods}} \\
    
    R-QA  & \underline{0.99}	&	\textbf{1.00}	&	0.94	&	\underline{0.99}	&	\textbf{1.00}	&	\textbf{1.00}	&	\underline{0.98}	&	\underline{0.95}	&	\underline{0.99}	&	0.61	&	0.38	&	0.66	&	0.87
 \\
    RADAR       & 0.95	&	\textbf{1.00}	&	0.48	&	0.97	&	\textbf{1.00}	&	0.78	&	0.79	&	0.92	&	0.43	&	0.90	&	0.88	&	0.52	&	0.80
 \\
     GPTZero     & 0.86	&	0.99	&	0.36	&	0.84	&	\underline{0.92}	&	0.65	&	0.76	&	0.58	&	0.50	&	0.61	&	0.53	&	0.46	&	0.67
 \\ 
    DeTeCtive   & 0.96	&	0.85	&	\textbf{0.98}	&	0.89	&	0.88	&	0.98	&	0.96	&	0.86	&	\textbf{1.00}	&	0.96	&	\underline{0.96}	&	\textbf{0.98}	&	\underline{0.94}
\\
    \midrule
    {\ours}     & \underline{0.99}	&	\textbf{1.00}	&	\underline{0.94}	&	\textbf{1.00}	&	\textbf{1.00}	&	0.98	&	\textbf{0.99}	&	\textbf{1.00}	&	0.85	&	\textbf{1.00}	&	\textbf{1.00}	&	\textbf{0.98}	&	\textbf{0.98}
\\
     
    \bottomrule
    \end{tabular}
    }   
\label{tab:humanm4_auc}
\end{table*}

\begin{table*}[!h]
\caption{MGT detection AUC performance comparisons with SOTA methods on HGT and MGT on RAID dataset. 
The best results are shown in bold font. The second-best results are shown in underlined. F-DetectGPT and R-QA are short for Fast-DetectGPT and RoBERTa-QA. } 
\centering
\setlength\tabcolsep{3.3pt}
\scalebox{0.9}{
    \begin{tabular}{lc@{\hspace{0.15cm}}c@{\hspace{0.15cm}}ccccccccccc}
        \toprule
        & \multicolumn{3}{c}{Llama} & \multicolumn{3}{c}{GPT-4} & \multicolumn{3}{c}{MPT} & \multicolumn{3}{c}{Mistral}  \\
        \cmidrule(r){2-4}\cmidrule(r){5-7}\cmidrule(r){8-10}\cmidrule(r){11-13}
   Method & recipe & poetry & review & recipe & poetry & review & recipe & poetry & review & recipe & poetry & review & Avg. \\
     \midrule
     
     \rowcolor{gray!20} \multicolumn{14}{c}{\textbf{Zero-Shot Methods}} \\
    NPR         &\underline{0.99}	&	0.87	&	\underline{0.98}	&	0.97	&	0.70	&	0.93	&	0.39	&	0.74	&	0.61	&	0.64	&	0.82	&	0.74	&	0.78\\
    LRR         &0.98	&	0.88	&	\underline{0.98}	&	0.94	&	0.60	&	0.83	&	0.44	&	0.83	&	0.84	&	0.62	&	0.89	&	0.84	&	0.81 \\
    Rank        &0.88	&	0.79	&	0.97	&	0.67	&	0.65	&	0.87	&	0.45	&	0.92	&	0.83	&	0.45	&	0.93	&	0.90	&	0.78 \\
    Entropy     &0.94	&	0.63	&	0.92	&	0.91	&	0.59	&	0.77	&	0.35	&	0.72	&	0.61	&	0.59	&	0.80	&	0.71	&	0.71 \\
    LogRank     &0.99	&	0.87	&	\underline{0.98}	&	0.97	&	0.69	&	0.94	&	0.38	&	0.73	&	0.60	&	0.64	&	0.81	&	0.74	&	0.78\\
    Likelihood  &\underline{0.99}	&	0.86	&	\underline{0.98}	&	0.98	&	0.72	&	0.95	&	0.38	&	0.67	&	0.53	&	0.64	&	0.80	&	0.71	&	0.77 \\

    Glimpse    & \textbf{1.00}	& 0.87	& 0.97		& \underline{0.99}	& 0.60	& 0.88		& 0.69	& 0.63	& 0.84		& 0.86	& 0.74	& 0.92		& 0.83
\\
 Binoculars & \underline{0.99}	& \underline{0.99}	& 0.97		& \textbf{1.00}	& \underline{0.98}	& \textbf{0.99}		& 0.55	& 0.66	& 0.59		& 0.72	& 0.79	& 0.68		& 0.83
\\
DNAGPT      & 0.96	&	0.75	&	0.95	&	0.80	&	0.75	&	0.88	&	0.38	&	0.55	&	0.49	&	0.57	&	0.72	&	0.60	&	0.70 \\ 
    
    DetectGPT   & 0.52	&	0.82	&	0.83	&	0.55	&	0.63	&	0.75	&	0.29	&	0.45	&	0.45	&	0.48	&	0.45	&	0.55	&	0.56 \\
    F-DetectGPT     &\underline{0.99}	&	\underline{0.97}	&	0.97	&	\underline{0.99}	&	0.88	&	\underline{0.99}	&	0.50	&	0.61	&	0.51	&	0.70	&	0.77	&	0.65	&	0.79\\

     \rowcolor{gray!20} \multicolumn{14}{c}{\textbf{Training-Based Methods}} \\
    R-QA  & 0.95	&	0.94	&	0.96	&	0.82	&	0.83	&	0.95	&	0.45	&	0.73	&	0.63	&	0.31	&	0.65	&	0.54	&	0.73 \\
    RADAR       & 0.98	&	0.85	&	0.89	&	\underline{0.99}	&	0.81	&	0.87	&	\underline{0.95}	&	0.83	&	0.74	&	0.80	&	0.86	&	0.63	&	0.85 \\
    
    GPTZero     & 0.74	&	0.47	&	0.73	&	0.61	&	0.46	&	0.73	&	0.53	&	0.52	&	0.57	&	0.58	&	0.57	&	0.51	&	0.59 \\ 
    DeTeCtive   & \textbf{1.00}	&	0.95	&	0.96	&\underline{0.99}	&{0.96}	&	\underline{0.99}	&	0.92	&	\textbf{0.91}	&	\textbf{0.99}	&	\underline{0.91}	&	\underline{0.90}	&	\underline{0.97}	&	\underline{0.95}\\
    \midrule
    {\ours}     & \textbf{1.00}	&	\textbf{1.00}	&	\textbf{1.00}	&	\textbf{1.00}	&	\textbf{1.00}	&	\textbf{1.00}	&	\textbf{0.99}	&	\underline{0.90}	&	\underline{0.95}	&	\textbf{0.99}	&	\textbf{0.94}	&	\textbf{0.98}	&	\textbf{0.98}\\
    \bottomrule
    \end{tabular}
    }   

\label{tab:humanraid_auc}
\end{table*}

\begin{table*}[!h]
\caption{MGT detection ACC performance comparisons with SOTA methods on HGT and MGT on Yelp, Essay, and Creative dataset. 
The best results are shown in bold font. The second-best results are shown in underlined. F-DetectGPT and R-QA are short for Fast-DetectGPT and RoBERTa-QA. } 
\centering
\scalebox{1.0}{
    \begin{tabular}{lc@{\hspace{0.4cm}}c@{\hspace{0.15cm}}cc@{\hspace{0.4cm}}c@{\hspace{0.15cm}}cc@{\hspace{0.4cm}}c@{\hspace{0.15cm}}cc}
        \toprule
        & \multicolumn{3}{c}{Claude3-Sonnet} & \multicolumn{3}{c}{Claude3-Opus} & \multicolumn{3}{c}{Gemini}  \\
        \cmidrule(r){2-4}\cmidrule(r){5-7}\cmidrule(r){8-10}
   Method & Yelp & Essay & Creative & Yelp & Essay & Creative & Yelp & Essay & Creative & Avg. \\
     \midrule
     
     \rowcolor{gray!20} \multicolumn{11}{c}{\textbf{Zero-Shot Methods}} \\
     NPR         &0.68	&	\underline{0.99}	&	0.94	&	0.66	&	0.99	&	0.98	&	0.49	&	0.98	&	0.76	&	0.83 \\
    LRR         &0.54	&	\textbf{1.00}	&	0.88	&	0.52	&	\textbf{1.00}	&	0.95	&	0.39	&	\textbf{0.99}	&	0.70	&	0.77 \\
    
    Rank        &0.54	&	0.85	&	0.85	&	0.49	&	\underline{0.99}	&	0.92	&	0.39	&	\underline{0.97}	&	0.65	&	0.74 \\
    Entropy     &0.64	&	0.83	&	0.83	&	0.57	&	0.95	&	0.88	&	0.42	&	0.91	&	0.57	&	0.73 \\
    LogRank     &0.69	&	0.93	&	0.93	&	0.68	&	\textbf{1.00}	&	0.98	&	0.50	&	\textbf{0.99}	&	0.74	&	0.83 \\
    Likelihood  &0.73	&	0.94	&	0.94	&	0.72	&	\textbf{1.00}	&	\textbf{0.99}	&	0.55	&	\textbf{0.99}	&	0.76	&	0.85 \\
    Glimpse    & 0.78    &\textbf{1.00}  &0.90                &0.83  &\textbf{1.00}   &0.96            &0.74   &\textbf{1.00}      &0.78  &0.89 \\
    Binoculars & 0.79   & \textbf{1.00}   & \underline{0.99}  & 0.87 & \textbf{1.00}  & \textbf{1.00}  & 0.73  & \underline{0.99}  & 0.79  & \underline{0.91}\\
    DNAGPT      &0.67	&	0.94	&	0.86	&	0.70	&	0.94	&	0.93	&	0.58	&	0.95	&	0.75	&	0.81 \\ 
    
    DetectGPT   & 0.53	&	0.75	&	0.71	&	0.43	&	0.74	&	0.78	&	0.37	&	0.80	&	0.64	&	0.64\\
    F-DetectGPT     &0.73	&	\textbf{1.00}	&	0.94	&	0.81	&	\textbf{1.00}	&	\textbf{0.99}	&	0.68	&	\textbf{0.99}	&	\textbf{0.79}	&	0.88 \\
    
     \rowcolor{gray!20} \multicolumn{11}{c}{\textbf{Training-Based Methods}} \\
    R-QA  &0.92	&	0.95	&	0.94	&	0.96	&	0.98	&	0.97	&	0.96	&	0.94	&	\underline{0.78}	&	0.93\\
    RADAR       &0.58	&	0.93	&	\underline{0.99}	&	0.68	&	\underline{0.99}	&	0.97	&	0.70	&	\textbf{0.99}	&	0.76	&	0.84\\
    GPTZero     &0.63	&	0.66	&	0.78	&	0.61	&	0.65	&	0.86	&	0.59	&	0.36	&	0.66	&	0.64\\ 
    DeTeCtive   & \underline{0.98}	&	0.86	&	0.96	&	\underline{0.99}	&	0.79	&	\textbf{0.99}	&	\underline{0.99}	&	0.85	&	0.76	&	\underline{0.91}\\
    \midrule
    {\ours}     & \textbf{1.00}	&	\textbf{1.00}	&	\textbf{1.00}	&	\textbf{1.00}	&	\textbf{1.00}	&	\textbf{0.99}	&	\textbf{1.00}	&	\textbf{0.99}	&	\underline{0.78}	&	\textbf{0.97} \\   
    \bottomrule
    \end{tabular}
    }  

\label{tab:humanyelp_auc}
\end{table*}

%% file: experiments/statistation_hc3.tex
\begin{table}[t]
\caption{Statistical significance analysis on HC3 dataset. We repeat the experiments 5 times and report the mean and standard deviation. } 
\centering
\setlength\tabcolsep{3.3pt}
\scalebox{1.0}{
    \begin{tabular}{lcccc}
        \toprule
        Metric & open-qa & wiki-csai \\ 
     \midrule
     \multirow{1}{*}{ACC}   &  \ms{0.9690}{0.0073}	& \ms{0.9410}{0.0097}	\\
     \multirow{1}{*}{AUC}  &  \ms{0.9965}{0.0005}	& \ms{0.9938}{0.0005}	 \\
     \hline
     \hline
    Metric &  medicine & finance  \\
    \multirow{1}{*}{ACC}    & \ms{0.9750}{0.0032}   & \ms{0.9810}{0.0037} \\
    \multirow{1}{*}{AUC}  & \ms{0.9993}{0.0001}   & \ms{0.9983}{0.0004} \\  
    \bottomrule
    \end{tabular}
    }   
\label{tab:statistic_model}
\end{table}

%% file: experiments/ablation_tokenizer.tex
\begin{table}[h]
\caption{MGT detection performance comparison on HC3 dataset between default(Bert) and GPT2 tokenizers. } 
\vspace{-0.4em}
\centering
\setlength\tabcolsep{3.3pt}
\scalebox{1.0}{
    \begin{tabular}{lcccccccc}
        \toprule
     & \multicolumn{2}{c}{open-qa} 
       & \multicolumn{2}{c}{wiki-csai} \\
       \cmidrule(r){2-3} \cmidrule(r){4-5}
    Metric   & Bert & GPT2    & Bert & GPT2    \\
     \midrule
     \multirow{1}{*}{ACC}   &   0.97	& 0.99	& 0.96	& 0.95	 \\
     \multirow{1}{*}{AUC} &   1.00	& 1.00	& 0.99	& 1.00	 \\
    \hline
    \hline
       & \multicolumn{2}{c}{medicine} 
       & \multicolumn{2}{c}{finance} \\
       \cmidrule(r){2-3} \cmidrule(r){4-5} 
    Metric   & Bert & GPT2    & Bert & GPT2   \\
     \midrule
     \multirow{1}{*}{ACC}  	& 0.98	& 0.98	& 0.98	& 0.97	 \\
     \multirow{1}{*}{AUC}    & 1.00	& 0.99	& 1.00	& 1.00	 \\

    \bottomrule
    \end{tabular}
    }   
\label{tab:tokenizers}
\vspace{-0.4em}
\end{table}

%% file: experiments/ablation_graph_hc3.tex
\begin{table}[!h]
\caption{Ablation analysis on HC3 dataset. The best results are shown in bold font.} 
\centering
\setlength\tabcolsep{3.3pt}
\scalebox{0.75}{
    \begin{tabular}{llccccc}
        \toprule
        & Method & open-qa & wiki-csai & medicine & finance & Avg.  \\
     \midrule
     \multirow{4}{*}{\rotatebox[origin=c]{90}{ACC} } & {\ours}       &  0.97	&	0.96	&	0.98	&	\textbf{0.98}	&	\textbf{0.97}  \\
                            & {\ours-U}   &  0.95	&	0.94	&	\textbf{1.00}	&	\textbf{0.98}	&	\textbf{0.97} \\
                            & {\ours-W}   &  \textbf{1.00}	&	0.84	&	\textbf{1.00}	&	0.94	&	0.95 \\
                            & {\ours-UW}   &  0.98	&	0.79	&	\textbf{1.00}	&	0.93	&	0.92 \\
                            & {{\ours-Bert}}   &  {\textbf{1.00}}	&	{0.79}	&	{0.98} &	{0.89}	&	{0.91} \\
     \midrule
     \multirow{4}{*}{\rotatebox[origin=c]{90}{AUC}} & {\ours}     & \textbf{1.00}	&	0.99	&	\textbf{1.00}	&	\textbf{1.00}	&	\textbf{1.00} \\
                            & {\ours-U}   &  0.99	&	\textbf{1.00}	&	\textbf{1.00}	&	\textbf{1.00}	&	\textbf{1.00} \\
                            & {\ours-W}   &  \textbf{1.00}	&	0.84	&	\textbf{1.00}	&	0.98	&	0.96 \\
                            & {\ours-UW}   & \textbf{ 1.00}	&	0.86	&	\textbf{1.00}	&	0.97	&	0.96 \\
                            & {{\ours-Bert}}   &  {\textbf{1.00}}	&	{0.85}	& {0.99}	&	{0.97}	&	{0.95} \\
     
    \bottomrule
    \end{tabular}
    }   
\label{tab:model_ablation}
\end{table}

%% file: experiments/time_comsumption.tex
\begin{table*}[!h]
\caption{Inference time(seconds) comparison on HC3 dataset. We repeat the experiments 10 times and report the average time consumption. - indicates the inference time is more than 10 minutes. The best results are shown in bold font.} 
\centering
\scalebox{1.0}{
    \begin{tabular}{lrrrr}
        \toprule
         & open-qa & wiki-csai & medicine & finance \\
     \midrule
     NPR       &  - & - &- & -  \\
     DNA-GPT   &  - & - &- & -  \\  
     DetectGPT &  442.0000 & 161.6530 &82.2744	 &255.4350  \\
     Fast-DetectGPT  &  28.0217	& 27.0673	 & 24.1440	&28.4082  \\
    RoBERTa-QA   &  2.9267	&2.5464	&2.5391	&2.5413   \\
    DeTeCtive &  18.7223	&13.2559	&17.2883	&17.5105 \\
    \midrule
    {\ours}      &  \textbf{0.0091}	& \textbf{0.0065} & \textbf{0.0051}	& \textbf{0.0058} \\     
    \bottomrule
    \end{tabular}
    }   
\label{tab:inference_time}
\vspace{-0.2cm}
\end{table*}

%% file: experiments/ablation_windowsize.tex
\begin{table}[t]
\caption{{Sliding window size ablation analysis on HC3 dataset. The best results are shown in bold font.}} 
\centering
\setlength\tabcolsep{3.3pt}
\scalebox{0.85}{
    \begin{tabular}{lcccccc}
        \toprule
        & {Method} & {open-qa} & {wiki-csai} & {medicine} & {finance} & {Avg.}  \\
     \midrule
     \multirow{4}{*}{\rotatebox[origin=c]{90}{{ACC}} } 
                            & {10}   &  {0.95}	&	{0.93}	&	{0.99}	&	{0.93}	&	{0.95}  \\
                            & {15}   &  {0.97}	&	{0.95}	&	{0.99}	&	{0.94}	&	{0.96} \\
                            & {20}   &  {0.97}	&	{\textbf{0.96}}	&	{0.98}	&	{\textbf{0.98}}	&	{\textbf{0.97}} \\
                            & {25}   &  {0.96}	&	{0.94}	&	{\textbf{1.00}}	&	{0.94}	&	{0.96} \\
                            & {30}   &  {\textbf{0.98}}	&	{0.93}	&	{0.99}    &	{0.93}	&	{0.96} \\
     \midrule
     \multirow{4}{*}{\rotatebox[origin=c]{90}{{AUC}}} 
                            & {10}   & {0.99}	&	{0.99}	& {\textbf{1.00}}	&	{0.98}	&	{0.99} \\
                            & {15}   & {\textbf{1.00}}	&	{0.99}	& {\textbf{1.00}}	&	{0.98}	&	{0.99} \\
                            & {20}   & {\textbf{1.00}}	&	{0.99}	& {\textbf{1.00}}	&	{\textbf{1.00}}	&	{\textbf{1.00}} \\
                            & {25}   & {\textbf{1.00}}	&	{\textbf{1.00}}	& {\textbf{1.00}}	&	{0.98}	&	{\textbf{1.00}} \\
                            & {30}   & {\textbf{1.00}}	&	{\textbf{1.00}}	& {\textbf{1.00}}	&	{0.98}	&	{\textbf{1.00}} \\
     
    \bottomrule
    \end{tabular}
    }   
\label{tab:window_ablation}
\end{table}

%% file: experiments/morf_lerf.tex
\begin{figure*}[!h]
  \centering
  \begin{subfigure}[h]{0.185\textwidth}
        \includegraphics[width=\textwidth, trim=25 30 25 25, clip]{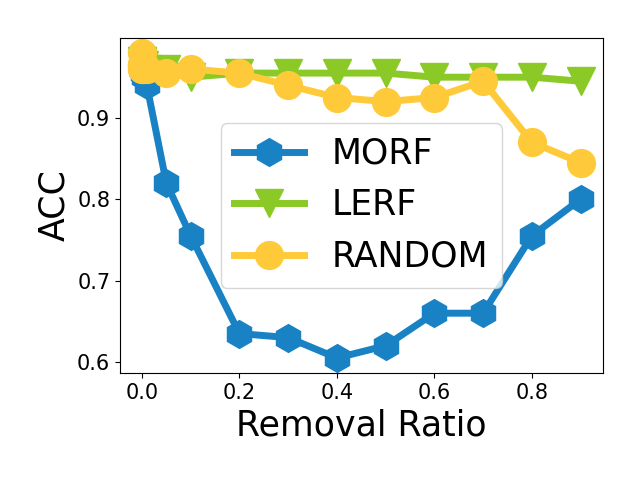}
        \caption{open-qa in HC3}
        \label{fig:explain_hc3_openqa}
  \end{subfigure}
  \begin{subfigure}[h]{0.185\textwidth}
        \includegraphics[width=\textwidth, trim=25 30 25 25, clip]{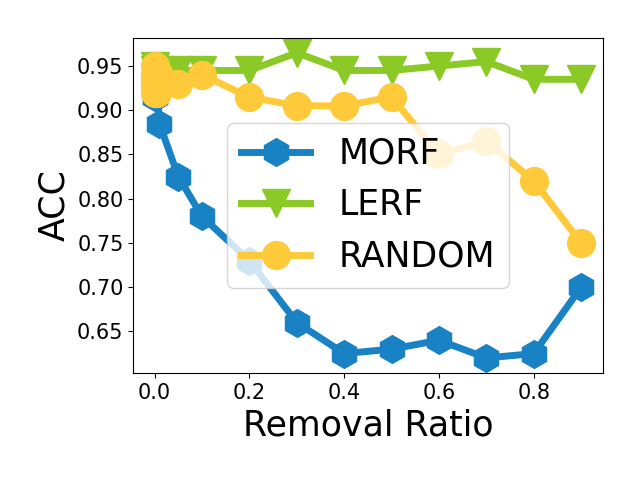}
        \caption{wiki-csai in HC3}
        \label{fig:explain_hc3_csai}
  \end{subfigure}
  \begin{subfigure}[h]{0.185\textwidth}
        \includegraphics[width=\textwidth, trim=25 30 25 25, clip]{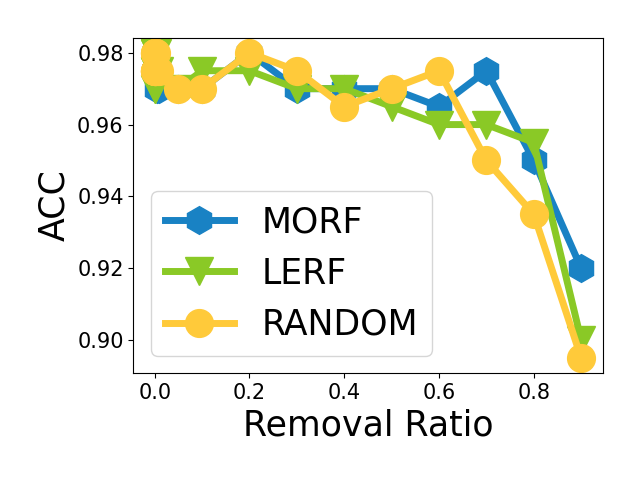}
        \caption{medicine in HC3}
        \label{fig:explain_hc3_med}
  \end{subfigure}
  \begin{subfigure}[h]{0.185\textwidth}
        \includegraphics[width=\textwidth, trim=25 30 25 25, clip]{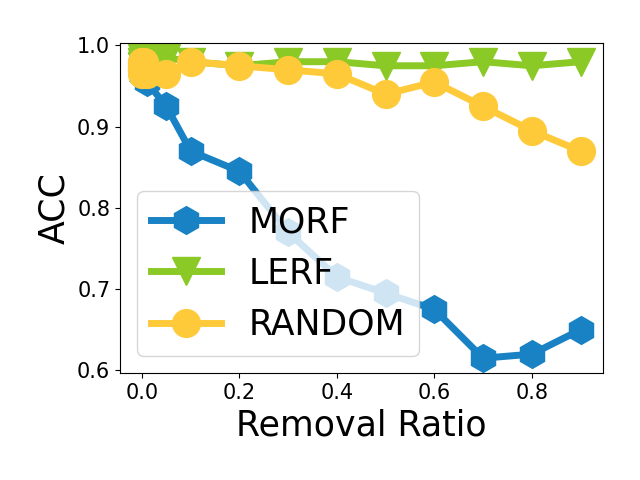}
        \caption{finance in HC3}
        \label{fig:explain_hc3_avg}
  \end{subfigure}
  \begin{subfigure}[h]{0.185\textwidth}
        \includegraphics[width=\textwidth, trim=25 30 25 25, clip]{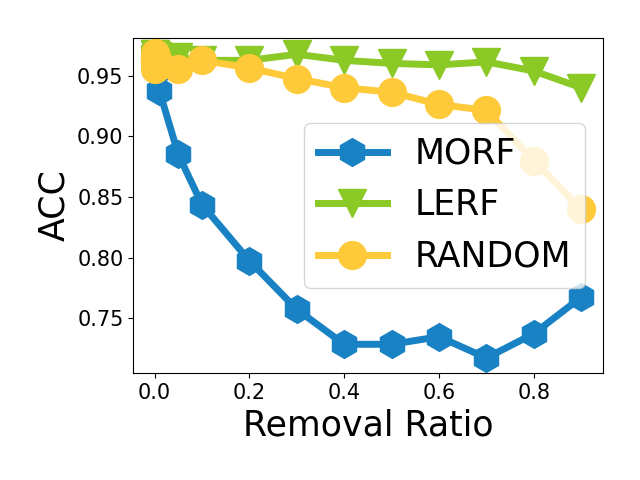}
        \caption{Average in HC3}
        \label{fig:explain_m4_avg}
  \end{subfigure} \\

  \begin{subfigure}[h]{0.185\textwidth}
        \includegraphics[width=\textwidth, trim=25 30 25 25, clip]{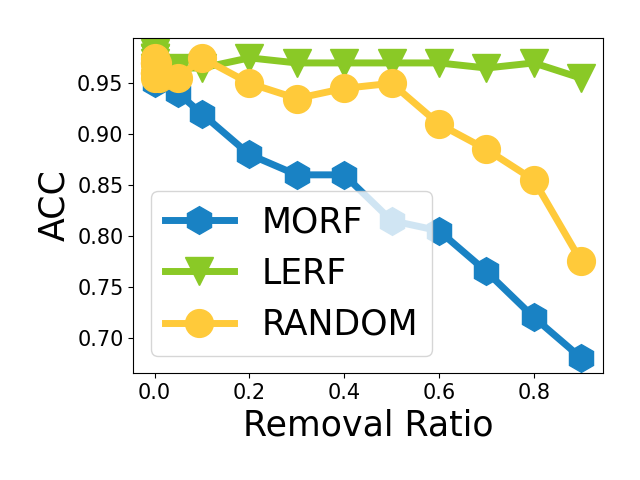}
        \caption{wikihow in M4}
        \label{fig:explain_m4_wikihow}
  \end{subfigure}
  \begin{subfigure}[h]{0.185\textwidth}
        \includegraphics[width=\textwidth, trim=25 30 25 25, clip]{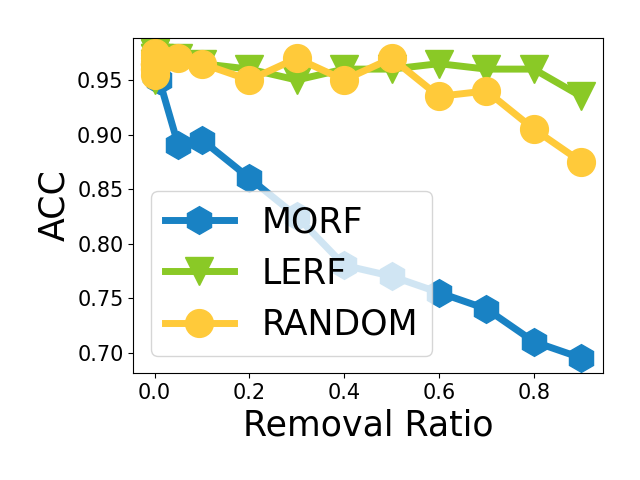}
        \caption{reddit in M4}
        \label{fig:explain_m4_reddit}
  \end{subfigure}
  \begin{subfigure}[h]{0.185\textwidth}
        \includegraphics[width=\textwidth, trim=25 30 25 25, clip]{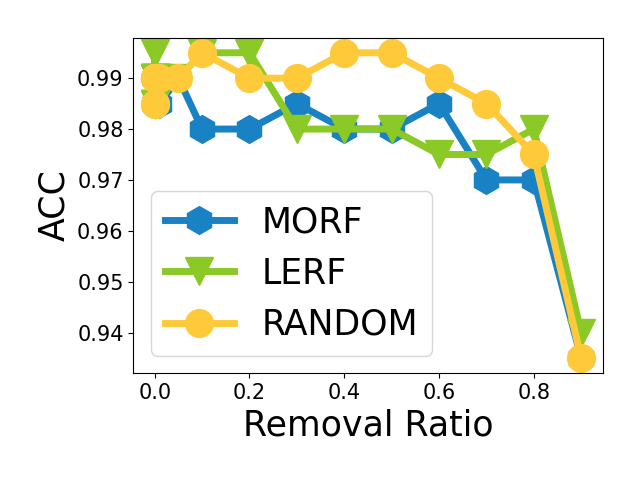}
        \caption{peerread in M4}
        \label{fig:explain_m4_peer}
  \end{subfigure}
  \begin{subfigure}[h]{0.185\textwidth}
        \includegraphics[width=\textwidth, trim=25 30 25 25, clip]{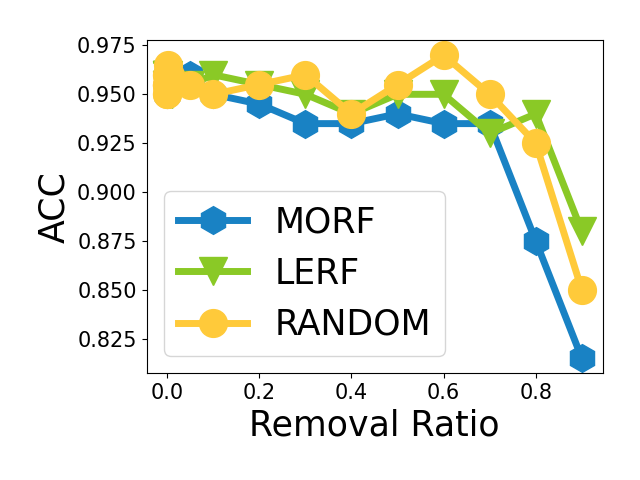}
        \caption{arxiv in M4}
        \label{fig:explain_m4_arxiv}
  \end{subfigure} 
  \begin{subfigure}[h]{0.185\textwidth}
        \includegraphics[width=\textwidth, trim=25 30 25 25, clip]{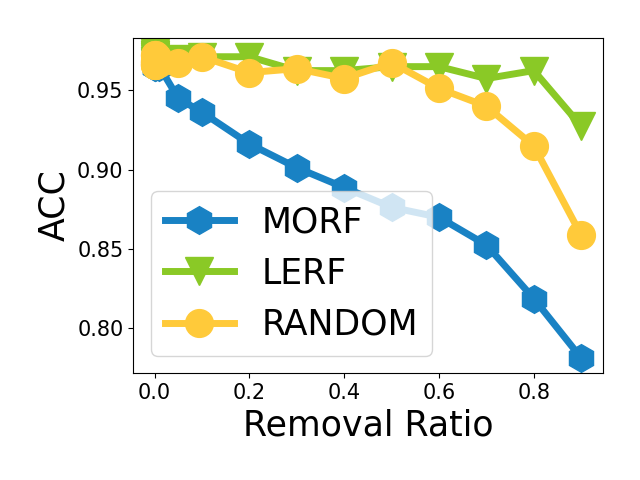}
        \caption{Average in M4}
        \label{fig:explain_raid_avg}
  \end{subfigure} \\

  \begin{subfigure}[h]{0.185\textwidth}
        \includegraphics[width=\textwidth, trim=25 30 25 25, clip]{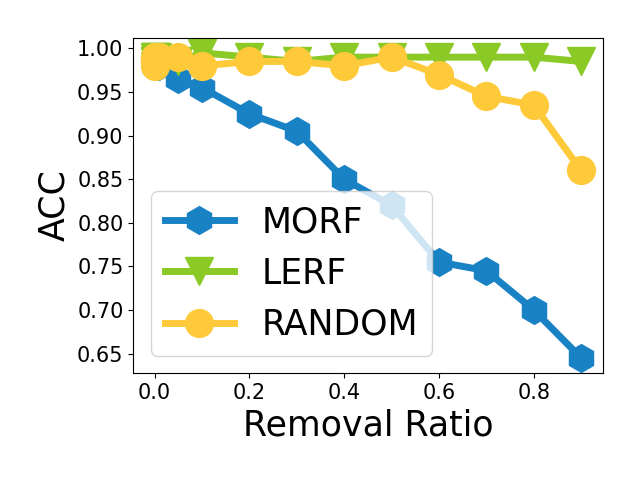}
        \caption{recipe in RAID}
        \label{fig:explain_RAID_recipe}
  \end{subfigure}
  \begin{subfigure}[h]{0.185\textwidth}
        \includegraphics[width=\textwidth, trim=25 30 25 25, clip]{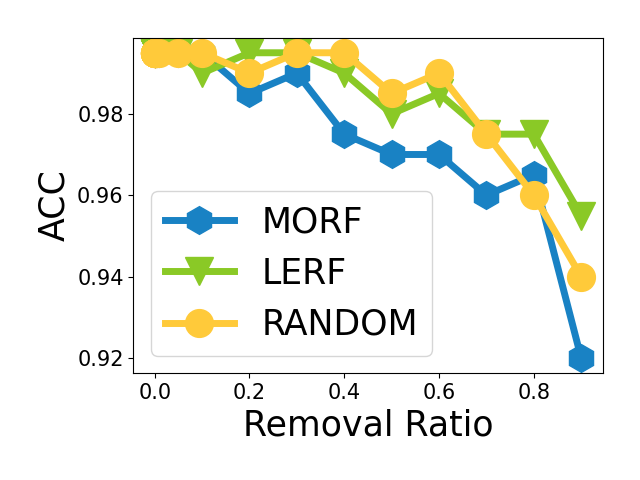}
        \caption{book in RAID}
        \label{fig:explain_RAID_book}
  \end{subfigure}
  \begin{subfigure}[h]{0.185\textwidth}
        \includegraphics[width=\textwidth, trim=25 30 25 25, clip]{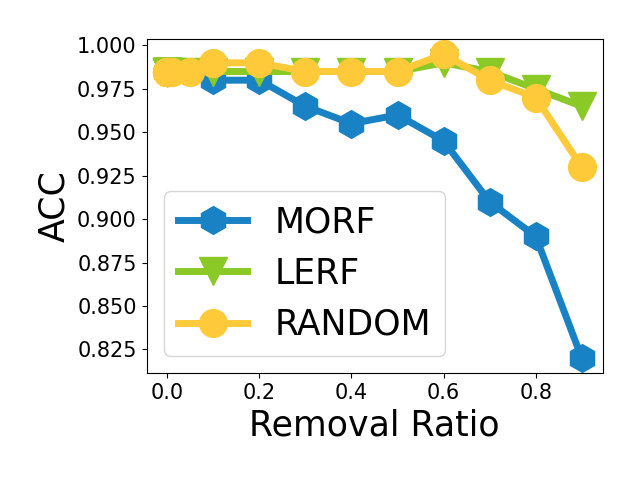}
        \caption{poetry in RAID}
        \label{fig:explain_RAID_poetry}
  \end{subfigure}
  \begin{subfigure}[h]{0.185\textwidth}
        \includegraphics[width=\textwidth, trim=25 30 25 25, clip]{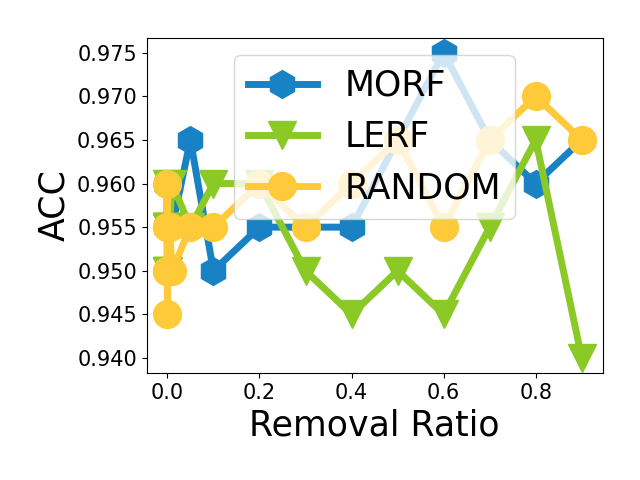}
        \caption{review in RAID}
        \label{fig:explain_RAID_review}
  \end{subfigure}
  \begin{subfigure}[h]{0.185\textwidth}
        \includegraphics[width=\textwidth, trim=25 30 25 25, clip]{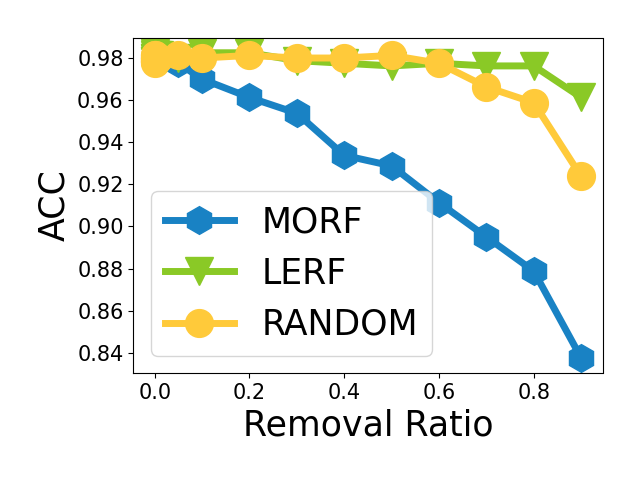}
        \caption{Average in RAID}
        \label{fig:explain_hc3_fin}
  \end{subfigure} 

\caption{Comparison results of MORF and LERF between explainable motifs extracted from {\ours} and random motifs on HGT and ChatGPT-generated texts.}
\label{fig:xai_motifs_evaluation}
\vspace{-0.5em}
\end{figure*}

%% file: experiments/figure_lerf_comparison.tex
\begin{figure*}[t]
  \centering
  \begin{subfigure}[h]{0.19\textwidth}
        \includegraphics[width=\textwidth, trim=25 30 25 25, clip]{figures/lerf_morf/lerf_open_qa.png}
        \caption{open-qa}
        \label{fig:baseline_openqa_lerf}
  \end{subfigure}
  \begin{subfigure}[h]{0.18\textwidth}
        \includegraphics[width=\textwidth, trim=50 30 25 25, clip]{figures/lerf_morf/lerf_wiki_csai.png}
        \caption{wiki-csai}
        \label{fig:baseline_csai_lerf}
  \end{subfigure}
  \begin{subfigure}[h]{0.18\textwidth}
        \includegraphics[width=\textwidth, trim=50 30 25 25, clip]{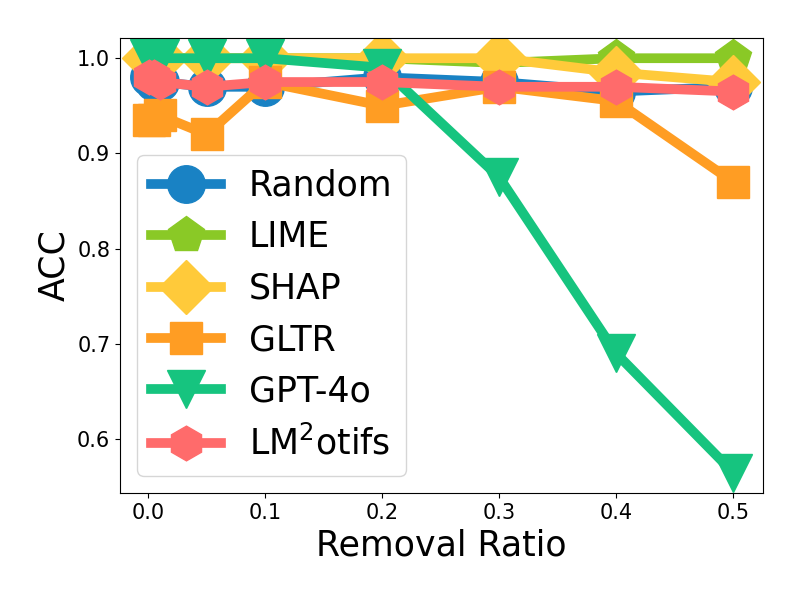}
        \caption{medicine}
        \label{fig:baseline_med_lerf}
  \end{subfigure}
  \begin{subfigure}[h]{0.18\textwidth}
        \includegraphics[width=\textwidth, trim=50 30 25 25, clip]{figures/lerf_morf/lerf_finance.png}
        \caption{finance}
        \label{fig:baseline_fin_lerf}
  \end{subfigure} 
  \begin{subfigure}[h]{0.18\textwidth}
        \includegraphics[width=\textwidth, trim=50 30 25 25, clip]{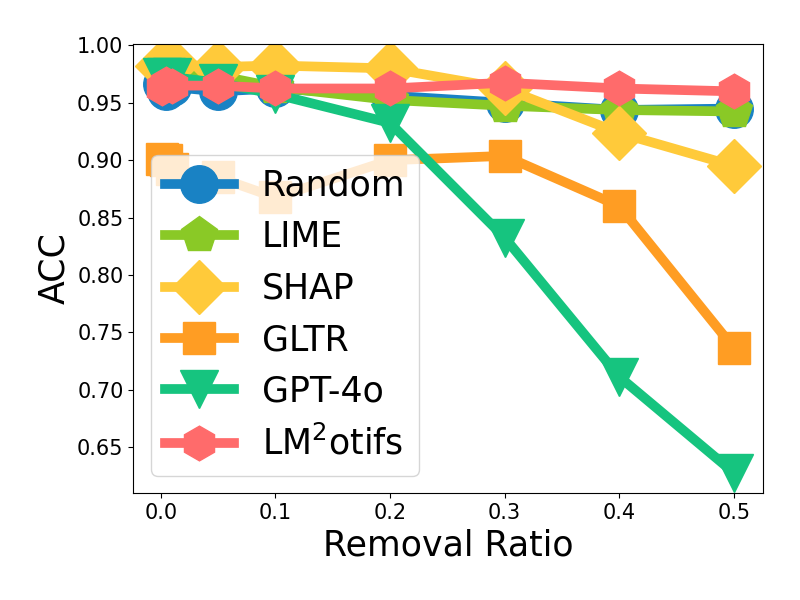}
        \caption{Average in HC3}
        \label{fig:appendix_baseline_avg_lerf}
  \end{subfigure} \\
\caption{Comparison results of LERF between {\ours} and adapted baselines.}
\label{fig:extensive_baseline_lerf}
\vspace{-0.5em}
\end{figure*}

%% file: experiments/figure_morf_comparison.tex
\begin{figure*}[!t]
  \centering
  \begin{subfigure}[h]{0.19\textwidth}
        \includegraphics[width=\textwidth, trim=25 30 25 25, clip]{figures/lerf_morf/morf_open_qa.png}
        \caption{open-qa}
        \label{fig:appendix_baseline_openqa_morf}
  \end{subfigure}
  \begin{subfigure}[h]{0.18\textwidth}
        \includegraphics[width=\textwidth, trim=50 30 25 25, clip]{figures/lerf_morf/morf_wiki_csai.png}
        \caption{wiki-csai}
        \label{fig:baseline_csai_morf}
  \end{subfigure}
  \begin{subfigure}[h]{0.18\textwidth}
        \includegraphics[width=\textwidth, trim=50 30 25 25, clip]{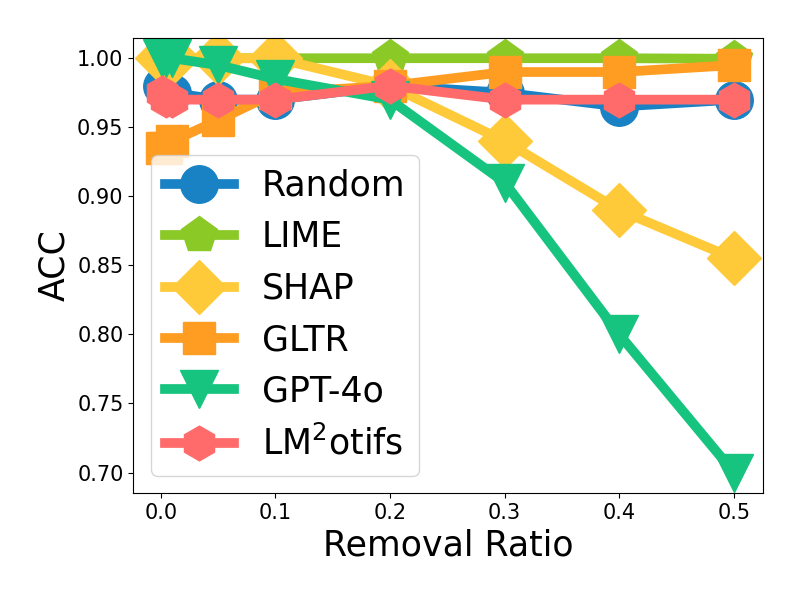}
        \caption{medicine}
        \label{fig:baseline_med_morf}
  \end{subfigure}
  \begin{subfigure}[h]{0.18\textwidth}
        \includegraphics[width=\textwidth, trim=50 30 25 25, clip]{figures/lerf_morf/morf_finance.png}
        \caption{finance}
        \label{fig:baseline_fin_morf}
  \end{subfigure} 
  \begin{subfigure}[h]{0.18\textwidth}
        \includegraphics[width=\textwidth, trim=50 30 25 25, clip]{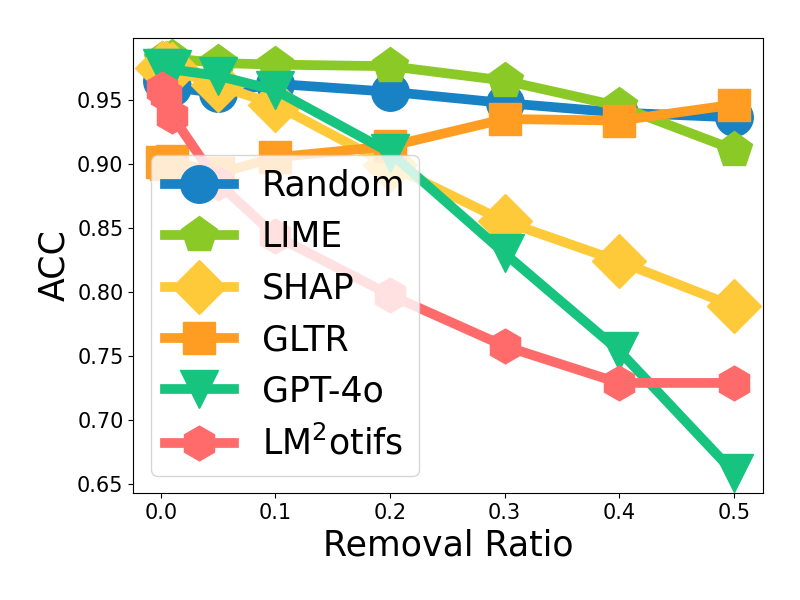}
        \caption{Average in HC3}
        \label{fig:baseline_avg_morf}
  \end{subfigure} \\
\caption{Comparison results of MORF between {\ours} and adapted baselines.}
\label{fig:extensive_baseline_morf}
\vspace{-0.5em}
\end{figure*}

%% file: experiments/figure_lambda.tex
\begin{figure*}[!h]
  \centering
  \begin{subfigure}[h]{0.19\textwidth}
        \includegraphics[width=\textwidth, trim=25 30 25 25, clip]{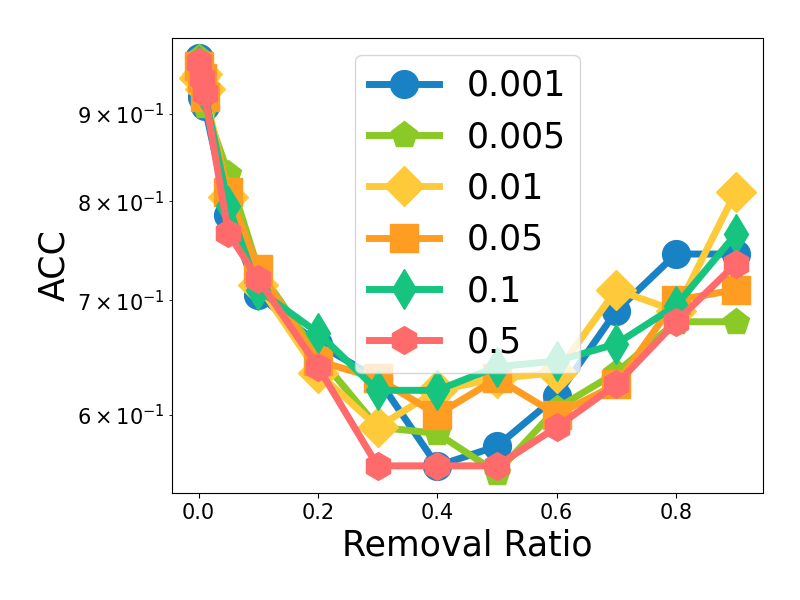}
        \caption{open-qa}
        \label{fig:appendix_lambda_hc3_openqa}
  \end{subfigure}
  \begin{subfigure}[h]{0.18\textwidth}
        \includegraphics[width=\textwidth, trim=50 30 25 25, clip]{figures/lambda/lambda_ablation_study_wiki_csai.png}
        \caption{wiki-csai}
        \label{fig:appendix_lambda_hc3_csai}
  \end{subfigure}
  \begin{subfigure}[h]{0.18\textwidth}
        \includegraphics[width=\textwidth, trim=50 30 25 25, clip]{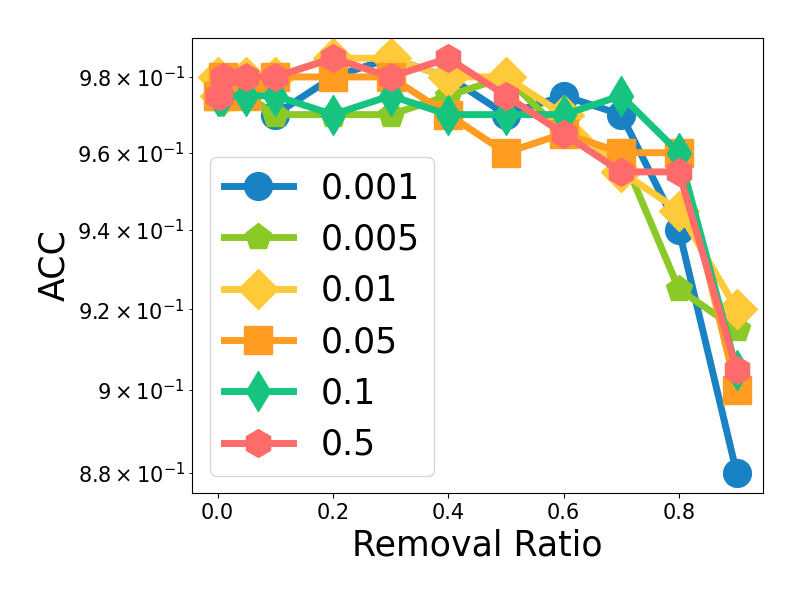}
        \caption{medicine}
        \label{fig:appendix_lambda_hc3_med}
  \end{subfigure}
  \begin{subfigure}[h]{0.18\textwidth}
        \includegraphics[width=\textwidth, trim=50 30 25 25, clip]{figures/lambda/lambda_ablation_study_finance.png}
        \caption{finance}
        \label{fig:appendix_lambda_hc3_fin}
  \end{subfigure} 
  \begin{subfigure}[h]{0.18\textwidth}
        \includegraphics[width=\textwidth, trim=50 30 25 25, clip]{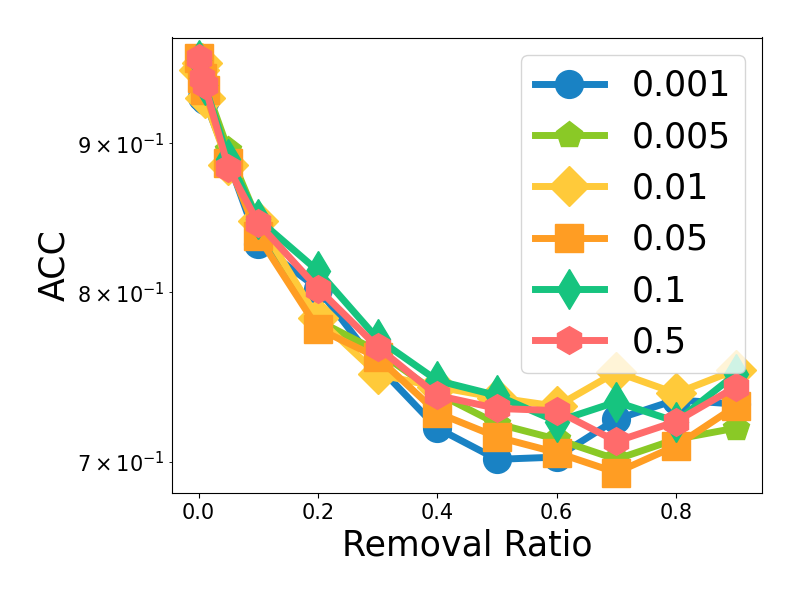}
        \caption{Average in HC3}
        \label{fig:appendix_lambda_hc3_avg}
  \end{subfigure} \\
\caption{{Comparison results of MORF between different $\lambda$ settings on HC3 dataset.}}
\label{fig:appendix_lambda_ablation}
\vspace{-0.5em}
\end{figure*}

%% file: experiments/table_lambda.tex
\begin{table}[!h]
\caption{{MORF average results of the $\lambda$ ablation study on HC3 dataset. Lower is better.}} 
\centering
\scalebox{0.8}{
    \begin{tabular}{cccccc}
        \toprule
    {$\lambda$} & {open-qa} & {wiki-csai} & {medicine} & {finance} & {Avg.}\\
     \midrule
     {0.001}  &  {0.73} & {0.72} & {0.97} & {\textbf{0.79}} &  {\textbf{0.80}} \\
     {0.005}  &  {0.72} & {0.74} & {\textbf{0.96}}  & {0.80} & {\textbf{0.80}} \\  
     {0.01}   &  {0.74} & {0.74} & {0.97}  & {0.80} & {0.81} \\
     {0.05}   &  {0.73} & {\textbf{0.71}} & {0.97}  & {\textbf{0.79}} & {\textbf{0.80}}  \\
     {0.1}    &  {0.74} & {0.74} & {0.97}  & {0.81} & {0.81} \\
     {0.5}    &  {\textbf{0.71}} & {0.76} & {0.97}  & {0.80} & {0.81} \\
    \bottomrule
    \end{tabular}
    }   
\label{tab:lambda_ablation}
\end{table}

%% file: experiments/statistics_analysis.tex
\begin{table*}[!h]
\caption{Statistics of text covered by explanation motifs on HC3 dataset. The sparsity of the explanation motifs is $0.05\%$.} 
\centering
\setlength\tabcolsep{3.3pt}
\scalebox{1.0}{
    \begin{tabular}{lcccccccc}  
        \toprule
      & \multicolumn{2}{c}{open-qa} 
       & \multicolumn{2}{c}{wiki-csai} 
       & \multicolumn{2}{c}{medicine} 
       & \multicolumn{2}{c}{finance}\\
       \cmidrule(r){2-3} \cmidrule(r){4-5} \cmidrule(r){6-7} \cmidrule(r){8-9}
     Statistic   & HGT & MGT   & HGT & MGT    & HGT & MGT  & HGT & MGT   \\
    \midrule
    Nodes    & 610	 &2407	 &1685	&777	&923	&990	&1251	&618	\\
    Edges   & 277	&3496	&2180	&1993	&797	&2086	&2004	&1816 \\
    \midrule
    Nodes/Edges   & 2.20	&0.69	&0.77	&0.39	&1.16	&0.47	&0.62	&0.34\\
    \bottomrule
    \end{tabular}
    }
\label{tab:statistic_motifs}
\vspace{-0.8em}
\end{table*}
\begin{table*}[!h]
\caption{Statistics of text covered by explanation motifs on RAID dataset. The sparsity of the explanation motifs is $0.05\%$.} 
\centering
\setlength\tabcolsep{3.3pt}
\scalebox{1.0}{
    \begin{tabular}{lcccccccc}
        \toprule
      & \multicolumn{2}{c}{recipes} 
       & \multicolumn{2}{c}{book} 
       & \multicolumn{2}{c}{poetry} 
       & \multicolumn{2}{c}{review} \\ 
       \cmidrule(r){2-3} \cmidrule(r){4-5} \cmidrule(r){6-7} \cmidrule(r){8-9}
     Statistic   & HGTs & MGTs   & HGTs & MGTs    & HGTs & MGTs  & HGTs & MGTs  \\
    \midrule
    Nodes        & 1100	 &458	 &4093	&1116	&2892	&760	&2674	&1560	\\
    Edges   & 3519  &2567	 &8583	&4163	&7731	&3452	&5100	&5791	\\
    \midrule
    Nodes/Edges   & 0.31	&0.18	&0.48	&0.27	&0.37	&0.22	&0.52	&0.27	\\
    \bottomrule
    \end{tabular}
    }   
\label{tab:statistic_motifs_raid}
\vspace{-0.8em}
\end{table*}
\begin{table*}[!h]
\caption{Statistics of text covered by explanation motifs on M4 dataset. The sparsity of the explanation motifs is $0.05\%$.} 
\centering
\setlength\tabcolsep{3.3pt}
\scalebox{1.0}{
    \begin{tabular}{lcccccccc}
        \toprule
      & \multicolumn{2}{c}{wikihow} 
       & \multicolumn{2}{c}{reddit} 
       & \multicolumn{2}{c}{peerread} 
       & \multicolumn{2}{c}{arxiv} \\
       \cmidrule(r){2-3} \cmidrule(r){4-5} \cmidrule(r){6-7} \cmidrule(r){8-9}
     Statistic   & HGT & MGT   & HGT & MGT    & HGT & MGT  & HGT & MGT   \\
    \midrule
    Nodes        & 4207	 &1894	 &3929	&1282	&2138	&609	&1449	&725	\\
    Edges   & 19511 &8819	 &8448	&2770	&10937	&6044	&2954	&2112	\\
    \midrule
    Nodes/Edges   & 0.22	&0.21	&0.47	&0.46	&0.20	&0.10	&0.49	&0.34	\\
    \bottomrule
    \end{tabular}
    }   
\label{tab:statistic_motifs_m4}
\end{table*}

%% file: figures/case_table.tex
\begin{table}[h]
    \caption{Samples of token-level explanation motifs.}
    \centering
    \scalebox{0.7}{
     \begin{tabular}{ccc}
        \toprule
          &  Graph Motifs & Words Mapping    \\  
        \midrule
        \rotatebox[origin=c]{90}{GPT-4}
        &\begin{minipage}[b]{0.48\columnwidth}
            \centering
            \vspace{2pt}
            \raisebox{-.5\height}{\includegraphics[width=\linewidth]{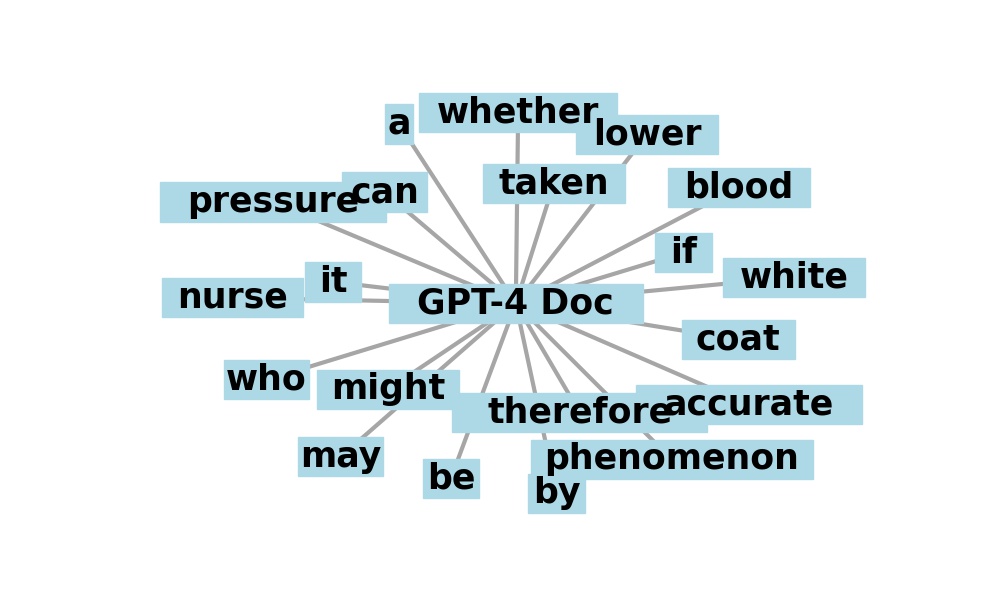}}
        \end{minipage} 
        & \begin{minipage}[b]{0.48\columnwidth}
            \centering
            \vspace{2pt}
            \raisebox{-.5\height}{\includegraphics[width=\linewidth, trim=0 30 50 0, clip]{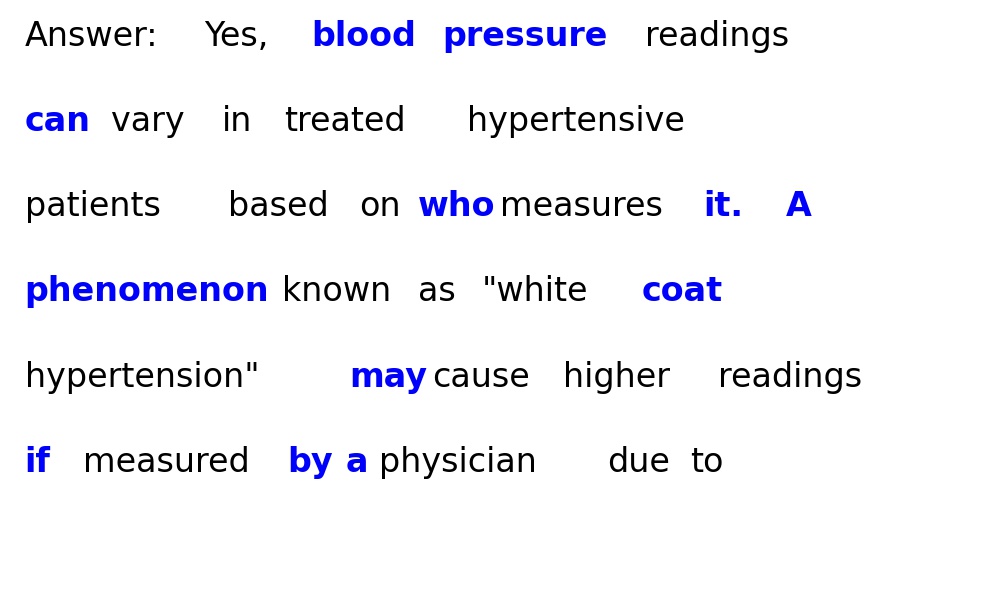}}
        \end{minipage} \\
        \rotatebox[origin=c]{90}{Claude-3} 
        & 
        \begin{minipage}[b]{0.48\columnwidth}
            \centering
            \vspace{2pt}
            \raisebox{-.5\height}{\includegraphics[width=\linewidth]{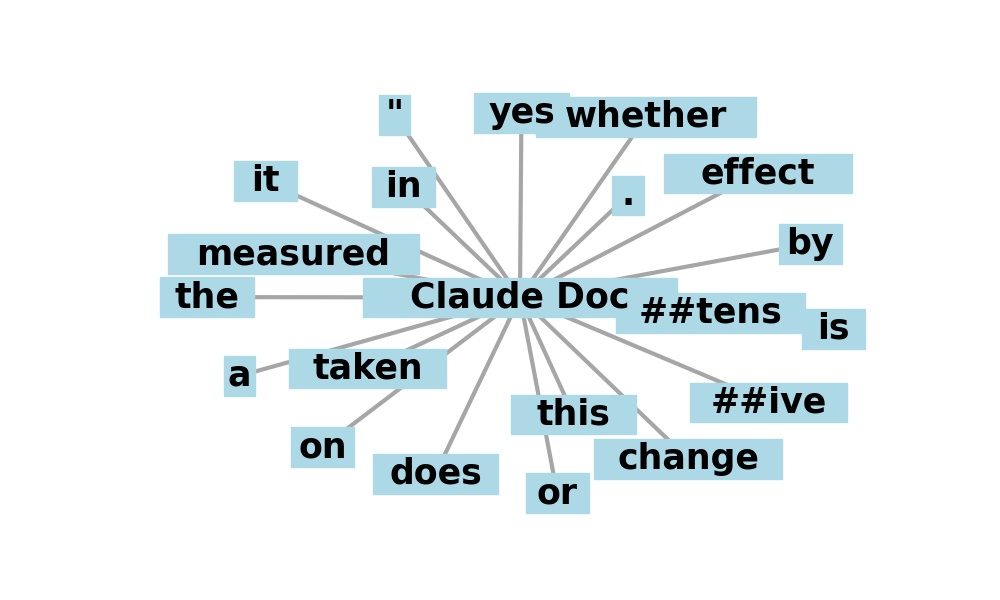}}
        \end{minipage}
        & \begin{minipage}[b]{0.48\columnwidth}
            \centering
            \vspace{2pt}
            \raisebox{-.5\height}{\includegraphics[width=\linewidth, trim=0 30 50 0, clip]{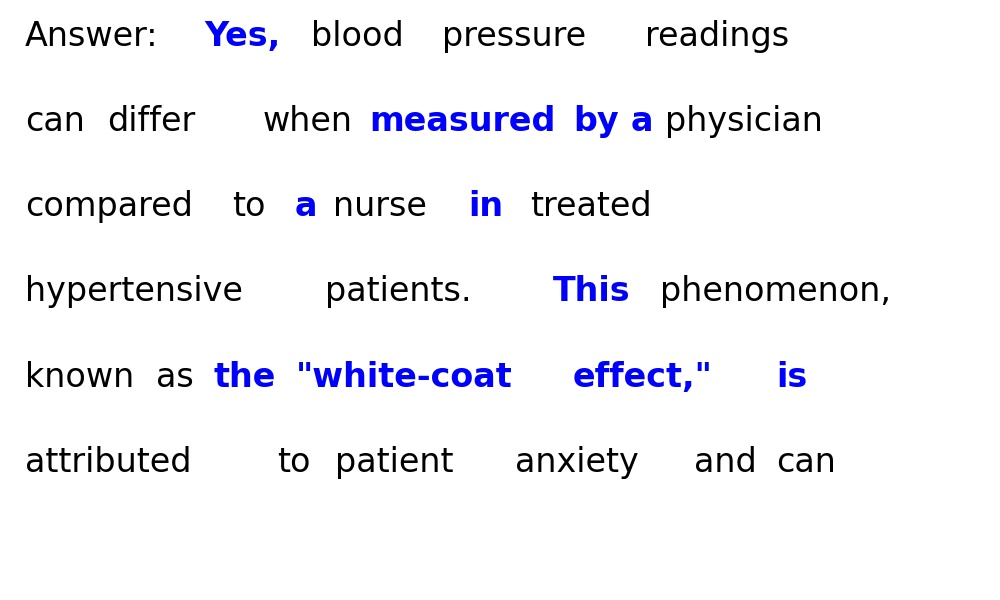}}
        \end{minipage} 
        \\
        \rotatebox[origin=c]{90}{Davinci}
        & 
        \begin{minipage}[b]{0.48\columnwidth}
            \centering
            \vspace{2pt}
            \raisebox{-.5\height}{\includegraphics[width=\linewidth]{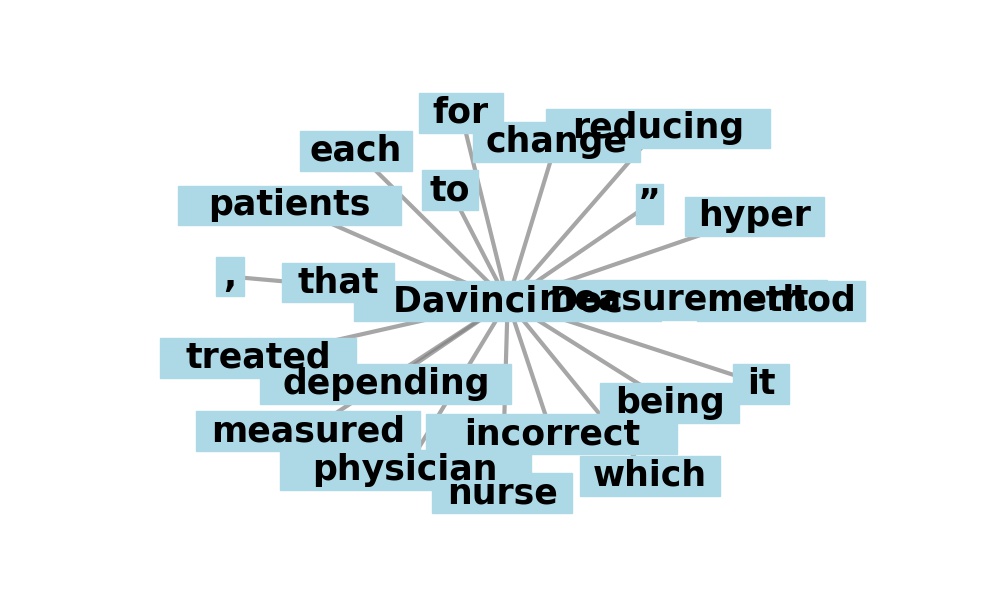}}
        \end{minipage} 
        & \begin{minipage}[b]{0.48\columnwidth}
            \centering
            \vspace{2pt}
            \raisebox{-.5\height}{\includegraphics[width=\linewidth, trim=0 60 50 0, clip]{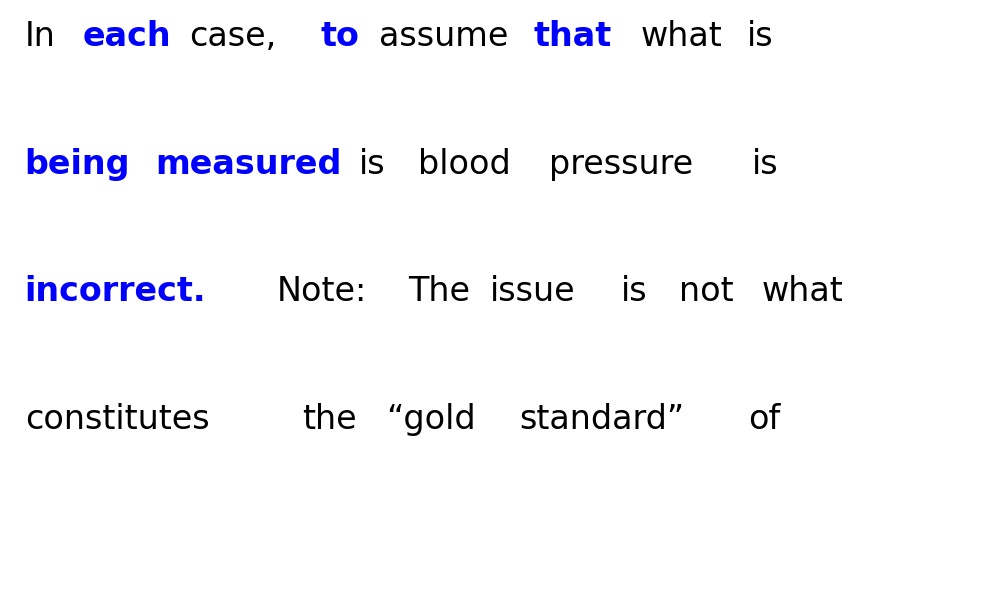}}
        \end{minipage} 
        \\
        \bottomrule
    \end{tabular}}
    \label{tab:fig:casestudy_words}
\end{table}
\begin{table}[h]
    \caption{Samples of token-token co-occurrence explanation motifs.}
    \centering
    \scalebox{0.7}{
     \begin{tabular}{ccc}
        \toprule
          &  Graph Motifs & Words Mapping    \\  
        \midrule
        \rotatebox[origin=c]{90}{GPT-4}
        &\begin{minipage}[b]{0.49\columnwidth}
            \centering
            \vspace{2pt}
            \raisebox{-.5\height}{\includegraphics[width=\linewidth, trim=60 60 60 60, clip]{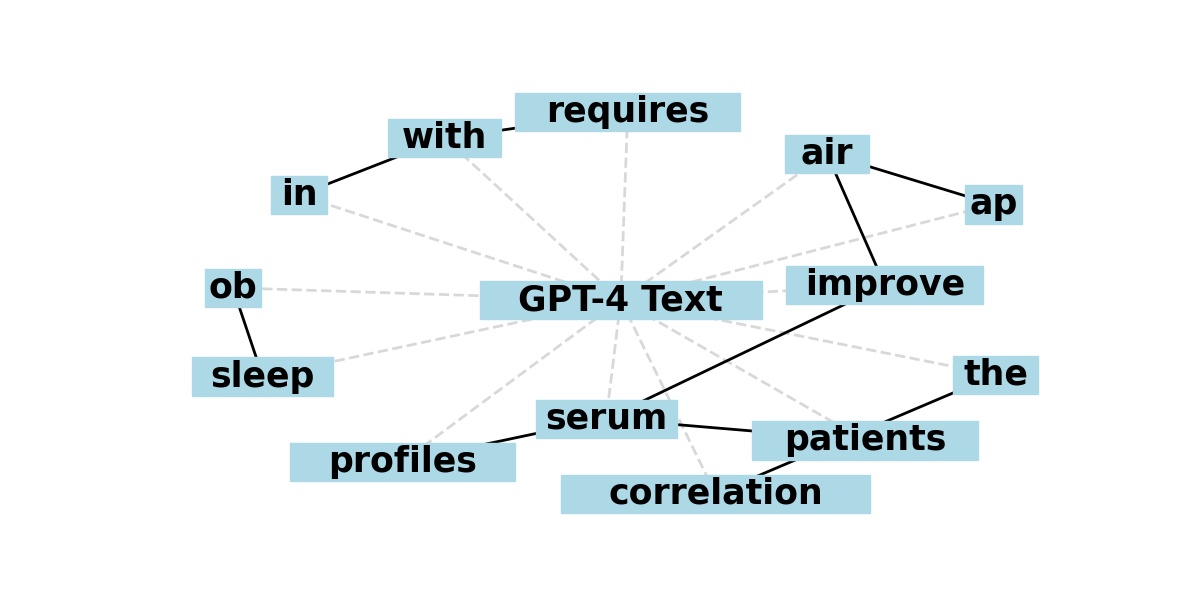}}
        \end{minipage} 
        & \begin{minipage}[b]{0.49\columnwidth}
            \centering
            \vspace{2pt}
            \raisebox{-.5\height}{\includegraphics[width=\linewidth, trim=0 30 60 0, clip]{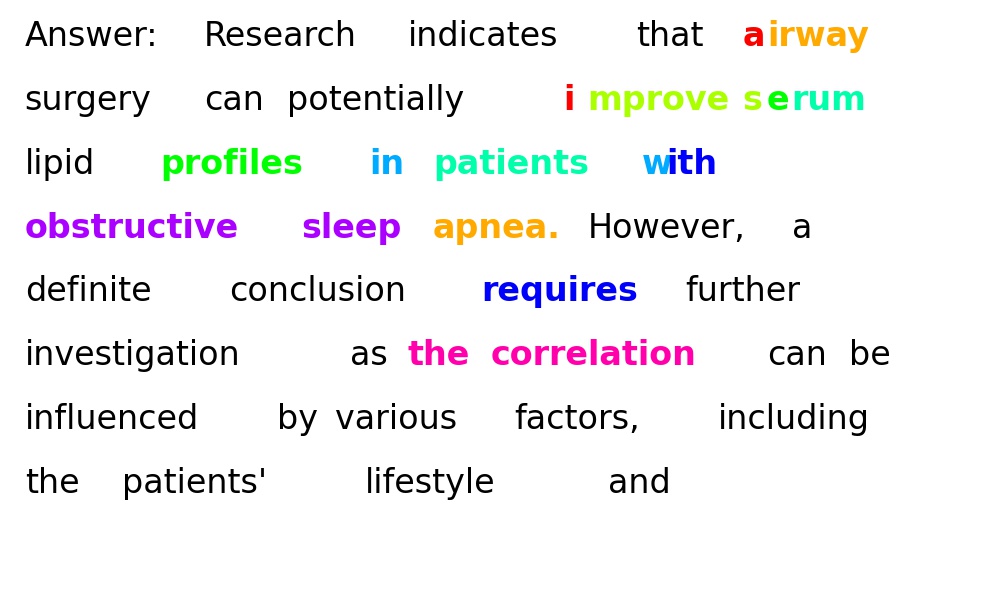}}
        \end{minipage} \\
        \rotatebox[origin=c]{90}{Claude-3} 
        & 
        \begin{minipage}[b]{0.49\columnwidth}
            \centering
            \vspace{2pt}
            \raisebox{-.5\height}{\includegraphics[width=\linewidth, trim=60 60 60 60, clip]{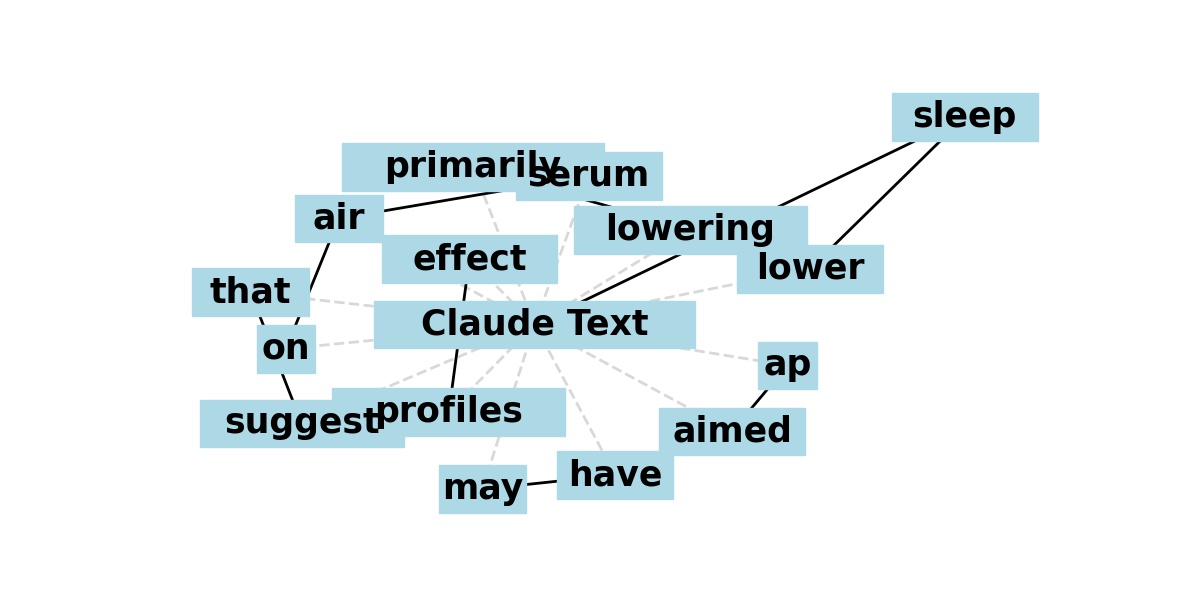}}
        \end{minipage}
        & \begin{minipage}[b]{0.49\columnwidth}
            \centering
            \vspace{2pt}
            \raisebox{-.5\height}{\includegraphics[width=\linewidth, trim=0 30 60 0, clip]{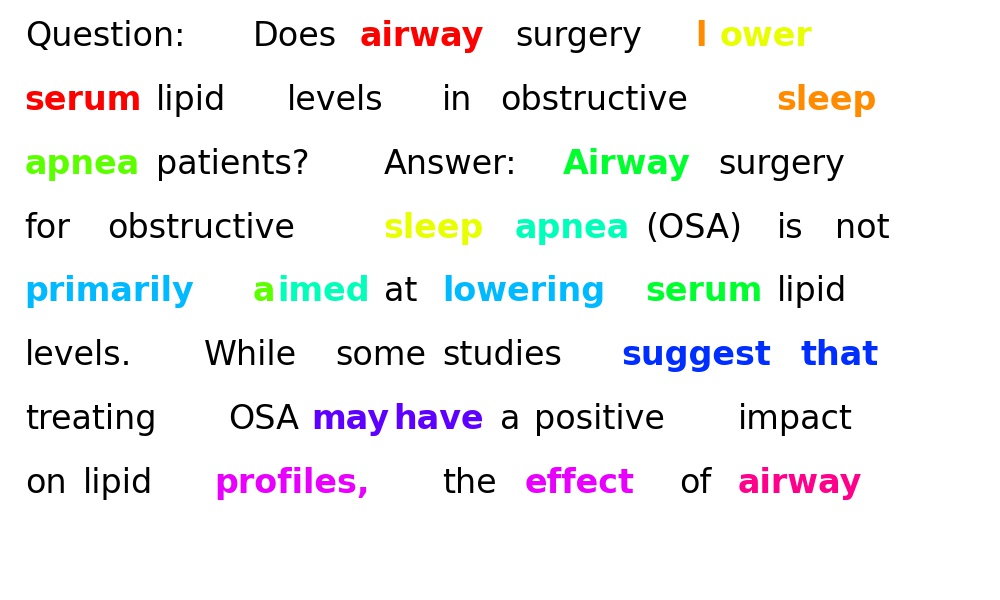}}
        \end{minipage} 
        \\
        \rotatebox[origin=c]{90}{Davinci}
        & 
        \begin{minipage}[b]{0.49\columnwidth}
            \centering
            \vspace{2pt}
            \raisebox{-.5\height}{\includegraphics[width=\linewidth, trim=60 60 60 60, clip]{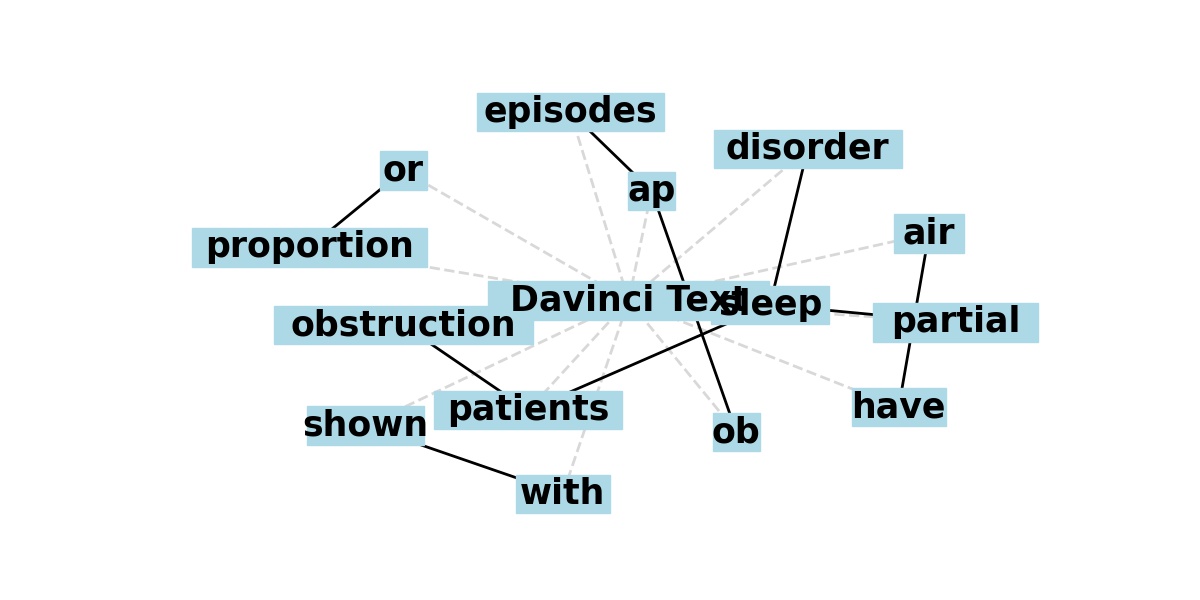}}
        \end{minipage} 
        & \begin{minipage}[b]{0.49\columnwidth}
            \centering
            \vspace{2pt}
            \raisebox{-.5\height}{\includegraphics[width=\linewidth, trim=0 30 60 0, clip]{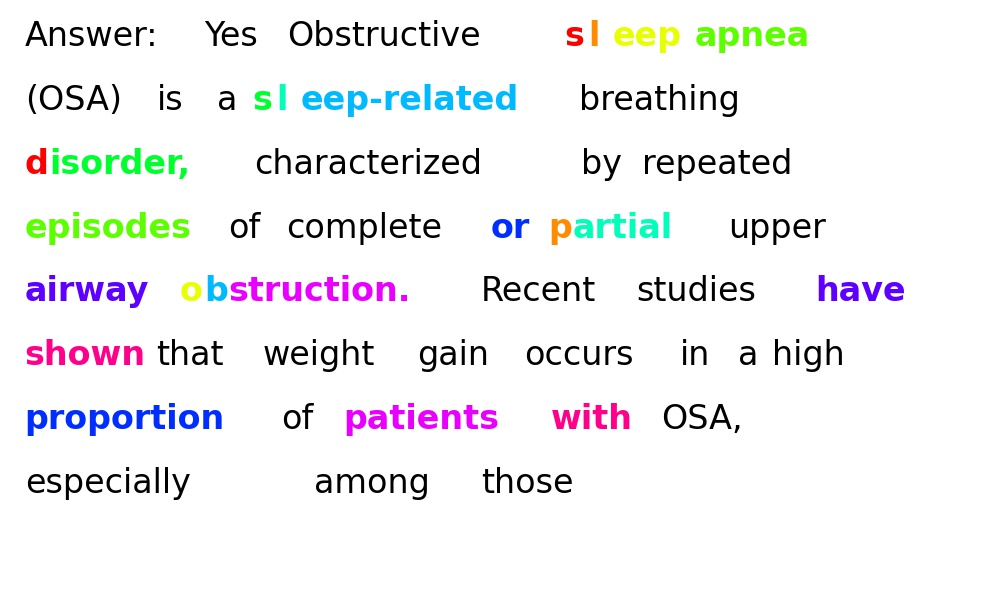}}
        \end{minipage} 
        \\
        \bottomrule
    \end{tabular}}
    \label{tab:fig:casestudy_phases}
\end{table}